%% file: main.tex
\documentclass[twoside,11pt]{article}
\usepackage[preprint]{jmlr2e}
\usepackage{amsmath}
\usepackage{booktabs,multirow,subcaption,longtable}
\usepackage{microtype}
\usepackage{needspace}
\usepackage{lastpage}
\hypersetup{hidelinks}
\graphicspath{{figures/}{../../results/figures/}}
\renewenvironment{proof}[1][Proof]{%
  \par\noindent\textbf{#1.}\hspace{0.4em}\ignorespaces
}{%
  \hfill\ensuremath{\square}\par\medskip
}
\newcounter{algorithm}
\makeatletter
\def\ftype@algorithm{4}
\def\ext@algorithm{loa}
\newenvironment{algfloat}[1][t]{\@float{algorithm}[#1]}{\end@float}
\makeatother

\newcommand{\E}{\mathbb{E}}
\newcommand{\R}{\mathbb{R}}

\jmlrheading{1}{2026}{1-\pageref{LastPage}}{11/26}{}{00-0000}{Mrigank Shekhar Singh}
\ShortHeadings{Finite-Cadence Equilibria for Prior-Anchored Ridge}{Singh}
\firstpageno{1}

\begin{document}
\title{Finite-Cadence Refresh Equilibria for Streaming Prior-Anchored Ridge: A Closed-Form Forgetting Window and Spectral Correction}
\author{\name Mrigank Shekhar Singh \email mrigank.codes@gmail.com \\
        \addr DLDAV Model School, Shalimar Bagh\\
        Delhi 110088, India}
\editor{}
\maketitle
\begin{abstract}%
Recursive estimators on drifting streams must forget stale data fast enough to track the target, yet slowly enough to average out noise.
In prior-anchored ridge regression, this trade-off admits a closed-form spectral decomposition in which each eigen-direction interpolates between the data solution and the prior with pull weight $s_i=\rho/(\mu_i+\rho)$.
When the anchor is refreshed every $k$ steps, we show that the moving-frame offset converges to a finite-cadence equilibrium that lags the pure-leaky path by an explicit $\Delta_k(s)$ while reducing estimation variance by a factor $\Psi(s,r_k)$.
Minimizing the cycle-averaged risk gives $\omega^*=(4\nu^2_\Sigma/c')^{1/3}$, $c'=\sigma^2d_{\mathrm{eff}}(\rho)$, and the plug-in strength $\rho^*=\sigma^2/(n\delta^2)$.
A random-matrix deterministic equivalent extends both laws to finite dimension, renormalizing the anchor to $\rho_{\mathrm{eff}}=\rho(e^{\zeta q}-1)/(\zeta q)$; it matches simulation within $5\%$ at fixed operating points and within $13\%$ at the method's own, and rank-deficient populations reduce exactly to their active subspace.
The resulting method, SPAR, calibrates window and strength online from residuals.
On a 27-configuration grid the calibrated window opens on drift without divergence, and the grid defaults transfer to an hourly electricity stream and to BCI IV 2b.
The core algebra, refresh chain, and regret bounds are formalized in Lean 4.
\end{abstract}

\begin{keywords}
streaming regression, prior-anchored ridge, forgetting factor, drift adaptation, deterministic equivalent
\end{keywords}

\section{Introduction}
A streaming estimator that never forgets accumulates lag under drift; one that forgets too fast amplifies noise.
Gradient-based test-time adaptation \citep{sun2020,wang2021tent} resolves this balance implicitly, at the cost of a gradient channel, per-data set hyperparameter selection, and imprecise guarantees.
Prior-anchored ridge regression makes this balance explicit. In the eigenbasis of the sample covariance, the anchored solution decomposes coordinatewise as $W_i - w_i = (1-s_i)(\widehat w_i - w_i) + s_i(G_i - w_i)$ with pull weight $s_i = \rho/(\mu_i + \rho)$.
While the underlying MAP algebra with a prior-centered penalty is classical \citep{hoerl1970,rasmussen2006,li2018l2sp}, this spectral view turns hyperparameter selection into a closed-loop estimation problem, in which measured conflict lowers the anchor strength so fresh data dominates and a rising drift-rate estimate opens the forgetting window.

Rather than proposing an ad hoc tuning heuristic, we analyze the closed-loop dynamics of the resulting recursion.
At refresh cadence $k$ the moving-frame anchor offset follows an affine recursion whose fixed point gives the exact tracking lag and whose variance factor gives the retained noise reduction.
Minimizing the cycle-averaged risk yields the optimal forgetting window in closed form.
A separable-covariance deterministic equivalent then extends the law to finite dimension, where the finite window renormalizes the anchor strength and amplifies the risk by a scalar factor.
The calibrated method, SPAR, implements these results; the experiments test the theory on synthetic and real streams.

\needspace{4\baselineskip}
\emph{Contributions.}
\begin{enumerate}
\item \emph{A finite-cadence theory of anchored forgetting.}
The pull identity turns the anchor strength into a per-direction interpolation weight.
The refresh equilibrium gives the exact lag $\Delta_k(s)$ and variance factor $\Psi(s,r_k)$ (Theorem~\ref{thm:cadence}); minimizing the cycle-averaged risk yields the closed-form window $\omega^*=(4\nu^2_\Sigma/c')^{1/3}$ (Theorem~\ref{thm:window}) and the plug-in strength $\rho^*=\sigma^2/(n\delta^2)$ (Proposition~\ref{prop:conflict}).

\item \emph{A finite-dimensional spectral correction.}
The separable-covariance deterministic equivalent gives the retention $q$, the renormalized anchor $\rho_{\mathrm{eff}}$, the variance functional $\Upsilon$, and the full excess risk with its $1/(1-\Upsilon)$ correction (Theorem~\ref{thm:rmt}).
The formula matches Monte Carlo within $4.0\%$ across the $45$-point grid (including the rank-deficient spectrum), within $13\%$ at the method's own operating points, and within $10.0\%$ for the conflict increment; rank-deficient populations reduce exactly to their active subspace (Corollary~\ref{cor:rmt-active}).

\item \emph{A formalized core.}
The algebraic core, the finite-cadence refresh chain, the OGD and Hedge regret bounds, and the coupled contraction are formalized in Lean 4 \citep{demoura2021,mathlib2020}: 111 declarations, 9 non-vacuity witnesses, no \texttt{sorry}, and only the three standard axioms (Appendix~\ref{app:lean}).

\item \emph{A calibrated streaming method.}
The forgetting window and the anchor strength are estimated online from residual and anchor-disagreement statistics; prior immunity is carried by the direction-selective refresh $\beta_i = 1 - s_i$ (one $\rho$ sets it), and the strength rule enters as an onset guard when the warm-up prior scale is unknown, not as a continuous tracking rule.
A Hedge composition over leaky windows, the frozen anchor, and the SPAR-Pull trajectory satisfies a no-regret bound under bounded losses (Proposition~\ref{prop:ocowrap}).

\item \emph{Evidence that the law transfers.}
On the 27-configuration grid the calibrated window responds to drift in the direction the law prescribes, and the calibrated methods are competitive with the strongest oracle-free baselines without diverging in any configuration.
The grid-default configuration transfers unchanged to an hourly electricity stream, and a reduced-range head transfers to BCI IV 2b (Section~\ref{sec:experiments}); Section~\ref{sec:limitations} discusses limitations.
\end{enumerate}

\begin{figure}[t]
\centering
\includegraphics[width=0.95\textwidth]{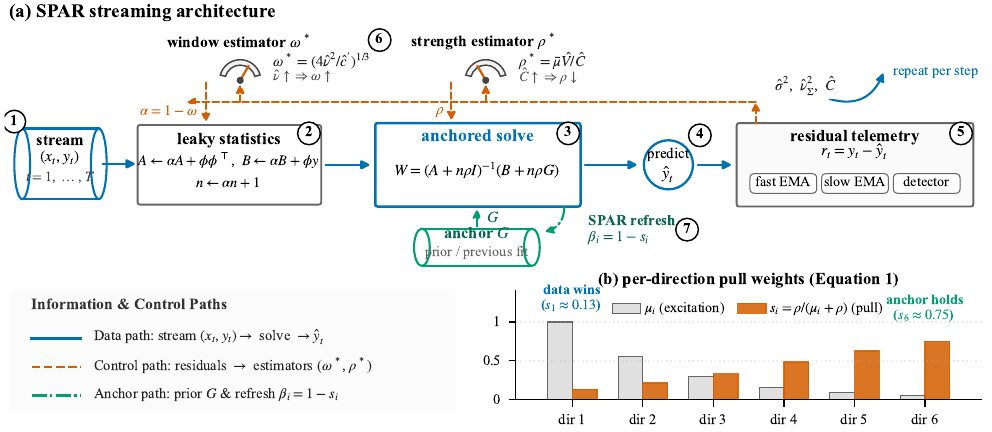}
\caption{The SPAR streaming loop.
(1) The stream enters (2) leaky statistics and (3) the anchored solve (Algorithm~\ref{alg:spar-pull}), which (4) predicts.
(5) The residual $r_t$ feeds fast and slow moving averages whose positive difference drives (6) the two online estimators ($\rho^*$ and $\omega^*$), which feed the solve and the statistics.
(7) The anchor refreshes at $\beta_i = 1-s_i$; one $\rho$ couples pull and refresh.
Panel (b): per-direction pull weights.}
\label{fig:schematic}
\end{figure}

\section{Related Work}
\emph{Analytic continual learning.}
The Analytic Class-Incremental Learning lineage \citep{zhuang2022acil,zhuang2023gkeal,zhuang2024dsal,zhuang2024gacl,tong2025anyssr,yan2025crossacl,zhang2025mdaa,kim2025ulta} replaces gradient-based class-incremental learning with closed-form recursive updates on frozen features; these methods address a stationary target, whereas the present setting has a drifting target and an adaptive regularization strength.

\emph{Interpolated adaptation.}
WiSE-FT established frozen-to-adapted interpolation with a single scalar \citep{wortsman2022wiseft}; the pull identity generalizes it per direction ($s_i = \rho/(\mu_i + \rho)$ in place of one $\alpha$), Proposition~\ref{prop:conflict} characterizes when interpolation reduces risk, and Section~\ref{sec:eeg} compares against it directly.

\emph{Forgetting and state space.}
When the nominal window $1/(1-\lambda)$ of leaky RLS drops below the input dimension, the covariance recursion loses rank and the estimate diverges \citep{sayed2008,bierman1977}.
Directional forgetting \citep{kulhavy1984,cao2000} and its subspace-information variants \citep{lai2024sift} preserve rank along excited directions; scalar EMA teachers \citep{tarvainen2017} update every direction at one fixed rate, and per-direction momentum \citep{hong2026} at excitation-dependent rates.
The SPAR refresh is direction-selective with no added hyperparameter: each direction's refresh rate is $1-s_i$, the complement of its pull weight.
Variable forgetting factor RLS \citep{fortescue1981,paleologu2008} automates the window from the squared residual, the same statistic used here, but it retains the covariance recursion and inherits its instability (Section~\ref{sec:experiments}).
AROW \citep{crammer2009arow} and random-walk Kalman filters \citep{kalman1960,sayed1994} are the nearest online relatives; the estimated-noise Kalman filter is the strongest oracle-free baseline on the grid (Table~\ref{tab:grid}).
Composite drift-triggered adaptations are surveyed in \citet{gama2014drift}.

\emph{Regret and two-time-scale analysis.}
The residual-driven estimators are coupled stochastic approximation: the accumulator tracks a moving target while its step size is estimated from the same residuals, for which two-time-scale conditions and drift tracking are classical \citep{benveniste1990,kushner2003,borkar2008}.
The contribution here is the closed-form target and the verified contraction.
Projected gradient, Hedge, and exponentiated-gradient arguments are standard \citep{zinkevich2003,freund1997,kivinen1997}; the composition and its basis dependence are made explicit in Proposition~\ref{prop:ocowrap}.

\emph{Random-matrix limits.}
The finite-\(D\) correction uses the separable-covariance deterministic equivalent \citep{silverstein1995,rubio2011,hachem2007}.
The nearest prior analysis is the large-system exponentially windowed RLS study of \citet{xiao2005}, which has no anchor, per-direction pull, refresh, or online calibration; the unweighted-batch ridge asymptotics \citep{hastie2022,pafka2004} do not carry the finite-window renormalization.

\emph{What is new.}
The nearest analyses differ from this work in three respects.
(i) Against the exponentially windowed RLS of \citet{xiao2005}, whose large-system limit has no anchor, per-direction pull, refresh, or online calibration, the finite-cadence refresh equilibrium adds the closed-form window law $\omega^*=(4\nu^2_\Sigma/c')^{1/3}$ and couples pull and refresh through $\beta_i=1-s_i$; the nearest prior has neither.
(ii) Against the unweighted-batch ridge asymptotics \citep{hastie2022,pafka2004,silverstein1995,rubio2011,hachem2007}, which are the $\zeta\to0$ limit, the streaming deterministic equivalent carries the finite-window renormalization $\rho_{\mathrm{eff}}=\rho(e^{\zeta q}-1)/(\zeta q)$ and the $1/(1-\Upsilon)$ variance amplification; on full-rank spectra it matches simulation to $4.0\%$ across the whole grid (Section~\ref{sec:experiments}).
(iii) Against single-scalar interpolation (WiSE-FT) and scalar-$\lambda$ forgetting, the anchor strength is per direction, $\rho/(\mu_i+\rho)$, and it also sets the refresh rate; the threshold is calibrated online from the measured conflict, and the coupling product stays below one at the operating points.
Table~\ref{tab:related} summarizes the coverage.

\begin{table}[t]
\centering
\footnotesize
\setlength{\tabcolsep}{4pt}
\caption{What is new relative to the nearest lines of work. ``DE'' is the finite-window deterministic-equivalent correction; ``Lean'' the machine-checked core.}
\label{tab:related}
\resizebox{\textwidth}{!}{%
\begin{tabular}{@{}p{1.72in}cccccc@{}}
\toprule
family & anchor & per-dir.\ pull & refresh & online window/strength & finite-window DE & Lean \\
\midrule
leaky / EW-RLS \citep{sayed2008,xiao2005} & -- & -- & -- & -- & $\zeta\to0$ & -- \\
VFF-RLS \citep{fortescue1981,paleologu2008} & -- & -- & -- & window & -- & -- \\
directional forgetting \citep{kulhavy1984,cao2000,lai2024sift} & -- & subspace & -- & -- & -- & -- \\
AROW \citep{crammer2009arow} & confidence & per-feature & -- & -- & -- & -- \\
Kalman / state space \citep{kalman1960,sayed1994} & process & -- & -- & $q$ fixed & -- & -- \\
WiSE-FT \citep{wortsman2022wiseft} & frozen model & -- & -- & -- & -- & -- \\
ACIL \citep{zhuang2022acil} & frozen features & -- & stationary & -- & -- & -- \\
batch-ridge DE \citep{silverstein1995,rubio2011,hachem2007} & -- & -- & -- & -- & unweighted & -- \\
\textbf{SPAR (this work)} & prior $G$ & $s_i=\frac{\rho}{\mu_i+\rho}$ & $\beta_i=1-s_i$ & both & $\rho_{\mathrm{eff}}$ & yes \\
\bottomrule
\end{tabular}}
\end{table}

\section{Method}
We call the resulting solver family SPAR (Streaming Prior-Anchored Ridge), the streaming instantiation of the Anchored Analytic Learning (AAL) framework.

\emph{Notation.}
$\omega \in (0,1)$ is the scalar forgetting rate and $n_{\mathrm{eff}} = (2-\omega)/\omega$ its effective sample count; $\widehat w_t$ and $w^*$ are parameter vectors (we write $\widehat w_t$ for the minimum-ridge data solve and reserve $W_t$ for the anchored solution); $\pi_{i,t}$ are Hedge weights; $\varsigma_t$ is the Hedge learning rate; $\iota$ is the OGD step; $\zeta = D\omega$ is the aspect ratio of Theorem~\ref{thm:rmt}. For the strength estimator, $d_{\mathrm{act}} = \max(1,\#\{i:\mu_i(\bar A_t) > 10^{-8}\max_j\mu_j(\bar A_t)\})$ is the active dimension, $\bar\mu_t = \mathrm{tr}(\bar A_t)/D$ the mean excitation, and $G_{\mathrm{ref}}$ the slow reference anchor, an exponential moving average of the refreshed solve with a $2000$-step horizon. Experiments label windows by the nominal $1/(1-\lambda)$, which differs from $n_{\mathrm{eff}}$ by the factor $2-\omega$.

\begin{figure}[t]
\centering
\includegraphics[width=0.68\textwidth]{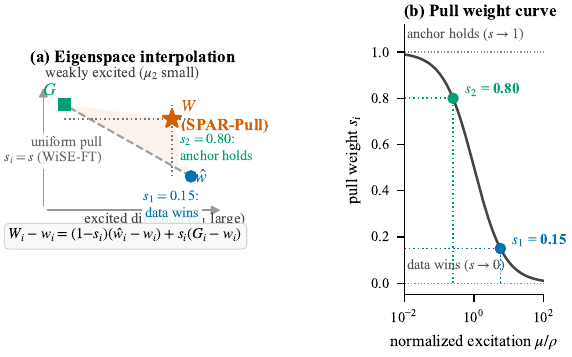}
\caption{The pull identity (Equation~\eqref{eq:pull}): the excited direction ($s_1 = 0.15$) holds the data solution, the weakly excited direction ($s_2 = 0.8$) holds the anchor, so $W$ departs from the uniform-pull segment (dashed).}
\label{fig:geometry}
\end{figure}

\emph{Anchored solve.}
\label{sec:method-solve}
Let $\phi_t = \phi(x_t)$ denote the feature vector of the incoming sample.
With optional forgetting factor $\alpha$, the learner updates the sufficient statistics via $A \leftarrow \alpha A + \phi\phi^\top$, $B \leftarrow \alpha B + \phi y$, and $n \leftarrow \alpha n + 1$.
The anchored solve is
\[
  W = (A + n\rho I)^{-1}(B + n\rho G),
\]
with anchor $G$ and strength $\rho$.
In the eigenbasis of $\bar A = A/n$ with excitation $\mu_i$, each coordinate is a scalar ridge solve, which yields the per-direction decomposition we call the \emph{pull identity} (the $w_i$ and $\widehat w_i$ are per-direction weight coordinates; the scalar forgetting rate $\omega$ of Section~\ref{sec:theory} is a separate object):
\begin{equation}
\label{eq:pull}
  W_i - w_i = (1-s_i)\,(\widehat w_i - w_i) + s_i\,(G_i - w_i),
  \qquad s_i = \frac{\rho}{\mu_i + \rho};
\end{equation}
Figure~\ref{fig:geometry} shows its geometry.
The solve in Equation~\eqref{eq:pull} is the $\gamma = 1$ case of the generalized prior scaling $W = (A + n\rho I)^{-1}(B + n\rho\gamma G)$: $\gamma = 1$ is the exact prior-centered MAP and $\gamma < 1$ adds a shrinkage bias (Section~\ref{sec:experiments}); the prior precision $\tau_0$ of Appendix~\ref{app:proofs} is a separate symbol.

\emph{Operator form.}
Both identities are basis-free: with the resolvent shrinkage operator $S_t(\rho) := \rho\,(\bar A_t + \rho I)^{-1}$ (so $0 \prec S_t(\rho) \preceq I$ and $I - S_t(\rho) = \bar A_t(\bar A_t + \rho I)^{-1}$), Equation~\eqref{eq:pull} reads $W_t - w^* = (I - S_t) (\widehat w_t - w^*) + S_t (G_t - w^*)$ for any $\bar A_t \succeq 0$, without choosing a basis, assuming distinct eigenvalues, or restricting to finite dimension.
The spectral refresh of the next paragraph, written per-direction as $\beta_i = 1 - s_i$, is likewise the operator identity $G \leftarrow (I - S_t)\,W + S_t\,G = W - \rho\,(\bar A_t + \rho I)^{-1}(W - G)$: since the factorization of $(\bar A_t + \rho I)$ is already available from the solve, the exact refresh can reuse it at $O(D^2)$ cost per refresh with no separate eigendecomposition (Algorithm~\ref{alg:spar-pull}'s \texttt{eigh} is the equivalent spectral realization used to expose the per-direction weights); this reduces the refresh cost, while the per-step solve remains $O(D^3)$ for the scalar method.

\emph{Spectral anchor refresh.}
To prevent the anchor from becoming stale during sustained drift, $G$ is updated every $k$ steps along the eigenbasis $U$ of $\bar A$: in each eigendirection $i$, $G_{i} \leftarrow \beta_i W_{i} + (1-\beta_i) G_{i}$ with $\beta_i = 1 - s_i = \mu_i/(\mu_i + \rho)$.
Because the prediction risk in direction $i$ scales with $\mu_i$, staleness along weakly excited eigendirections ($\beta_i \approx 0$) incurs negligible excess error; well-excited directions ($\beta_i \approx 1$) incorporate new information into the anchor rapidly.
A single $\rho$ thus controls both the pull toward the anchor and its refresh rate.

\emph{Online calibration.}
SPAR updates the window and the strength at each refresh block from fast and slow exponential moving averages of the squared prequential residual $r_t^2 = (y_t - \hat y_t)^2$; here $\omega$ is the forgetting rate, with effective memory $n_t \approx 1/\omega$ samples, so opening the window means increasing $\omega$.
Under Proposition~\ref{prop:conflict}, the risk-minimizing anchor strength decreases inversely with the prediction-unit conflict $\widehat C^\Sigma_t = \max\{C^\Sigma_{\mathrm{ref}}, C^\Sigma_{\mathrm{resid}}, C^\Sigma_{\mathrm{floor}}\}$: $\rho^*_t = \bar\mu_t \widehat V^\Sigma_t/\widehat C^\Sigma_t$, with $\widehat V^\Sigma_t = \hat\sigma^2_t d_{\mathrm{act}}/n_t$ the prediction variance and every branch in $[y^2]$ (Online Appendix~\ref{app:hyper}).
Formulating both branches in prediction units ensures that the estimator never inverts $\bar A_t$, is invariant to feature rescaling, and contributes nothing in dead directions. Section~\ref{sec:experiments} reports the covariate-shift behavior.
Under Theorem~\ref{thm:window}, the forgetting window trades estimation variance against drift lag: $\omega^*_t = (4\widehat\nu^2_{\Sigma,t} / \hat c'_t)^{1/3}$, with $\hat c'_t = (1-\bar s_t)^2 \hat\sigma^2_t D$ and the drift energy $\widehat\nu^2_{\Sigma,t} = \max(0, (\widehat E_t - \hat c'_t \omega_t/(2-\omega_t))\omega_t^2)$, where $\widehat E_t = \hat\sigma^2_{\mathrm{fast},t} - \hat\sigma^2_{\mathrm{slow},t}$ isolates excess error above the stationary variance level, the subtraction required for a positive fixed point (Appendix~\ref{app:proofs}).
The strength rule is an onset guard: its average effect is neutral, and its value is the temporary lower bound it places on $\rho$ when the residual detector fires (Section~\ref{sec:bridges}).
Algorithm~\ref{alg:spar-pull} gives the complete step loop. The only fixed controls are the default constants $k{=}10$, $h_f{=}30$, the $\rho$-lower bound $0.02$, and the clip ranges and EMA horizons, shared by every configuration (Online Appendix~\ref{app:hyper}); all other behavior adapts at runtime.

\emph{Estimator branches and solver variants.}
Step~(iv) of the algorithm takes the maximum of three prediction-unit branches, all in $[y^2]$:
\[
\begin{array}{@{}l@{\qquad}l@{}}
C^\Sigma_{\mathrm{ref}} = (G_{\mathrm{ref}} - \widehat w)^\top \bar A_t\,(G_{\mathrm{ref}} - \widehat w) - \widehat V^\Sigma, & C^\Sigma_{\mathrm{resid}} = E_t, \\
C^\Sigma_{\mathrm{floor}} = \tfrac12\,\hat\sigma^2 d_{\mathrm{act}}/T_w, & \widehat V^\Sigma = \hat\sigma^2 d_{\mathrm{act}}/n,
\end{array}
\]
with $\widehat w = (\bar A_t + \rho_{\mathrm{ridge}} I)^{-1}\bar b_t$, where $\rho_{\mathrm{ridge}} = 10^{-6}$ is the fixed numerical ridge of the reference-branch data solve ($10^{-8}$ is the ridge used in the active-dimension and trace computations), and $\rho_{\min} = 0.02$ while the window is at least half open ($\omega_t \ge \tfrac12\omega_{\max}$), otherwise the stationary value $10^{-5}$, is the anchor-strength clip floor.
The reference branch compares against the slow copy $G_{\mathrm{ref}}$ rather than the refreshed $G$, so the conflict does not collapse immediately after a refresh. The residual branch supplies the fast disengagement, and the floor branch is the irreducible warm-up conflict.
No branch inverts $\bar A_t$, so the rule is invariant to feature rescaling and immune to the rank-deficient trace blow-up.
The legacy parameter-unit form (with $\mathrm{tr}(\bar A^{-1})$) is reported only as an ablation (Online Appendix~\ref{app:hyper}).

\begin{algfloat}[t]
\refstepcounter{algorithm}\label{alg:spar-pull}
\noindent\begin{minipage}{\textwidth}
\hrule\vspace{3pt}
\noindent\textbf{Algorithm \thealgorithm: SPAR-Pull}\quad{\footnotesize Inputs: a stream $(\phi_t,y_t)$, a warm-up length $T_w$, and a prior $G_0$; fixed constants $k$, $h_f$, $\rho_{\min}$, $\rho_{\max}$, $\eta_C$, and $\eta_\nu$.}\vspace{3pt}\hrule\vspace{3pt}
{\footnotesize
\emph{Initialization.} Build the sufficient statistics $A$, $B$, and $n$, the anchor $G$, and the solve $W$ from the warm-up stream. Set $\sigma^2_0 = \mathrm{var}(r_{\mathrm{warm}})$. For the first $K_{\mathrm{cal}} = 10$ refresh blocks the strength is capped at its initial value and the window held at $\omega_{\min}$, a burn-in for the two online estimators.

\emph{Loop.} For $t = T_w,\dots,T$, execute Steps~(i), (ii), (vii), and~(viii) at every step; at the start of each refresh block (every $k$ steps) run Steps~(iii)--(vi) on the current statistics, and at the end of the block run the refresh Step~(ix).
\begin{enumerate}
\renewcommand{\labelenumi}{(\roman{enumi})}
\setlength{\itemsep}{1pt}
\setlength{\parsep}{0pt}
\setlength{\topsep}{2pt}
\setlength{\partopsep}{0pt}
\item Predict $\hat y_t = \phi_t^\top W$.
\item Compute the residual $r_t = y_t - \hat y_t$, and update the fast and slow exponential moving averages $\sigma^2_{\mathrm{fast}}$ and $\sigma^2_{\mathrm{slow}}$ of $r_t^2$.
\item Form the excess error $E_t = \max(0, \sigma^2_{\mathrm{fast}} - \sigma^2_{\mathrm{slow}})$. This quantity drives both the conflict estimator and the drift-energy estimator.
\item Form the conflict estimate $\widehat C_t = \max\{C^\Sigma_{\mathrm{ref}}, C^\Sigma_{\mathrm{resid}}, C^\Sigma_{\mathrm{floor}}\}$ and the drift-energy estimate $\widehat\nu^2_t$ of Theorem~\ref{thm:window}. Then smooth both quantities as $\bar C \leftarrow (1-\eta_C)\bar C + \eta_C\widehat C_t$ and $\bar\nu^2 \leftarrow (1-\eta_\nu)\bar\nu^2 + \eta_\nu\widehat\nu^2_t$.
\item Update the anchor strength as $\rho \leftarrow \mathrm{clip}(\bar\mu\widehat V^\Sigma/\bar C,\ \rho_{\min}(E_t),\ \rho_{\max})$.
\item Update the forgetting rate as $\omega \leftarrow \mathrm{clip}((4\bar\nu^2/\hat c')^{1/3},\ \omega_{\min},\ \omega_{\max})$, and set $\alpha \leftarrow 1-\omega$.
\item Update the sufficient statistics as $A \leftarrow \alpha A + \phi_t\phi_t^\top$, $B \leftarrow \alpha B + \phi_t y_t$, and $n \leftarrow \alpha n + 1$.
\item Solve the anchored system $W \leftarrow (A + n\rho I)^{-1}(B + n\rho G)$.
\item Every $k$ steps (after the first), compute $(\mu,U) \leftarrow \mathrm{eigh}(\bar A)$ and refresh the anchor as $G \leftarrow U[\mathrm{diag}(\beta_i)U^\top W + \mathrm{diag}(1-\beta_i)U^\top G]$; then update the slow reference as $G_{\mathrm{ref}} \leftarrow (1-\eta_{\mathrm{ref}})G_{\mathrm{ref}} + \eta_{\mathrm{ref}} W$ with $\eta_{\mathrm{ref}} = k/2000$.
\end{enumerate}
}
\vspace{2pt}\hrule
\end{minipage}
\end{algfloat}

\begin{table}[t]
\centering
\footnotesize
\setlength{\tabcolsep}{4pt}
\renewcommand{\arraystretch}{0.95}
\caption{Notation and the fixed runtime constants, shared by every configuration (Online Appendix~\ref{app:hyper}); every other quantity adapts online.}
\label{tab:notation}
\begin{tabular}{@{}l@{\ }l@{\qquad}l@{\ }l@{}}
\toprule
\multicolumn{2}{@{}l}{\emph{notation}} & \multicolumn{2}{l}{\emph{fixed constants}}\\
\midrule
$D$ & input dimension & $k$ & refresh cadence $10$ \\
$T_w,T$ & warm-up, stream length & $h_f$ & fast residual EMA $30$ \\
$A,B,n$ & sufficient statistics & $h_{\mathrm{slow}}$ & slow residual EMA $20{,}000$ \\
$\bar A_t,A_t$ & mean, running accumulator & $2000$ & $G_{\mathrm{ref}}$ EMA horizon \\
$\mu_i,U$ & excitation eigenpairs & $\rho_{\min}$ & strength clip floor $0.02$ \\
$s_i$ & pull weight $\rho/(\mu_i{+}\rho)$ & $\rho_{\max}$ & strength clip cap $1.0$ \\
$\rho,\omega$ & anchor strength, forgetting rate & $\omega_{\min},\omega_{\max}$ & window range $[5{\cdot}10^{-4},\,0.05]$ \\
$G,G_{\mathrm{ref}}$ & anchor, slow reference & $K_{\mathrm{cal}}$ & burn-in blocks $10$ \\
$W,\widehat w_t$ & anchored solve, estimate & $\eta_C,\eta_\nu$ & conflict, drift EMA $0.15,\,0.2$ \\
$d_{\mathrm{act}},d_{\mathrm{eff}}$ & active, effective dimension & $\rho_{\mathrm{ridge}}$ & data-solve ridge $10^{-6}$ \\
$\bar\mu$ & mean excitation $\mathrm{tr}(\bar A_t)/D$ & $\varepsilon_{\mathrm{act}}$ & active-direction threshold $10^{-8}$ \\
$\widehat C,\widehat V^\Sigma$ & conflict, variance (prediction units) & $\rho_{\mathrm{lo}}$ & open-window strength floor $10^{-5}$ \\
\bottomrule
\end{tabular}
\end{table}

The same loop defines the solver family used below.
\emph{SPAR-Pull} is Algorithm~\ref{alg:spar-pull} (two estimators, one scalar $\rho$, exact $O(D^3)$ solve).
\emph{SPAR-Pull-D} keeps a per-direction strength $\rho_i = \sigma^2/(n\delta_i^2)$ and caches the refresh eigenbasis, giving $O(D^2)$ per step with an $O(D^3)$ refresh every $k$ steps.
\emph{SPAR-Pull-S} is the scalar-pull simplification on the data--anchor segment, used to isolate the regret guarantee.
\emph{SPAR-EG} replaces the single solve by a Hedge mixture over five batch-solved leaky windows plus the frozen anchor; \emph{SPAR-EG$^+$} adds the SPAR-Pull trajectory as a sixth expert.

\section{Theoretical Analysis}
\label{sec:theory}

The section is organized around the refresh recursion.
Lemma~\ref{lem:zerolag} establishes the continuous-refresh limit.
Theorem~\ref{thm:cadence} gives the finite-cadence equilibrium, and Theorem~\ref{thm:window} minimizes the resulting risk to obtain the window law.
Propositions~\ref{prop:conflict}--\ref{prop:windup} give the anchor calibration, the minimax and false-alarm guarantees, the regret wrappers, and the conditioning bound.
Theorem~\ref{thm:rmt} is the finite-dimensional spectral correction, and Theorem~\ref{thm:closedloop} closes the loop with the coupled contraction.
Theorems~\ref{thm:cadence} and~\ref{thm:window} are analyzed in the mean-field (population-design) model in which the random accumulator $\bar A_t$ is replaced by its expectation $\Sigma$; this is the model that yields the closed-form window target, and it is exact in the proportional limit $\zeta = D\omega \to 0$.
Theorem~\ref{thm:rmt} supplies the finite-$D$ random-design correction at $\zeta > 0$; Appendix~\ref{app:proofs} states the two models explicitly and quantifies their gap ($\zeta \le 0.13$ on the pre-drift grid, up to $6.4$ during drift-phase adaptation).

\begin{lemma}[Zero-lag refresh equilibrium]
\label{lem:zerolag}
Assume drift at constant rate $\nu$ per step along a direction with excitation $\mu$, forgetting rate $\omega$, and continuous anchor refresh.
(i) The accumulator's implied data solution lags the target by exactly $\nu/\omega$.
(ii) At the refresh equilibrium $G = W$ the anchored solution's lag equals the pure-leaky lag, independent of $s$; the anchor is inert for mean tracking.
(iii) At refresh cadence $k$ an extra lag $\Delta_k(s)$ of Theorem~\ref{thm:cadence} appears, whose prediction cost is weighted by $\mu_i$ and therefore concentrates in weakly excited directions.
\end{lemma}

\begin{theorem}[Finite-cadence refresh equilibrium]
\label{thm:cadence}
Let $a = s(2-s) = 1 - (1-s)^2 \in (0,1)$ for $s \in (0,1)$, and let $r_k = \alpha^k \in [0,1)$ be the data-path noise correlation across one refresh cycle.
At cadence $k$ with data-path lag $L$ (Lemma~\ref{lem:zerolag}) and target velocity $\nu$ per step, the moving-frame anchor offset follows the affine recursion
\[
  g_{m+1} \;=\; a\,g_m \;-\; (1-s)^2\,L \;-\; a\,\nu k,
  \qquad a := s(2-s) = 1-(1-s)^2.
\]
The recursion contracts geometrically at rate $a \in (0,1)$ per cycle to the fixed point $g^* = -L - \nu k\,a/(1-a)$, and the cycle-averaged solution lags the pure-leaky path by $\Delta_k(s) = s\nu k\,(a/(1-a) + \tfrac{1}{2})$.
The lag is linear in the cadence and vanishes as $s \to 0$.
The steady-state solution variance is $v\,\Psi(s,r_k)$ with $\Psi(s,r_k) := 1 - 2s(1+s-s^2)(1-r_k)/((1+a)(1-a r_k)) \in [(1-s)^2,\,1]$ for all $s \in (0,1)$, $r_k \in [0,1)$: the variance reduction of Proposition~\ref{prop:conflict} survives at every finite cadence, vanishing at the inert continuous-refresh limit $r_k \to 1$ of Lemma~\ref{lem:zerolag}(ii) and largest at $r_k \to 0$, so the cadence trades lag against variance reduction (Appendix~\ref{app:proofs}).
\end{theorem}

\begin{theorem}[Finite-cadence optimal forgetting window]
\label{thm:window}
Let the target drift at constant rate $\nu_i$ in direction $i$ with prediction-units drift power $\nu^2_\Sigma = \sum_i \mu_i \nu_i^2$, let the refresh cadence be $k$, and set $r_k = (1-\omega)^k$ and $a_i = s_i(2-s_i)$.
At the refresh equilibrium of Theorem~\ref{thm:cadence}, the cycle-averaged prediction risk at pull profile $s$ is
\[
\begin{aligned}
  R_k(\omega,s) &= \sigma^2\,\frac{\omega}{2-\omega}\sum_{i=1}^D \Psi(s_i,r_k) \\
  &\quad + \sum_{i=1}^D \mu_i\Bigl(\frac{\nu_i}{\omega} + \Delta_k(s_i)\Bigr)^2
  + O(k^2\nu^2 s^2),
\end{aligned}
\]
with $\Psi(s,r) = 1 - \frac{2s(1+s-s^2)(1-r)}{(1+a)(1-ar)}$ and $\Delta_k(s) = s\nu k\bigl(\frac{a}{1-a}+\frac12\bigr)$.
The runtime cadence need not satisfy $1 \ll k\omega$, since the grid defaults have $k\omega \in [0.005, 0.5]$; two monotone facts justify the reduced target.
\emph{(i) Variance lower bound.}
Since $(1-s_i)^2 \le \Psi(s_i,r_k) \le 1$ for every $r_k \in [0,1)$ (Theorem~\ref{thm:cadence}), the surrogate $c'\,\omega/(2-\omega)+\nu^2_\Sigma/\omega^2$ with $c' = \sigma^2 d_{\mathrm{eff}}(\rho)$ lower-bounds the exact variance term at every cadence; Jensen's inequality gives $D(1-\bar s)^2 \le d_{\mathrm{eff}}(\rho)$, so the runtime plug-in is itself a lower bound (the gap is exactly the pull variance, $\sum_i(1-s_i)^2 - D(1-\bar s)^2 = D\operatorname{Var}(s)$, formalized in \texttt{Core.lean}),
\[
  \hat c'_t \;=\; \sigma^2 D(1-\bar s_t)^2 \;\le\; \sigma^2\,d_{\mathrm{eff}}(\rho) \;\le\; \sigma^2 \sum_{i=1}^D \Psi(s_i,r_k).
\]
Its minimizer $\omega^* = (4\nu^2_\Sigma/c')^{1/3}$ is at least the exact-optimal window (a smaller coefficient returns a larger $\omega$, the over-forgetting direction), which is the cheap side of the safety asymmetry of Proposition~\ref{prop:minimax}.
\emph{(ii) Finite-cadence lag residual.}
With $b(s) = s(2-s)/(1-s)^2$ and $k\omega \le 1/2$, the residual of Theorem~\ref{thm:cadence} obeys
\[
  \Delta_k(s_i) = s_i\,\nu_i\,k\bigl(b(s_i)+\tfrac12\bigr) \;\le\; \tfrac12\,s_i\bigl(b(s_i)+\tfrac12\bigr)\,\frac{\nu_i}{\omega},
\]
a fixed fraction of the leading lag that vanishes as $s_i \to 0$, whose prediction cost $\mu_i\,\Delta_k(s_i)^2$ is $\mu_i$-weighted and concentrated in weakly excited directions.
Over the runtime range $k\omega \in [0.005, 0.5]$ the surrogate's cubic minimizer stays within $2.1\%$ of the exact cycle-averaged optimum for $s_i \le 0.2$ (within $5.1\%$ for $s_i \le 0.3$, the largest gaps at the window floor; Appendix~\ref{app:proof-window}).
In the quasi-static limit $1 \ll k\omega \ll 1/s$ (so $s \ll 1/(k\omega) \ll 1$) both corrections vanish identically, with $\Psi(s_i,r_k) = (1-s_i)^2(1+O(s_i^2+r_k))$ and $\Delta_k(s_i) = o(\nu_i/\omega)$, and the reduced form is exact:
\[
  R_k(\omega,s) = c'\,\frac{\omega}{2-\omega} + \frac{\nu^2_\Sigma}{\omega^2} + o(1),
  \qquad c' = \sigma^2\,d_{\mathrm{eff}}(\rho),
  \qquad d_{\mathrm{eff}}(\rho) := \sum_{i=1}^D (1-s_i)^2,
\]
which is strictly convex with unique minimizer $\omega^* = (4\nu^2_\Sigma/c')^{1/3}$ to leading order (for $\nu_\Sigma=0$ the optimum is the boundary $\omega^*=0$).
The runtime estimator uses the mean-field form $D(1-\bar s)^2$ (exact on isotropic spectra, conservative otherwise); it recovers the drift energy from the two-scale residual excess, $\widehat\nu^2_\Sigma = \max\bigl(0,\,(\widehat E_t - \hat c'\,\omega/(2-\omega))\,\omega^2\bigr)$ with $\widehat E_t = \widehat\sigma^2_{\mathrm{fast}} - \widehat\sigma^2_{\mathrm{slow}}$, and under stationarity $\widehat E_t \to 0$ in expectation, pinning $\omega$ at $\omega_{\min}$ (Appendix~\ref{app:proofs}).
\end{theorem}

\begin{corollary}[Mixed-drift master law]
\label{cor:mixed}
For target increments $\Delta w^*_t = \nu + \vartheta_t$ with coherent velocity $\nu$ and martingale-difference innovation power $q_\Sigma = \E[\vartheta_t^\top \Sigma\, \vartheta_t]$, the exact tracking lag is $\mathcal{L}_{\mathrm{lag}}(\omega) = q_\Sigma/(\omega(2{-}\omega)) + \nu^2_\Sigma/\omega^2$, giving $\omega^* = (4\nu^2_\Sigma/c')^{1/3}$ in the coherent regime ($q_\Sigma = 0$; Theorem~\ref{thm:window}) and the classical $\omega^* = (q_\Sigma/c')^{1/2}$ random-walk law of \citet{widrow1976}; the runtime estimator inverts the total lag, converging to the appropriate regime automatically (Appendix~\ref{app:proofs}).
\end{corollary}

\begin{proposition}[Anchored risk and the optimal anchor strength]
\label{prop:conflict}
Let $v$ be the data-path variance and $\delta^2$ the anchor conflict in a given direction.
(i) The anchored risk profile is $R(s) = (1-s)^2 v + s^2 \delta^2$; with a correct anchor ($\delta^2 = 0$) it is strictly below $v$ for every $s > 0$, with the largest gain in weakly excited directions.
(ii) $R(s) = v\delta^2/(v+\delta^2) + (v+\delta^2)\,(s - v/(v+\delta^2))^2$, so the optimal pull is $s^* = v/(v+\delta^2)$ and the corresponding per-direction strength is $\rho^*_i = \sigma^2/(n\,\delta^2_i)$, where $\delta^2_i$ is the expected anchor conflict.
The optimal anchor strength thus decreases in measured conflict and in $n$.
\end{proposition}

\begin{proposition}[Minimax safety of conflict overestimation]
\label{prop:minimax}
At a fixed conflict budget, the minimax and maximin values coincide (the parallel sum of the variance and the budget).
Overestimating conflict is safe.
Under a calibrated overestimate, the excess of the anchored risk over the no-anchor risk is at most $(v+\delta^2)(s-s^*)^2$. Underestimating conflict costs quadratically in the same expression; the twice-the-optimum boundary is the James--Stein domination condition \citep{james1961}.
\end{proposition}

\begin{proposition}[False-alarm calibration of the sliding two-scale EMA]
\label{prop:falsealarm}
Let $z_t$ be a centered sub-exponential martingale-difference sequence with variance proxy $\nu_z^2$ and scale $b_z$, and let the fast two-scale excess be its causal moving average $Z_t = \sum_{k\ge 0} w_k z_{t-k}$ with geometric weights $w_k = \eta_f(1-\eta_f)^k$, so $\sum_k w_k^2 = \eta_f/(2-\eta_f) = 1/m_{\mathrm{eff}}$ and $\max_k w_k = \eta_f$.
For every horizon $T$ and $\delta \in (0,1)$, with
\[
\tau(\delta) \;=\; \eta_f b_z L + \sqrt{\eta_f^2 b_z^2 L^2 + \frac{2\nu_z^2}{m_{\mathrm{eff}}}\,L},
\qquad L = \log(T/\delta),
\]
the sliding trigger satisfies the time-uniform false-alarm bound $\Pr[\exists\, t \le T:\ Z_t > \tau(\delta)] \le \delta$; in the Gaussian corner ($b_z \to 0$) it reduces to $\tau = \sqrt{2\log(T/\delta)/m_{\mathrm{eff}}}$.
The variance-floor subtraction and conflict floor induce strictly positive effective thresholds $\tau_\omega = \hat c' \omega_{\min}/(2-\omega_{\min})$ and $\tau_\rho = C_{\mathrm{floor}}$, so applying the bound at $\tau = \min(\tau_\omega,\tau_\rho) > 0$ pins $\omega_t$ at $\omega_{\min}$ with probability at least $1-\delta$ under stationarity.
This calibrates the implemented EMA directly, with measured rates in Section~\ref{sec:bridges} and proof in Appendix~\ref{app:proofs}.
\end{proposition}

\begin{proposition}[No-regret wrappers]
\label{prop:ocowrap}
(i) The cumulative loss is convex in the pull profile $s \in [0,1]^D$ (affine in $s$ by Equation~\eqref{eq:pull} in the instantaneous eigenbasis $U_t$ of $\bar A_t$, with the minimum-ridge data solve $\widehat w_t = (\bar A_t + \rho_{\mathrm{ridge}}I)^{-1}\bar b_t$, then squared), so projected gradient descent on the profile attains regret at most $G_B\sqrt{2DT}$ against every fixed profile in the box, including the data path ($s = \mathbf 0$), the anchor path ($s = \mathbf 1$), and, when the accumulator is quasi-stationary, every fixed scalar strength $\rho$, for arbitrary adaptive bounded losses \citep{zinkevich2003}.
A fixed profile $s$ is basis-relative: when $U_t$ rotates it corresponds to the time-varying physical operator $U_t\operatorname{diag}(s)U_t^\top$, so the comparator is the realized profile sequence rather than a fixed linear operator in feature space (Appendix~\ref{app:proof-composition}). The class contains a basis-invariant sub-class: a constant profile $s \equiv c$ realizes $cI$ in every orthonormal basis ($U\operatorname{diag}(c\mathbf 1)U^\top = cI$; \texttt{conj\_const\_diagonal} in \texttt{Core.lean}), so the data path ($c=0$), the anchor path ($c=1$), and every fixed isotropic ridge are basis-invariant comparators, while a general fixed $s$ competes through its realized physical sequence.
The $s = \mathbf 1$ comparator is the learner-dragged anchor $G_t$ (external regret on the interpolation segment); the frozen $G_0$ enters only in (ii), where the counterfactual guarantee against it lives.
(ii) Exponentially weighted mixture prediction over $K$ expert paths (batch-solved leaky windows plus the frozen anchor) satisfies $\sum_t \ell_t(\mathrm{mix}) \le (\min_i L_{i,T} + \ln K/\eta)/(1 - \eta M)$ for arbitrary bounded adaptive losses; the comparator class contains every benchmarked window and the anchor, a best-of-both-worlds statement (Appendix~\ref{app:proofs}).
\end{proposition}

\begin{proposition}[Information-form conditioning bound]
\label{prop:windup}
At any window and for any data, $\kappa(A + n\rho I) \le 1 + \mu_{\max}/\rho$, independent of the drift.
The anchored solve never iterates an inverse, so it needs no implementation care at short windows; Section~\ref{sec:experiments} measures what happens when the standard covariance recursion does.
\end{proposition}
\begin{proof}
The eigenvalues of $A + n\rho I$ lie in $[\,n\rho,\ n(\mu_{\max} + \rho)\,]$, whose ratio is $1 + \mu_{\max}/\rho$ (the pull-weight matrix identity, the anchored resolvent, the floor nonsingularity, and the eigenbasis conditioning ratio are formalized as \texttt{pull\_matrix\_sum\_one}, \texttt{diagonal\_resolvent}, \texttt{anchored\_det\_pos}, and \texttt{anchored\_cond\_bound} in \texttt{Core.lean}).
\end{proof}

\begin{theorem}[Spectral retention and the streaming risk]
\label{thm:rmt}
Let the feature vectors have independent sub-Gaussian entries with population spectral measure $H$ supported on $[0,\tau_{\max}]$ whose nonzero part lies in $[\tau_{\min}^+,\tau_{\max}]$, $\tau_{\min}^+>0$ (a possibly rank-deficient population covariance), and let $\zeta = D \omega$ be held fixed as $D \to\infty$ (proportional regime).
(i) The average data retention $q = 1 - \bar s$ is the unique solution in $(0,1)$ of
\[
\begin{aligned}
q \;&=\; \int \frac{\tau}{\tau + \rho_{\mathrm{eff}}}\, dH(\tau), \\
\rho_{\mathrm{eff}} \;&=\; \rho\,\frac{e^{\zeta q} - 1}{\zeta q},
\end{aligned}
\]
and the anchored one-step excess prediction risk converges to $\sigma^2\, \Upsilon/(1 - \Upsilon)$ with
$\Upsilon = \frac{1}{q}\int \bigl(\frac{\tau}{\tau + \rho_{\mathrm{eff}}}\bigr)^{2} dH(\tau)\,\bigl(1 - \frac{1 - e^{-\zeta q}}{\zeta q}\bigr)$.
(ii) The pull sensitivities are closed form: $\rho\, \partial \bar s / \partial \rho = \mathcal{I}\,/\,(1 + \zeta\,\phi(\zeta q)\,\mathcal{I})$, with $\mathcal{I} = \int \tilde s (1 - \tilde s)\, dH(\tau)$, $\tilde s(\tau) = \rho_{\mathrm{eff}}/(\tau + \rho_{\mathrm{eff}})$, and $\phi(u) = \tfrac{1}{1 - e^{-u}} - \tfrac{1}{u}$.
(iii) Under an anchor conflict $\delta = G - w^*$ with direction-weighted conflict measure $H_\delta$ (atom $\delta_i^2$ at direction $\tau_i$), the full excess prediction risk converges to
\[
R_{\mathrm{pred}} \;\longrightarrow\; \frac{1}{1 - \Upsilon}\left[\sigma^2 \Upsilon + \rho_{\mathrm{eff}}^2 \int \frac{\tau}{(\tau + \rho_{\mathrm{eff}})^2}\, dH_\delta(\tau)\right].
\]
The oracle anchor recovers Part~(i), and the classical limit $\zeta \to 0$ recovers Proposition~\ref{prop:conflict}'s per-direction conflict bias $\mu_i(\rho/(\mu_i+\rho))^2\delta_i^2$. Part~(i) is therefore the variance component of the risk, while Part~(iii) is the full risk.
\end{theorem}
The convergence step is the separable-covariance deterministic equivalent \citep{silverstein1995,rubio2011,hachem2007}; the nearest prior analysis, the large-system exponentially windowed RLS of \citet{xiao2005}, has no anchor, pulls, refresh, or online calibration, and the ridge-risk line \citep{hastie2022} treats unweighted batches.
The finite-window spreading is a single renormalization $\rho \mapsto \rho_{\mathrm{eff}} \ge \rho$: the formula matches simulation within $4.0\%$ across the $45$-point fixed-point grid (including the rank-deficient spectrum) and the measured operating pairs within $13\%$.
It is a correction to the analysis, not a term fed back into the runtime rule: Algorithm~\ref{alg:spar-pull} keeps the mean-field plug-in $\hat c'_t$, and Theorem~\ref{thm:rmt} quantifies what that forgoes as $\zeta$ grows.
When $\Sigma$ is rank-deficient, exact null directions carry zero out-of-sample risk, so the deterministic equivalent restricts exactly to $\operatorname{range}(\Sigma)$, with the aspect ratio rescaled by the rank fraction $\zeta^+ = p\zeta$ (Corollary~\ref{cor:rmt-active}).
Part~(ii) supplies the analytic coupling product used in Theorem~\ref{thm:closedloop}: in logarithmic coordinates the product is at most $\tfrac{4}{3(2-\omega)}\,\bar s < 1$ for all $\omega \in (0, 2/3)$ (Online Appendix~\ref{app:protocol}).

The per-direction variant inherits the Davis--Kahan eigenvector fragility \citep{davis1970,wedin1972}: it degrades on flat and rotating spectra (Section~\ref{sec:limitations}), while the scalar method is invariant to basis rotation (Proposition~\ref{prop:invariance}, Appendix~\ref{app:proofs}).
Variable forgetting factor RLS \citep{fortescue1981,paleologu2008} adapts its factor from the squared prequential residual; in covariance form it suffers the same short-window instability (Section~\ref{sec:experiments}).

\begin{theorem}[Closed-loop stability of the two estimators]
\label{thm:closedloop}
Let the window and strength errors satisfy the coupled recursions of Appendix~\ref{app:proofs} (estimator gains $\eta_\nu, \eta_C$; target sensitivities $L_{\omega\rho}, L_{\rho\omega}$; perturbations bounded by $\xi$).
The recursions model the quasi-static two-time-scale regime: the smoothing horizons $1/\eta_C, 1/\eta_\nu$ are slow relative to the fast EMA horizon $h_f$, and, in the drift-active interior where the window is open ($1/\omega \lesssim k/\eta_\nu$), to the accumulator mixing time $1/\omega$; finite-$k$ lag then enters as an additive perturbation absorbed into $\xi_\omega,\xi_\rho$, so the contraction and tracking radius are uniform in its bound. Outside that interior the detector floor pins $\omega_t = \omega_{\min}$ (Theorem~\ref{thm:window}), the pinned specialization of part~(ii).
(i) If $L_{\omega\rho} L_{\rho\omega} < 1$, the pair contracts geometrically in a weighted max-norm, with explicit weight and rate below one, to within $\xi/(1-\gamma_{\mathrm{cl}})$ of the coupled fixed point.
(ii) When either quantity is pinned (the drift-active lower bound on $\rho$, or the window floor or cap on $\omega$), the system is triangular and the free quantity's contraction governs alone.
(iii) The stationary fluctuation obeys $\limsup_m \E[V(m)^2] \le q_*/(1 - \gamma_{\mathrm{cl}})^2$ for the weighted max-norm $V = \max\{|e_\omega|, r|e_\rho|\}$ and innovation second moment $q_* \ge \E\|\xi(m)\|_r^2$ (Proposition~\ref{prop:fluct}); the cross-quantity covariance is bounded by the same norm, $|\E[e_\omega e_\rho]| \le \E[V^2]/r \le q_*/(r(1-\gamma_{\mathrm{cl}})^2)$, so the full stationary second-moment matrix is controlled entrywise.
\end{theorem}
The proof assembles Theorem~\ref{thm:window}'s exact window contraction with the plug-in rule of Proposition~\ref{prop:conflict}. The coupled contraction is verified in \texttt{Coupled.lean}; the measured coupling product is below one in all 20 diagnostic runs, with one quantity pinned in three of four configurations (Online Appendix~\ref{app:protocol}).

\section{Experiments}
\label{sec:experiments}

\subsection{The Window Law}
\emph{Protocol.}
We evaluate 31 configurations at $D = 128$ and $T = 10^4$ over 5 seeds each: a 27-configuration full-factorial grid (three spectra, three drift geometries, three prior qualities) plus four stress tests (loss of persistent excitation, oscillating drift, a stationary control, and a heavy-tailed variant).
One fixed control parametrization is used for every SPAR configuration (Online Appendix~\ref{app:hyper}); baselines use the best single hyperparameter in their family.
Every baseline family runs through a shared verification step that confirms exact reductions to RLS at the degenerate parameter values.

\emph{Pre-drift floor.}
Before drift, the calibrated window sits at its lower bound and the error is the noise floor: SPAR-Pull holds $1.01$--$1.14\times\sigma^2$ across the three spectra ($0.0106$ grand mean vs.\ $\sigma^2=0.01$) and varies by $4.1\%$ across oracle, warm, and adversarial priors ($0.301$, $0.305$, $0.313$).

\emph{Window response.}
On the abrupt-jump stream the window opens after the jump and the strength falls; Online Appendix Figure~\ref{fig:cliffdials} traces both.
A transient sweep (Online Appendix Table~\ref{tab:transient}) shows that the adaptive strength stays within $1$--$5\%$ of the best fixed strength at cadence $10$, while a large fixed anchor costs $87\%$ overall.
Online Appendix Figure~\ref{fig:cliffapp} sweeps the window and locates the sharp degradation at short windows, confirming that the calibrated method stays on the stable side.

\emph{Finite-$D$ correction.}
The deterministic-equivalent formula matches the Monte Carlo stationary risk within $4.0\%$ across all $45$ operating points of the grid, including the rank-deficient spectrum (Appendix~\ref{app:rmt-validation}), within $13\%$ at the method's own operating points, and the conflict increment within $10.0\%$.
Rank-deficient spectra reduce cleanly to the active subspace (Corollary~\ref{cor:rmt-active}).

\subsection{Fixed-protocol Comparison}
Table~\ref{tab:grid} reports the fixed-protocol comparison.
Two oracle-free families lead the mean: the slow-horizon estimated-noise Kalman ($0.215$) and nominal-length VFF ($0.267$).
Both are static-target specialists; SPAR-Pull wins all nine covariance-rotation configurations ($0.011$ vs.\ $0.023$ and $0.027$).
On moving targets the unanchored information-form leaky method wins $12$ of the $18$ configurations (each baseline family at its per-configuration best), but the outcome is spectrum-structured rather than uniform: the $12$ wins are all small-margin, six isotropic by $2.1\%$ (paired mean difference $-0.026$) and six rank-deficient by $0.8\%$, statistically a tie (paired mean difference $-0.0015$, max magnitude $0.0024$), while SPAR-Pull wins all six ill-conditioned (decay3) configurations by a factor of $2.0$ ($0.166$ vs.\ $0.331$); on the moving-target mean SPAR's best variant is $9.3\%$ below the leaky method ($0.439$ vs.\ $0.484$).
The leading baseline's margin rests on hindsight: its process-noise level is tuned post hoc on the evaluation grid, and selecting it causally from the warm-up alone instead degenerates the estimated-noise Kalman to mean MSE $1.69$ against the post-hoc $0.215$ (Online Appendix~\ref{app:protocol}).
The calibrated SPAR methods attain $0.297$--$0.306$ mean MSE with $1.22$ worst-case, and no SPAR method diverges, while the unsymmetrized covariance-form recursion diverges at short windows ($135/135$ main-grid and $20/20$ stress-suite runs).
Per-step symmetrization stabilizes the covariance form on full-rank spectra \citep{bierman1977,sayed2008} but not on rank-deficient spectra, where the dead directions inflate deterministically; the anchored information form avoids the issue entirely (Proposition~\ref{prop:windup}).

\begin{table}[t]
\centering
\footnotesize
\setlength{\tabcolsep}{4pt}
\renewcommand{\arraystretch}{0.92}
\caption{Factorial grid (27 main configurations $\times$ 5 seeds): mean, geometric-mean, and worst overall MSE (worst single run of 135; the geometric mean is scale-invariant across the three spectra).
Baselines use their family's best single hyperparameter, selected post hoc on the evaluation grid; the causal Kalman instead selects its $q$ from the warm-up alone.
The true-noise Kalman is an oracle comparator (best single $q$, $0.204/0.86$); the WiSE-FT envelope sweeps $\alpha$ per configuration.
VFF-RLS covariance reports the nominal-length calibration's best setting ($N_0 = 100$); the single-step calibration $\xi = \sigma_0^2$ diverged in all 135 runs (Online Appendix~\ref{app:protocol}).
Standard errors of the mean over the $135$ runs are $0.026$--$0.050$, so aggregate differences below about $0.05$ are not resolved by the mean alone; the paired tests use configuration-level pairing.}
\label{tab:grid}
\setlength{\tabcolsep}{3pt}
\begin{tabular}{lccc}
\toprule
method & mean MSE & geo.\ mean & worst MSE \\
\midrule
\multicolumn{3}{l}{\emph{baselines (single hyperparameter, tuned post hoc on the evaluation grid)}}\\
leaky RLS ($\lambda = 0.997$) & $0.555$ & $0.233$ & $2.12$ \\
leaky RLS, information form ($\lambda = 0.99$) & $0.342$ & $0.188$ & $1.19$ \\
SGD ($\mathrm{lr} = 0.01$) & $0.377$ & $0.147$ & $1.29$ \\
Adam \citep{kingma2015adam} ($\mathrm{lr} = 0.01$) & $0.503$ & $0.291$ & $1.92$ \\
directional forgetting & $1.81$ & $0.567$ & $7.54$ \\
AROW & $2.74$ & $0.448$ & $10.55$ \\
Kalman, $R$ estimated online (100-step horizon) & $1.09$ & $0.265$ & $5.36$ \\
Kalman, $R$ estimated online (20{,}000-step horizon) & $0.215$ & $0.090$ & $0.93$ \\
VFF-RLS, covariance form ($N_0 = 100$) \citep{fortescue1981} & $0.267$ & $0.125$ & $0.98$ \\
\multicolumn{3}{l}{\emph{causal baseline (no test-grid tuning)}}\\
Kalman, $R$ estimated online (causal $q$ from warm-up) & $1.69$ & $0.266$ & $9.42$ \\
\multicolumn{3}{l}{\emph{regret-guaranteed methods (one fixed configuration)}}\\
OGD, scalar pull (segment) & $0.585$ & $0.226$ & $2.16$ \\
OGD, $D$-dim profile (Proposition~\ref{prop:ocowrap}(i)) & $0.707$ & $0.324$ & $2.61$ \\
\midrule
\multicolumn{3}{l}{\emph{oracle comparator}}\\
Kalman, best single $q$, true $\sigma^2$ & $\mathbf{0.204}$ & $\mathbf{0.089}$ & $\mathbf{0.86}$ \\
\midrule
\multicolumn{3}{l}{\emph{ours (closed-form streaming, one fixed configuration)}}\\
SPAR-EG (Proposition~\ref{prop:ocowrap}(ii)) & $0.358$ & $0.144$ & $1.27$ \\
SPAR-EG$^+$ (pool incl.\ SPAR-Pull) & $0.297$ & $0.099$ & $1.22$ \\
SPAR-Pull & $0.306$ & $0.100$ & $1.22$ \\
\bottomrule
\end{tabular}
\end{table}

The stability gap between the covariance and information forms stems from how each handles unexcited directions.
In covariance form, reducing $\lambda$ from $0.997$ (nominal window $333$) to $0.99$ (window $100$) raises the 27-configuration mean MSE from $0.555$ to $182$ and then to $1.2 \times 10^6$ (diverging in all $135$ main-grid and $20$ stress-suite runs).
Because any component invisible to the rank-one downdate is multiplied by $(1{-}\omega)^{-1}$ at each step, errors grow as $(1{-}\omega)^{-t}$.
On full-rank spectra, this unstable component arises from asymmetric roundoff in the iterated inverse, and per-step symmetrization restores stability at every window tested ($0.61$ at $\lambda = 0.99$, $D = 128$). On rank-deficient spectra it is the dead directions themselves, which inflate deterministically even under symmetrization (stable on all 90 full-rank runs, divergent on 42 of 45 rank-deficient). The anchored information form never iterates an inverse and is immune to both. Neither directional forgetting nor any SPAR method diverged in any configuration (Online Appendix Figure~\ref{fig:cliffapp}).

Exponentiated-gradient weights over $K = 5$ expert paths (batch-solved leaky windows plus the frozen anchor), with the anytime rate $\varsigma_t = \sqrt{\ln K / S_t}$, attain $0.358/1.27$ with no tuning, within $17\%$ of the plug-in rule; on oscillating streams the mixture is within $2.4\%$ of the best single window ($6.36$ vs.\ $6.21$).
Adding the SPAR-Pull trajectory itself as a sixth expert (SPAR-EG$^+$) attains $0.297$ mean MSE ($1.22$ worst-case), $2.9\%$ below the plug-in method, so the Proposition~\ref{prop:ocowrap}(ii) guarantee holds relative to a pool containing the plug-in method. The gain appears in 11 of 27 configurations (paired $t$ $p = 0.032$; Wilcoxon $p = 0.61$).

\subsection{Ablation and Cost}
\label{sec:bridges}
On four representative configurations, removing the window estimator, the two-scale detector, or the refresh costs $+189$--$190\%$ overall MSE (the first two pin $\omega_t = \omega_{\min}$), and $\gamma = 0.5$ costs $+560\%$; freezing the strength rule \emph{lowers} mean MSE by $2.2\%$.
Online Appendix Table~\ref{tab:ablation} localizes prior immunity: freezing the spectral anchor refresh costs $2.2\times$ on warm-prior configurations and $7.4\times$ on adversarial ones (early MSE $47.0$ vs.\ $16.4$), while freezing $\rho$ leaves both within $2.6\%$.
The strength rule's contribution is therefore the onset guard of Section~\ref{sec:experiments}, not a continuous tracking gain.
SPAR-Pull solves step (viii) exactly each step ($O(D^3)$); SPAR-Pull-D caches the refresh eigenbasis and runs at $O(D^2)$ per step with $O(D^3)$ every $k$ refreshes ($51.3$ vs.\ $225$--$363$ ms at $D = 2048$; Online Appendix Table~\ref{tab:app-ops}).

An abrupt-jump strength sweep locates the value of the rule (Online Appendix Table~\ref{tab:transient}): a large fixed anchor, the choice a good warm prior invites, multiplies pre-drift MSE by $4.2$ relative to a small fixed anchor and costs $87\%$ overall at cadence 10, while the adaptive rule stays within $1$--$5\%$ of the best fixed strength.
The 50-step post-jump window is dominated by the jump itself and is flat across the adaptive and small-anchor policies, so $\rho_t$ acts as a guard against an oversized anchor, not as a post-jump accelerator.

The false-escalation rate of Proposition~\ref{prop:falsealarm} is at or below nominal except at $\delta = 0.01$ on ill-conditioned spectra, as the sub-exponential tail predicts.
The OGD regret slopes ($-0.65$ isotropic, $-1.08$ ill-conditioned) lie within the $O(T^{-1/2})$ rate, with pre-drift dominance and prior immunity in Online Appendix Figures~\ref{fig:regret} and~\ref{fig:prior} (Online Appendix~\ref{app:protocol}).

\subsection{Robustness Outside the Model Class}
Three generators outside the linear-Gaussian, constant-rate class: nonlinear targets ($y = \tanh(x^\top w^*) + \varepsilon$), covariate shift (feature scale doubles mid-stream), and a noise-regime change ($\sigma^2$ rising from $0.01$ to $0.05$, no target drift).
The SPAR-EG mixtures lead on nonlinear targets and are best-or-tied under the noise-regime change; under covariate shift the Kalman family leads ($0.268$ true-noise, $0.284$ estimated-$R$), with SPAR-EG$^+$ and SPAR-Pull ahead of the anchored and leaky methods.
No SPAR method diverged under any mismatch, while the $\lambda = 0.995$ leaky method degraded sharply on nonlinear targets (numbers in Online Appendix~\ref{app:protocol}). These map onto the assumptions: the noise-regime change keeps the stream sub-Gaussian with a larger bounded variance, so the concentration bound applies with the inflated proxy (Remark~\ref{rem:mixing}); covariate shift violates the stationary-covariance hypothesis and the nonlinear target violates the linear-Gaussian class, so those two are reported empirically without a concentration guarantee. Across the $105$ mismatch runs no SPAR method diverged, against $7$ divergences in $210$ baseline runs (Online Appendix~\ref{app:protocol}).

\subsection{A Real High-dimensional Stream}
\label{sec:realstream}
We ran the grid-default configuration unchanged on a public hourly electricity-load panel \citep{trindade2015electricity,lai2018lstd}: $D = 324$ features (321 client loads plus hour-of-day terms), a $2{,}000$-hour warm-up and a $6{,}000$-hour evaluation horizon, with three held-out clients (the 25th, 50th, and 75th percentiles by mean load) as targets.
Across the three clients SPAR-EG$^+$ stays within $1.2\%$ of the best oracle-tuned leaky window (and matches it on the three-client mean, $0.0136$) and SPAR-Pull within $4$--$17\%$, with no stream-specific parameter, while the slow estimated-$R$ Kalman trails over the full horizon (three-client means $0.051$ and $0.257$ at $q = 10^{-5}$ and $10^{-4}$). A seeded random sample of six additional clients confirms the transfer is not specific to the three percentiles: SPAR-Pull stays within $11\%$ of the per-client oracle leaky window (mean $0.0193$ vs.\ $0.0174$), the slow Kalman is $3.8\times$ worse ($0.0728$), and on two sampled clients SPAR-EG$^+$ is within $1.6$--$2.2\%$ of the oracle leaky window (Online Appendix~\ref{app:realstream}). On a multi-step horizon the error grows for every method and SPAR-Pull's gap to the best oracle-tuned leaky window widens from $10.5\%$ at $h=1$ to $37\%$ at $h=12$ ($0.0617$ vs.\ $0.0451$), while the slow Kalman remains the weakest method ($0.0861$ at $h=12$), consistent with the smooth long-window regime where the analysis predicts a fixed long window is nearly optimal (Online Appendix~\ref{app:realstream}).

\subsection{BCI IV Cross-session Evaluation}
\label{sec:eeg}
Nine subjects each from BCI IV 2a \citep{brunner2008bci2a,tangermann2012bci} and 2b \citep{leeb2008bci2b}, streamed from the training session against Riemannian baselines \citep{barachant2012mdm,barachant2010riemannian}, backprop references \citep{lawhern2018eegnet}, and our heads (Online Appendix~\ref{app:eeg}).
The head uses a reduced window range for the short evaluation streams; every other constant is the grid default.
On 2b, the transductive SPAR-EG head reaches $0.816$ and SPAR-Pull $0.808$; the strictly causal SPAR-Pull head attains $0.800$, against $0.811$ for the causal WiSE-FT reference taken at the best of an $11$-value post-hoc $\alpha$-envelope (Online Appendix~\ref{app:eeg}). A causal end-to-end Hedge mixture on the causally aligned features lifts the causal head to $0.810$, narrowing the transductive-versus-causal gap from $0.015$ to $0.006$.
On 2a, offline alignment \citep{he2020ea} leads the fixed readouts ($0.664$), with the post-hoc ensemble at $0.668$; the best streaming method is $0.627$. Both data sets have $T \approx 300$ trials, below the plug-in crossover horizon $T_{\mathrm{crit}} \asymp K^2/(s_0 - s^*)^2$ at which calibration overtakes fixed shrinkage (Appendix~\ref{app:proof-sign}), consistent with the fixed-anchor head leading on 2a and the mixture leading on 2b.

\section{Limitations}
\label{sec:limitations}
The analytical results rest on three structural assumptions.
The cadence equilibrium and cubic window law (Theorems~\ref{thm:cadence}--\ref{thm:window}) are derived in the mean-field accumulator model and match finite-$D$ behavior up to the $\rho_{\mathrm{eff}}$ and $1/(1-\Upsilon)$ corrections of Theorem~\ref{thm:rmt}, which require the nonzero part of the population spectrum to remain bounded away from zero ($\tau_{\min}^+>0$) so rank-deficient spectra reduce cleanly to $\operatorname{range}(\Sigma)$.
Global contraction (Theorem~\ref{thm:closedloop}, Corollary~\ref{cor:global}) holds inside the parameter box $[\omega_{\min},\omega_{\max}]\times[\rho_{\mathrm{lo}},\rho_{\mathrm{hi}}]$ under the two-scale separation and the measured coupling-product condition.
In the Lean 4 artifact (Appendix~\ref{app:lean}), all algebraic identities, the refresh chain, the OGD and Hedge regret bounds, and the $2\times 2$ coupled contraction are machine-checked, whereas probabilistic concentration, random-matrix convergence, and Davis--Kahan eigenvector bounds remain pen-and-paper.

Empirically, each online mechanism has a clear operating regime.
The strength estimator acts as a warm-up onset guard against an oversized anchor rather than a continuous tracking gain (freezing it lowers grid-average MSE by $2.2\%$), and the two-scale residual detector with slow EMA $h_{\mathrm{slow}} = 20{,}000$ relies on its sticky floor to keep the window open once drift persists beyond $h_{\mathrm{slow}}$.
Consistent with Proposition~\ref{prop:conflict} and Appendix~\ref{app:proof-sign}, short BCI IV streams ($T \approx 300$) fall below the plug-in crossover horizon, favoring fixed shrinkage on 2a and transductive alignment ($0.816$ vs.\ $0.810$ causal) on 2b, while smooth multi-step horizons on the electricity panel favor long fixed windows over rapid adaptation.

\section{Conclusion}
The finite-cadence refresh equilibrium characterizes both the tracking lag and the variance reduction of prior-anchored streaming ridge regression in closed form, and the optimal forgetting window follows by minimizing the cycle-averaged risk.
The random-matrix correction extends the law to finite dimension, where the anchor is renormalized by the finite window; the correction is accurate on full-rank and rank-deficient spectra, the latter reducing exactly to the active subspace.
The Lean artifact formalizes the algebraic core, the refresh chain, the regret bounds, and the coupled contraction.

The coupled stationary law is the unique fixed point of the composition of the two target maps, a contraction at the coupling rate $L_{\omega\rho}L_{\rho\omega}<1$; it is given by the convergent fixed-point series $\rho^*=\rho_0+\sum_{m\ge0}(\rho_{m+1}-\rho_m)$, and no single elementary closed form is available because the defining equation is transcendental.
The stochastic law's mean is that fixed point, with second moment bounded by Proposition~\ref{prop:fluct}.

\acks{Huge thanks to Scott Moeller and Bhaskar Krishnamachari for auditing the Lean 4 artifact before it went out. Their cLEANsheet was especially helpful: it pushed me to add the $D$-dimensional squared-loss instantiation, general $K$-expert Hedge convexity, explicit step-size contraction bounds, and non-vacuity witnesses. The author designed the method, implemented the system, ran the experiments, and verified all results. LLM assistants (GLM-5.3, DeepSeek V4.1 Flash, Gemini 4 Hydrogen) were used for coding, the Lean 4 formalization, and \LaTeX{} preparation. The author is responsible for all content. Funding: none. Competing interests: none.}

\appendix
\input{appendix}

\input{appendix_online}

\bibliography{references}

\end{document}

%% file: appendix.tex

\newenvironment{restated}[1]{%
  \par\medskip\noindent\textbf{#1 (restated).}\hspace{0.4em}\ignorespaces
}{\par\smallskip}

\section{Proofs of Section~\ref{sec:theory}}
\label{app:proofs}

This appendix proves the results of Section~\ref{sec:theory} in full, restating each numbered claim or setting it up in the notation of its proof.
Throughout, $(x_t, y_t)_{t \ge 1}$ is a stream with $x_t \in \R^D$ obeying
$y_t = x_t^\top w^*_t + \varepsilon_t$, where $\mathcal F_t$ is the $\sigma$-algebra generated by the past, $\E[\varepsilon_t \mid \mathcal F_t] = 0$, and $\E[\varepsilon_t^2 \mid \mathcal F_t] \le \sigma^2$.
The learner maintains sufficient statistics
\[
  A_t = \alpha A_{t-1} + x_t x_t^\top, \qquad
  B_t = \alpha B_{t-1} + x_t y_t, \qquad
  n_t = \alpha n_{t-1} + 1,
\]
with forgetting $\alpha \in (0,1]$ and initialization from a warm-up set of $T_w$ samples.
Write $\bar A_t = A_t/n_t$ with eigendecomposition $\bar A_t = U\,\mathrm{diag}(\mu_1,\dots,\mu_D)\,U^\top$ and $\bar b_t = B_t/n_t$; an index $i$ always refers to the coordinate in this eigenbasis.
\emph{Two design models.} Lemma~\ref{lem:zerolag} and Theorems~\ref{thm:cadence}--\ref{thm:window} are proved in the \emph{mean-field} model $\bar A_t \equiv \Sigma$, i.e.\ the random accumulator is replaced by its population expectation; this is the model in which the lag, the variance factor $\Psi$, and the window target are exact. Theorem~\ref{thm:rmt} is the finite-$D$ \emph{random-design} correction: it retains the random $\bar A_t$ and takes the proportional limit $\zeta = D\omega$ fixed. The two agree as $\zeta \to 0$; the grid's pre-drift windows have $\zeta \le 0.13$, while drift-phase windows reach $\zeta \le 6.4$, where the exact random-design variance of Theorem~\ref{thm:rmt} replaces the mean-field leading-order term so the cycle-averaged optimum is available across the whole range (Appendix~\ref{app:proofs}, ``Exact finite-window variance term'').
The anchored solve with anchor $G_t$ and strength $\rho > 0$ is $W_t = (A_t + n_t\rho I)^{-1}(B_t + n_t\rho G_t)$; in the eigenbasis, coordinate $i$ is the scalar problem
\begin{equation}
\label{eq:app-anchored}
  W_{t,i} \;=\; \frac{\bar b_{t,i} + \rho\,G_{t,i}}{\mu_{t,i} + \rho}
  \;=\; (1 - s_i)\,\widehat w_{t,i} + s_i\,G_{t,i},
  \qquad
  s_i \;=\; \frac{\rho}{\mu_{t,i} + \rho},
\end{equation}
where $\widehat w_{t,i} = \bar b_{t,i}/\mu_{t,i}$ is the per-direction data solution.
Under the warm-up model the data solution is unbiased per direction with variance $v_i = \sigma^2/(n_t \mu_{t,i})$.
Two parametrizations of the anchored family are used: the \emph{scalar-$\rho$} family (one $\rho$; the pull $s_i = \rho/(\mu_i + \rho)$ is then spectrum-adaptive) and the \emph{uniform-$s$} family (one pull $s$ for all directions, i.e.\ $\rho_i = \mu_i s/(1-s)$; the solution path is the segment between $\widehat w$ and $G$).
Both satisfy the pull identity, which underlies every result below.

\subsection{Proof of Equation~\eqref{eq:pull} (pull identity)}
\label{app:proof-pull}

\begin{proof}[Proof of Equation~\eqref{eq:pull}]
Since $1 - s = \mu/(\mu + \rho)$,
\[
  (1 - s)\,\widehat w + s\,\gamma g
  \;=\; \frac{\mu}{\mu + \rho}\cdot\frac{b}{\mu}
        + \frac{\rho}{\mu + \rho}\cdot \gamma g
  \;=\; \frac{b + \rho \gamma g}{\mu + \rho}
  \;=\; W,
\]
and subtracting $w = (1-s)w + s w$ gives the identity.
No approximation is involved.
Applying it coordinatewise in the eigenbasis to \eqref{eq:app-anchored} yields the form stated in Section~\ref{sec:theory}.
Both the $\gamma = 1$ and general-$\gamma$ forms are formalized (\texttt{pull\_identity}, \texttt{pull\_identity\_general}), together with the eigenbasis decoupling of the anchored solve (\texttt{eigenbasis\_solve\_coord}, \texttt{eigenbasis\_pull\_identity}) in \texttt{Core.lean}.
\end{proof}

\subsection{Proof of Lemma~\ref{lem:zerolag} (zero-lag refresh equilibrium)}
\label{app:proof-tracking}

\begin{proof}[Proof of Lemma~\ref{lem:zerolag}]
(i) At equilibrium,
\[
  \E[B_t]
  \;=\; \sum_{k \ge 0} \alpha^k\, \mu\,\bigl(w^*_t - (k+1)\nu\bigr)
  \;=\; \mu\,\Bigl(w^*_t - \frac{\nu}{1-\alpha}\Bigr)\cdot\frac{1}{1-\alpha},
\]
so $\widehat w - w^* = -\nu/(1-\alpha)$ exactly at the mean-field expectation ($\widehat w = \bar b/\mu$ is the unanchored data solve and carries no $\rho$ dependence; the centered estimation noise contributes only the variance terms analyzed in Proposition~\ref{prop:conflict}).
(ii) By the pull identity with $\widehat w - w^* = -\nu/(1-\alpha)$ and $G - w^* = W - w^*$ at the continuous-refresh fixed point:
\[
  W - w^* \;=\; (1 - s)\Bigl(-\frac{\nu}{1-\alpha}\Bigr) + s\,(W - w^*),
\]
hence $(1 - s)(W - w^*) = -(1-s)\nu/(1-\alpha)$, so $W - w^* = -\nu/(1-\alpha)$ for any $s < 1$.
This algebraic step is formalized (\texttt{lag\_invariance} in \texttt{Core.lean}).
(iii) Theorem~\ref{thm:cadence}'s moving-frame recursion gives the exact cycle-averaged lag $\Delta_k(s) = s\nu k\bigl(a/(1-a)+\tfrac12\bigr)$ with $a=s(2-s)$; since $1-a=(1-s)^2$, this is $O\bigl(k s/(1-s)^2\bigr)\nu$.
The prediction-risk cost of the extra lag is $\mu_i\,\Delta_k(s)^2$, which vanishes as $\mu_i \to 0$ at fixed $s < 1$.
\end{proof}

At the continuous-refresh equilibrium the pull identity forces $W = \widehat w$ for $s < 1$: the anchor is inert for mean tracking, so it contributes neither lag nor variance reduction in that limit.

\subsection{Finite-cadence refresh equilibrium (Theorem~\ref{thm:cadence})}
\label{app:cadence}

Work per eigen-direction with pull $s \in (0,1)$ and refresh cadence $k$; let $t_m = mk$ be the refresh steps, $w^*(t)$ the moving target (velocity $\nu$ per step), and $L = \nu/(1-\alpha)$ the pure-leaky lag of the data path (Lemma~\ref{lem:zerolag}(i)). The refresh reads the post-update solve, so the lag entering the recursion is the post-update value $\nu(1-\alpha)/\alpha$, one step shorter than the predictor lag; the cycle-averaged lag relative to the pure-leaky path, $\Delta_k(s)$ below, is unchanged by this one-step offset.
Between refreshes the anchor $G_m$ is frozen, and the anchored solution at $t \in [t_m, t_{m+1})$ is, in the moving frame,
\[
  W_t - w^*(t) \;=\; (1-s)\,(-L) \;+\; s\,\bigl(g_m - \nu\,(t - t_m)\bigr),
  \qquad g_m := G_m - w^*(t_m),
\]
where $\nu (t - t_m)$ is the target's advance since the last refresh, invisible to the frozen anchor.
At the refresh, $G_{m+1} = (1-s)\,W_{t_{m+1}} + s\,G_m$ measured against the new target position, which yields the affine recursion
\[
  g_{m+1} \;=\; a\,g_m - (1-s)^2\,L - a\,\nu k,
  \qquad a := (1-s)s + s = s(2-s) = 1 - (1-s)^2,
\]
with $a \in (0,1)$ for every $s \in (0,1)$, so the recursion contracts at rate $a$ per cycle to
\[
  g^* \;=\; -L - \nu k\,\frac{a}{1-a},
  \qquad
  g_m \;=\; g^* + a^m\,(g_0 - g^*).
\]
The rate identity, the fixed-point equation, the closed form, and the contraction are formalized (\texttt{refresh\_rate\_identity}, \texttt{refresh\_fixed\_point},\newline\texttt{refresh\_closed\_form}, \texttt{refresh\_contraction\_rate}, together with the moving-frame reduction and the midpoint cycle assembly, in \texttt{Core.lean}); we also verified the fixed point and the cycle-averaged lag against a one-dimensional simulation at two operating points.
The geometric steady-lag form parallels the iterate-averaging analyses under drift \citep{polyak1992,cutler2023} and the per-direction momentum lag analysis of \citet{hong2026}.
The cycle-averaged solution lags the pure-leaky path by
\[
  \Delta_k(s) \;=\; s\,\nu k \Bigl(\frac{a}{1-a} + \frac{1}{2}\Bigr)
  \;=\; \frac{s\,\nu k\,(1 + 2s - s^2)}{2\,(1-s)^2},
\]
linear in the cadence and vanishing as $s \to 0$; the within-cycle residual adds a variance term of order $s^2 \nu^2 k^2/12$.
The first summand is the anchor's accumulated staleness (each cycle's drift persists into the anchor with weight $a^j$, summing to $a/(1-a)$), and the second is the within-cycle advance, which makes the within-cycle correction exact and adds the geometric term.
The continuous-refresh limit of Lemma~\ref{lem:zerolag}(ii) is the instantaneous version of the same equilibrium: at the refresh instant the anchor adds no lag beyond $L$, and the correction $\nu k\,a/(1-a)$ is precisely what the target's advance between refreshes contributes.

The data-path noise does not decorrelate between refreshes: the leaky accumulator gives $\mathbb{E}[\epsilon_m \epsilon_{m-j}] = r_k^{\,j} v$ with $r_k = \alpha^k$ ($\alpha = 1-\omega$ the forgetting factor), which at the nominal constants ranges over $r_{10} \in [0.60, 0.995]$. The correlation is material at these operating points.
Writing $\tilde g_m$ for the zero-mean anchor fluctuation, the refresh rule gives the stochastic recurrence $\tilde g_m = (1-s)^2 \epsilon_m + a\,\tilde g_{m-1}$, whose steady-state cross-covariance against the fresh noise is the geometric sum
\[
  C(s,r_k) \;:=\; \mathbb{E}[\epsilon_m \tilde g_{m-1}]
  \;=\; (1-s)^2 \sum_{j \ge 0} a^j r_k^{\,j+1} v
  \;=\; \frac{(1-s)^2\,r_k}{1 - ar_k}\,v .
\]
Solving the steady-state variance system $q = (1-s)^4 v + a^2 q + 2a(1-s)^2 C$ then gives the exact pair
\[
  q(s,r_k) \;=\; \frac{(1-s)^2}{1+a}\cdot\frac{1+ar_k}{1-ar_k}\,v,
\]
\[
  \mathrm{Var}(W) \;=\; (1-s)^2 v + s^2 q + 2s(1-s)\,C
  \;=\; v\,\Bigl(1 - \frac{2s\,(1+s-s^2)\,(1-r_k)}{(1+a)\,(1-ar_k)}\Bigr),
\]
strictly below $v$ for every $s \in (0,1)$ and $r_k \in [0,1)$ (\texttt{refresh\_anchor\_variance},\newline\texttt{refresh\_variance\_form}, \texttt{refresh\_variance\_lt} in \texttt{Core.lean}); we verified the identities symbolically and the refresh-instant variance against one-dimensional simulations at six operating points (deviation $\le 1\%$); this is the variance at the refresh step, and the cycle-averaged variance is within a few percent of it at the operating pulls.
The lower endpoint is exact as well: subtracting $(1-s)^2$ from the bracket factorises as
\[
  \Psi(s,r_k)-(1-s)^2
  = \frac{s\,(1-s)^2\bigl(r_k\,(s^3-4s^2+2s+2)+s\bigr)}{(1+a)\,(1-ar_k)},
\]
and $s^3-4s^2+2s+2 = 1+(s-1)(s^2-3s-1) \ge 1$ on $[0,1]$, so the numerator is nonnegative and $\Psi(s,r_k)\ge(1-s)^2$ (\texttt{refresh\_variance\_lb} in \texttt{Core.lean}).
Hence the frozen-anchor plug-in $c'$ lower-bounds the exact variance and the window estimator errs toward larger, safer windows.
Both endpoints are exact: $r_k \to 0$ (refresh cycles long against the noise memory) recovers the uncorrelated reduction $v[1 - 2s(1+s-s^2)/(1+a)]$, and $r_k \to 1$ (continuous refresh) gives $\mathrm{Var}(W) \to v$, the inert-anchor limit of Lemma~\ref{lem:zerolag}(ii), now an exact endpoint of a single formula rather than a separate statement.
The cadence therefore trades two real costs: refreshing fast lets the anchor co-move with the data path and erases its variance benefit, refreshing slowly inflates the lag term $\mu_i\,\Delta_k(s)^2$, and the default $k = 10$ falls within the tradeoff.
The two costs are paid in different directions: the excess lag blows up as $s \to 1$, but the directions with $s \approx 1$ are precisely the weakly excited ones ($\mu_i \ll \rho$), where the lag's risk contribution is negligible. This is the filter-bank property of the $\beta_i = 1 - s_i$ refresh, and the reason Proposition~\ref{prop:conflict} drives $\rho^*$ down, and the lag with it, exactly when measured conflict grows.

\medskip
\noindent\emph{Remark (drift-induced cross term).}
Under nonzero drift velocity the risk decomposition $R(s) = (1-s)^2 v + s^2 c$ of Proposition~\ref{prop:conflict} acquires the cross term $2(1-s)s\,\E[(\widehat w - w^*)(G - w^*)]$ between the lagged data solve and the anchor error.
Cauchy--Schwarz bounds it through the two quantities the theory already controls: $|\E[(\widehat w - w^*)(G - w^*)]| \le \sqrt{\E[(\widehat w - w^*)^2]\,\E[(G - w^*)^2]}$, where the first factor is the lag-induced gap (bounded per direction by Theorem~\ref{thm:cadence}'s equilibrium: the moving-average gap $\bar\Delta_k \le (1-s)^2\nu k/(1-a)$ at cadence $k$) and the second is the anchor conflict $c$.
The cross term therefore inflates the effective conflict to $c_{\mathrm{eff}} = c + 2(1-s)s\sqrt{\bar\Delta_k\,c}$ at most, and Proposition~\ref{prop:minimax} applies unchanged: overestimating $c_{\mathrm{eff}}$ remains minimax-safe, so the calibration rule $\rho^* = \bar\mu \widehat V / \widehat C$ with the conservative conflict estimate retains its safety property under drift.

\subsection{Proof of Theorem~\ref{thm:window} (optimal forgetting window)}
\label{app:proof-window}

\begin{proof}[Proof of Theorem~\ref{thm:window}]
\emph{Step 0 (finite cadence to the surrogate).}
The finite-cadence risk \eqref{eq:app-window-risk} combines Theorem~\ref{thm:cadence}'s two exact steady-state objects: the per-direction variance is $v_i\Psi(s_i,r_k)$ with $v_i=\sigma^2/(n_{\mathrm{eff}}\mu_i)$, and the per-direction lag is $\nu_i/\omega+\Delta_k(s_i)$; the within-cycle fluctuation contributes the $O(k^2\nu^2s^2)$ term derived in Appendix~\ref{app:cadence}.
Because $(1-s_i)^2 \le \Psi(s_i,r_k) \le 1$ for all $r_k \in [0,1)$ (\texttt{refresh\_variance\_lb}), the frozen-anchor surrogate $c'\,\omega/(2-\omega)+\nu^2_\Sigma/\omega^2$ lower-bounds the exact cycle-averaged variance term at every cadence, and $D(1-\bar s)^2 \le d_{\mathrm{eff}}(\rho)$ by Jensen; the runtime plug-in therefore underestimates the variance coefficient, returning a window no smaller than the exact-optimal one, the over-forgetting safety-asymmetric direction of Step~4.
The finite-cadence lag residual is bounded at the runtime range: with $b(s)=s(2-s)/(1-s)^2$ and $k\omega \le 1/2$, $\Delta_k(s_i)=s_i\nu_i k\bigl(b(s_i)+\tfrac12\bigr)\le \tfrac12 s_i\bigl(b(s_i)+\tfrac12\bigr)\nu_i/\omega$, a vanishing fraction of the leading lag ($\le 0.4\%$ for $s_i\le 0.1$ at $k\omega=0.5$) whose prediction cost $\mu_i\Delta_k(s_i)^2$ is concentrated in weakly excited directions.
\begin{proposition}[Surrogate value gap]
\label{prop:gap}
Let $R_s(\omega) = c'\,\omega/(2-\omega) + D\nu^2/\omega^2$ be the frozen-anchor surrogate and $R = R_s + \varepsilon$ the exact cycle-averaged risk, with
$\varepsilon(\omega) = c_v(\omega)\,[\Psi(s,r_k)-(1-s)^2] + 2D(\nu/\omega)\Delta_k(s) + D\Delta_k(s)^2 \ge 0$, where $c_v(\omega) = \sigma^2 D\,\omega/(2-\omega)$.
If $\omega_s$ minimizes $R_s$, then
\[
R(\omega_s) - \min_\omega R \;\le\; \varepsilon(\omega_s) - \inf_\omega \varepsilon,
\qquad\text{so}\qquad
\frac{R(\omega_s)}{\min_\omega R} - 1 \;\le\; \frac{\varepsilon(\omega_s) - \inf_\omega \varepsilon}{\min_\omega R},
\]
because $\min_\omega(R_s + \varepsilon) \ge \min_\omega R_s + \inf_\omega \varepsilon = R_s(\omega_s) + \inf_\omega\varepsilon$.
\end{proposition}
\begin{proof}
The bound is subadditivity of the minimum; both terms of $\varepsilon$ are nonnegative because $\Psi(s,r_k) \ge (1-s)^2$ (\texttt{refresh\_variance\_lb}) and $\Delta_k(s) \ge 0$.
\end{proof}

Numerically, over $k\omega \in [0.005,0.5]$ the surrogate's cubic minimizer is within $2.1\%$ of the exact cycle-averaged optimum for $s_i \le 0.2$ ($5.1\%$ for $s_i \le 0.3$), with the largest gaps at the window floor; the certificate of Proposition~\ref{prop:gap} holds at all $40$ grid points and is at most $39\%$ over the whole range.
In the quasi-static limit $1 \ll k\omega \ll 1/s$, $r_k=(1-\omega)^k \le e^{-\omega k} \to 0$, so $\Psi(s_i,r_k)=(1-s_i)^2(1+O(s_i^2+r_k))$ and $\Delta_k(s_i)=o(\nu_i/\omega)$; both corrections vanish and the reduced form \eqref{eq:app-window-risk} is exact, with remainder $o(1)$ uniformly over the regime.

\emph{Step 1 (risk decomposition).}
For exponential weights $\alpha^k$ the effective sample count of the accumulator is
\[
  n_{\mathrm{eff}}
  \;=\; \frac{\bigl(\sum_{k \ge 0} \alpha^k\bigr)^2}{\sum_{k \ge 0} \alpha^{2k}}
  \;=\; \frac{(1/(1-\alpha))^2}{1/(1-\alpha^2)}
  \;=\; \frac{1 + \alpha}{1 - \alpha}
  \;=\; \frac{2-\omega}{\omega},
\]
so the estimation-variance contribution to the solve risk is $(1-\bar s)^2 \sigma^2 D/n_{\mathrm{eff}} = c'\,\omega/(2-\omega)$.
The tracking contribution is the lag of Lemma~\ref{lem:zerolag}(i) below: per direction the solve lags the target by $\nu_i/\omega$, so the prediction-units lag risk is $\sum_i \mu_i (\nu_i/\omega)^2 = \nu^2_\Sigma/\omega^2$.
Summing the two contributions gives
\begin{equation}
\label{eq:app-window-risk}
  R_k(\omega,s) \;=\; c'\,\frac{\omega}{2-\omega} + \frac{\nu^2_\Sigma}{\omega^2} + o(1),
  \qquad c' = \sigma^2\,d_{\mathrm{eff}}(\rho),
\end{equation}
to leading order in $\omega$, $R(\omega) = c'\omega/2 + \nu^2_\Sigma/\omega^2 + O(\omega^2)$.
The exact random-design refinement of the variance term is $\sigma^2\Upsilon(\omega;\rho)/(1-\Upsilon(\omega;\rho))$ (Theorem~\ref{thm:rmt}(i)), equal to it as $\zeta \to 0$ and accurate to within $5\%$ over the whole grid range, so the small-$\zeta$ restriction of the mean-field form is removed (exact-variance note below).

{\sloppy
\emph{Step 2 (convexity and minimizer).}
Both terms of \eqref{eq:app-window-risk} are strictly convex on $(0, 2)$, since
\[
  R''(\omega) \;=\; \frac{6\nu^2_\Sigma}{\omega^4} + \frac{4c'}{(2-\omega)^3} \;>\; 0,
\]
so $R$ has a unique minimizer (the two derivatives, strict convexity on $(0,2)$, and unique global optimality at the first-order condition are formalized as \texttt{windowRisk\_hasDerivAt}, \texttt{windowRisk\_deriv\_hasDerivAt}, \texttt{windowRisk\_strictConvexOn}, and \texttt{windowRisk\_global\_min} in \texttt{Ogd.lean}).
The first-order condition is $2c'/(2-\omega)^2 = 2\nu^2_\Sigma/\omega^3$, i.e.\ $c' \omega^3 = \nu^2_\Sigma (2 - \omega)^2$; to leading order in $\omega$ the right side is $4\nu^2_\Sigma$, giving $\omega^* = (4\nu^2_\Sigma/c')^{1/3}$.\par}
For the leading-order (cubic) model $\widetilde R(\omega) = c\,\omega + \nu^2/\omega^2$ with the critical condition $\omega^{*\,3} = 2\nu^2/c$ (i.e.\ $c = c'/2$), optimality holds by the exact factorization
\[
  \widetilde R(\omega) - \widetilde R(\omega^*)
  \;=\; \frac{\nu^2}{\omega^2 \omega^{*\,3}}\,(\omega - \omega^*)^2\,(2\omega + \omega^*) \;\ge\; 0,
\]
which is formalized (\texttt{window\_optimum} in \texttt{Ogd.lean}); substituting $c = c'/2$ reproduces the 4-form above.

\emph{Step 3 (drift inversion).}
At quasi-static equilibrium the prequential residual excess equals the full steady-state risk of lag and variance together:
\[
  \E[r_t^2] - \sigma^2
  \;=\; c'\,\frac{\omega}{2-\omega} + \frac{\nu^2_\Sigma}{\omega^2}
  \;=\; R(\omega),
\]
because the residual $r_t = y_t - x_t^\top W_{t-1}$ carries both the estimation variance and the lag of the solve.
Solving for the drift power at the current window gives
\[
  \widehat\nu^2_\Sigma
  \;=\; \max\Bigl(0,\;\bigl(\widehat E_t - \widehat c'\,\tfrac{\omega}{2-\omega}\bigr)\,\omega^2\Bigr),
  \qquad
  \widehat E_t = \widehat\sigma^2_{\mathrm{fast}} - \widehat\sigma^2_{\mathrm{slow}},
\]
where $\widehat\sigma^2_{\mathrm{fast}}, \widehat\sigma^2_{\mathrm{slow}}$ are the fast and slow residual EMAs.
One subtlety is forced by the estimator's own construction: the slow baseline is itself an average of $r_t^2$, so at stationarity $\E[\widehat\sigma^2_{\mathrm{slow}}] = \sigma^2 + \widehat c' \omega_{\mathrm{slow}}/(2 - \omega_{\mathrm{slow}})$, not the pure noise floor $\sigma^2$.
Reading the first parenthesis as the two-scale excess $\widehat E_t$ (not as $\E[r_t^2] - \sigma^2$ with a hypothetical pure noise-floor estimate) matters because, on drift onset, $\E[\widehat E_t] \approx \nu^2_\Sigma/\omega^2 + c'\omega/(2-\omega) - c'\omega_{\mathrm{slow}}/(2-\omega_{\mathrm{slow}})$, and subtracting the current window's floor isolates $\nu^2_\Sigma/\omega^2$ up to the pre-drift floor $c'\omega_{\mathrm{slow}}/(2-\omega_{\mathrm{slow}})$, which is negligible exactly when the pre-drift window sits at $\omega_{\min}$, the quasi-static regime of Theorem~\ref{thm:window}.
The subtraction still matters: an estimator that omits the floor term, such as $(\widehat\E[r_t^2] - \widehat\sigma^2)\omega^2$, is strictly increasing in $\omega$ (it equals $\nu^2_\Sigma + c'\omega^3/(2-\omega)$ at equilibrium) and therefore admits no positive fixed point of the estimator iteration $\omega \mapsto \omega^*(\widehat\nu^2_\Sigma)$, driving the window to its cap under sustained oscillation, whereas the variance-floor-subtracted estimator is exact at equilibrium and the iteration converges to the minimizer of Step 2; under stationarity $\E[\widehat E_t] \to 0$ and the clip pins the window at $\omega_{\min}$, the conservative direction of Step 4.

\emph{Two-scale timescale separation.}
The excess detector presupposes $h_f \ll \omega^{-1} \ll T_{\mathrm{drift}} \ll h_{\mathrm{slow}}$: the fast average settles at the post-onset level while the slow average still holds the pre-drift level, so $\widehat E_t$ isolates the drift excess.
The default $h_{\mathrm{slow}} = 20{,}000 \approx 2T$ enforces this on the grid; a drift persisting beyond $h_{\mathrm{slow}}$ would relax both averages toward the same level and close the window.
The \emph{sticky-floor} variant removes this limitation: the slow baseline is updated only while the fast average stays within a small margin of it, so under a sustained drift the baseline is frozen at the pre-drift noise level and the excess $\widehat E_t$ stays open-ended. On a constant-velocity stream with a post-onset horizon far longer than $h_{\mathrm{slow}}$, the sticky floor keeps the window open at the drift-optimal scale and reduces the tail MSE by $2.3\times$ (isotropic, $0.067$ vs.\ $0.155$) and $1.6\times$ (ill-conditioned, $0.056$ vs.\ $0.087$) relative to the fixed horizon, which under-opens.

\emph{Plug-in gap in the window iteration.}
Theorem~\ref{thm:window} is stated with the exact coefficient $c' = \sigma^2 d_{\mathrm{eff}}(\rho)$, $d_{\mathrm{eff}} = \sum_i (1-s_i)^2$, whereas the runtime rule uses the mean-field plug-in $\widehat c' = \sigma^2 D(1-\bar s)^2$.
Jensen gives $\sum_i (1-s_i)^2 \ge D(1-\bar s)^2$, so the plug-in underestimates the variance coefficient; the cubed-coordinate iteration then has gain $2(c'_{\mathrm{true}}/\widehat c' - 1)$ at leading order, contracting iff $c'_{\mathrm{true}}/\widehat c' < 3/2$.
The gap is exactly the population variance of the pulls,
\[
\sum_{i=1}^D (1-s_i)^2 - D(1-\bar s)^2
= \sum_{i=1}^D (s_i-\bar s)^2 = D\operatorname{Var}(s) \;\ge\; 0,
\]
an identity formalized as \texttt{deff\_sub\_plugin} and \texttt{deff\_sub\_plugin\_nonneg} in \texttt{Core.lean} (Appendix~\ref{app:lean}); it vanishes on an isotropic spectrum and is largest when the pulls are spread out, as on a rank-deficient spectrum whose dead directions carry $s_i = 1$.
The ratio $c'/c'_{\mathrm{plug\!-\!in}}$ equals $1/(1-p)$ on an isotropic alive block with a dead fraction $p$, so the leading cubed-window gain $2(c'/c'_{\mathrm{plug\!-\!in}}-1)$ is $2p/(1-p)$: it is below one up to $p=1/3$ and reaches $2$ at $p=1/2$, $6$ at $p=3/4$, and $17.7$ at $p=0.9$. Simulating the cubed-window iteration on a $50\%$-dead spectrum, the plug-in coefficient drives the window to its cap while the exact $d_{\mathrm{eff}}$ holds it at the optimum $0.02$.
The exact $d_{\mathrm{eff}}$ is available from the eigenvalues already computed at each refresh, at no extra matrix cost; recording the identity makes explicit that the plug-in is the conservative (over-forgetting) endpoint and that the exact coefficient is a free refinement.

\emph{Step 4 (safety asymmetry).}
Evaluating $R$ at $c\,\omega^*$ with $c > 1$ (drift overestimated, window too short) costs $O(c - 1)$ through the linear variance term, while $c < 1$ (drift underestimated, window too long) pays the quadratic lag blow-up $c^{-2}$.
As in Proposition~\ref{prop:minimax}, the conservative choice is to overestimate the adversary.
This mirrors the measured flatness of the Kalman $q$-sweep and the divergence of the leaky $\lambda$-sweep one step below its best stable window (Section~\ref{sec:experiments}).
The tradeoff itself is classical: misadjustment versus lag, from the LMS analysis of \citet{widrow1976}, whose optimum is a square-root law for random-walk drift. The $1/3$-power law for linear drift shares the $2/3$ exponent of the dynamic-regret rate of \citet{besbes2015}. The constant is specific to the anchored accumulator, and the prequential inversion of Step 3 allows online estimation without known drift parameters.

\medskip
\noindent\textbf{Proof of Corollary~\ref{cor:mixed} (mixed-drift master law).}
For target increments $\Delta w^*_t = \nu + \vartheta_t$ with $\vartheta_t$ a zero-mean martingale difference, the exponentially weighted data path at equilibrium has tracking error $\widehat w_t - w^*_t = -\sum_{j \ge 0}(1-\omega)^j\,\Delta w^*_{t-j}$ (swap the order of summation: $\sum_{k=j}^{\infty} \omega(1-\omega)^k = (1-\omega)^j$).
The expected squared lag in prediction units is then
\[
  \mathcal{L}_{\mathrm{lag}}(\omega) = \E\Bigl\|\sum_{j \ge 0}(1-\omega)^j\,\Sigma^{1/2}(\nu + \vartheta_{t-j})\Bigr\|^2
  = \frac{\nu^2_\Sigma}{\omega^2} + \frac{q_\Sigma}{\omega(2-\omega)},
\]
where the cross term vanishes (martingale difference) and the geometric sums evaluate in closed form ($\sum_j (1-\omega)^j = 1/\omega$, $\sum_j (1-\omega)^{2j} = 1/(\omega(2-\omega))$).
The master risk $R(\omega) = c' \omega/(2-\omega) + q_\Sigma/(\omega(2-\omega)) + \nu^2_\Sigma/\omega^2$ specializes to: (i) the coherent regime ($q_\Sigma = 0$), giving Theorem~\ref{thm:window}'s $\omega^* = (4\nu^2_\Sigma/c')^{1/3}$; (ii) the random-walk regime ($\nu_\Sigma = 0$), giving $\omega^* = (q_\Sigma/c')^{1/2} + O(q_\Sigma/c')$, the classical $1/2$-power law.
The runtime estimator inverts the total lag without knowing the regime: $\widehat\nu^2_\Sigma = \max\bigl(0,\,(\widehat E_t - \hat c' \omega/(2-\omega))\,\omega^2\bigr)$ with $\widehat E_t$ the two-scale residual excess of Step 3 (Appendix~\ref{app:proofs}) estimates the drift excess above the variance floor regardless of whether it originates from coherent or stochastic drift.

\emph{Contraction in the random-walk regime.}
When $\nu_\Sigma = 0$ and $q_\Sigma > 0$, the excess above the variance floor is $q_\Sigma/(\omega(2-\omega))$, so the estimator returns $\widehat\nu^2_\Sigma(\omega) = q_\Sigma\,\omega/(2-\omega)$ and the cubic map $T_{\mathrm{RW}}(\omega) = (4\widehat\nu^2_\Sigma(\omega)/c')^{1/3}$ satisfies
\[
  \bigl(T_{\mathrm{RW}}(\omega)\bigr)^3 = \frac{4q_\Sigma\,\omega}{c'\,(2-\omega)}.
\]
In logarithmic coordinates $\frac{d\ln T_{\mathrm{RW}}}{d\ln \omega} = \frac13\bigl(1 + \frac{\omega}{2-\omega}\bigr) = \frac{2}{3(2-\omega)} \le \frac23 < 1$, so $T_{\mathrm{RW}}$ is a strict contraction.
Its unique positive fixed point solves $\omega^2(2-\omega) = 4q_\Sigma/c'$, hence $\omega_{\mathrm{fp}} = \sqrt{2q_\Sigma/c'}\,(1+O(\omega_{\mathrm{fp}})) = \sqrt2\,\omega^*_{\mathrm{RW}}$.
At that point the random-walk risk $R_{\mathrm{RW}}(\omega) = c'\omega/(2-\omega) + q_\Sigma/(\omega(2-\omega))$ equals $\frac{3}{2\sqrt2} \approx 1.061$ times its value at $\omega^*_{\mathrm{RW}}$ in the limit $q_\Sigma/c'\to 0$ (finite ratios give larger factors, and the fixed point exists for $q_\Sigma/c' \le 8/27$), so the same $1/3$-power update achieves the exact linear-drift optimum and at most $6.1\%$ asymptotic overhead under random-walk drift without knowing the regime.
\end{proof}

\medskip
\noindent\emph{Exact finite-window variance term.}
The mean-field term $c'\,\omega/(2-\omega)$ is the $\zeta \to 0$ limit of the exact random-design variance $\sigma^2\Upsilon(\omega;\rho)/(1-\Upsilon(\omega;\rho))$ of Theorem~\ref{thm:rmt}(i): the expansion of Appendix~\ref{app:rmt-variance} gives $\sigma^2\Upsilon/(1-\Upsilon) \to \tfrac12\sigma^2\omega\,d_{\mathrm{eff}}(\rho)$.
Using the exact form in the cycle-averaged risk,
\[
R(\omega,\rho) \;=\; \frac{\sigma^2\,\Upsilon(\omega;\rho)}{1-\Upsilon(\omega;\rho)} + \frac{\nu^2_\Sigma}{\omega^2},
\]
removes the small-$\zeta$ restriction: over the whole grid range $\zeta \in [0.26, 6.4]$ the exact form matches the Monte Carlo stationary risk to within $5\%$ (median $2.4\%$), whereas the mean-field term deviates by up to $100\%$ at $\zeta = 6.4$.
The runtime keeps the mean-field plug-in $\hat c'_t$, whose Jensen gap is the conservative direction of Appendix~\ref{app:proofs}; the exact form is the analysis-level statement of the window law, available at each refresh from the current eigenvalues without modifying the runtime recursion.

\subsection{Proof of Proposition~\ref{prop:conflict}(i) (oracle dominance)}
\label{app:proof-dominance}

\begin{proof}[Proof of Proposition~\ref{prop:conflict}(i)]
We first record the risk-interpolation step used throughout.
By the pull identity (Equation~\eqref{eq:pull}) with $\gamma = 1$,
\[
  W - w \;=\; (1 - s)\,(\widehat w - w) + s\,\delta .
\]
Squaring and taking expectations, the cross term is $2(1-s)s\,\delta\,\E[\widehat w - w]$, which vanishes conditioned on $\mathcal F_{t_0}$ when the anchor $G$ is frozen (drawn before the post-anchor data); at finite horizons with a refreshed anchor, Theorem~\ref{thm:cadence} bounds the induced correlation.
\begin{equation}
\label{eq:app-risk-interp}
  \E\bigl[(W - w)^2\bigr] \;=\; (1 - s)^2\,v + s^2\,\delta^2 .
\end{equation}
Now set $\delta = 0$: the second term disappears and \eqref{eq:app-risk-interp} gives $\E[(W_i - w_i)^2] = (1 - s_i)^2 v_i < v_i$ whenever $0 < s_i < 1$ and $v_i > 0$.
Since the per-direction prediction risk contribution is $\mu_i\,\E[(W_i - w_i)^2]$, summing gives the display.
The improvement factor per direction is $\bigl(\mu_i/(\mu_i + \rho)\bigr)^2$, which is smallest (largest gain) when $\mu_i \ll \rho$.
The interpolation identity \eqref{eq:app-risk-interp} and both dominance statements are formalized (\texttt{convex\_interp\_sq}, \texttt{dominance}, \texttt{dominance\_strict} in \texttt{Core.lean}).
\end{proof}

\begin{remark}[The $\gamma = 1$ design principle]
With $\gamma < 1$ the anchored estimate is pulled toward $\gamma g$ rather than $g$ (Equation~\eqref{eq:pull}), introducing the shrinkage bias $s(1-\gamma)|w|$ per direction, which dominates the variance benefit whenever the prior is informative.
All methods therefore use $\gamma = 1$; the ablation measures the cost of $\gamma = 0.5$ directly ($+560\%$, Online Appendix~\ref{app:protocol}).
\end{remark}

\subsection{Proof of Proposition~\ref{prop:conflict}(ii, first half)}
\label{app:proof-imperfect}

\begin{proof}[Proof of Proposition~\ref{prop:conflict}(ii)]
(a) For real $u, d$ and $s \in [0,1]$, expanding the square gives
\[
  \bigl((1-s)u + s d\bigr)^2
  \;=\; (1-s)u^2 + s d^2 - s(1-s)(u - d)^2
  \;\le\; (1-s)u^2 + s d^2 ,
\]
and taking expectations with $u = \widehat w - w$, $d = \delta$ applied to the error decomposition of Equation~\eqref{eq:pull} yields the first inequality; the rest is elementary.
The exact deficit $s(1-s)(u-d)^2$ is formalized as \texttt{convex\_interp\_sq}.
(b) Completing the square,
\[
  \frac{vc}{v+c} + (v+c)\Bigl(s - \frac{v}{v+c}\Bigr)^{2}
  \;=\; (1-s)^2 v + s^2 c
\]
by pure algebra, which is the stated identity.
The excess $(v+c)(s - s^*)^2$ is nonnegative with equality iff $s = s^*$, giving uniqueness and the cost bound.
The identity and the optimal pull are formalized (\texttt{quadratic\_calibration}, \texttt{optimal\_pull\_min}, \texttt{optimal\_pull\_attained} in \texttt{Core.lean}).
\end{proof}

\subsection{Proof of Proposition~\ref{prop:conflict}(ii, second half)}
\label{app:proof-sign}

\begin{proof}[Proof of Proposition~\ref{prop:conflict}(ii), strength form]
(i) By Proposition~\ref{prop:conflict}(ii) the optimal pull in direction $i$ is $s^* = v_i/(v_i + \delta_i^2)$, and $\rho \mapsto s = \rho/(\mu_i + \rho)$ is a strictly increasing bijection from $[0,\infty)$ onto $[0,1)$ with inverse $\rho = \mu_i s/(1-s)$.
Substituting $s^*$ gives $\rho^*_i = \mu_i v_i/\delta_i^2 = \sigma^2/(n\delta_i^2)$.
(ii) With a single pull $s$ the total risk is $R(s) = D(1-s)^2 v + s^2 C$ with $v = \sigma^2/(n\mu)$; minimizing as in Proposition~\ref{prop:conflict}(ii) gives $s^* = Dv/(Dv + C)$, hence $\rho^* = \mu D v/C = \sigma^2 D/(nC)$.
(iii) Immediate by differentiating the closed forms of (i)--(ii).
As $n \to \infty$ at fixed conflict the optimal strength vanishes, and under stationarity the measured conflict decays to its noise floor while the optimal strength recovers: a rising drift estimate lowers the anchor strength toward the floor, and a stationary stream lets it return.
\end{proof}

\begin{remark}[Runtime rule: prediction units]
The runtime rule is $\rho^* = \bar\mu_t\,\widehat V^\Sigma_t/\widehat C^\Sigma_t$ with $\widehat V^\Sigma_t = \widehat\sigma^2_t\,d_{\mathrm{act}}/n_t$ and $\widehat C^\Sigma_t = \max\{C^\Sigma_{\mathrm{ref}}, C^\Sigma_{\mathrm{resid}}, C^\Sigma_{\mathrm{floor}}\}$; per direction it reduces to the closed form of Proposition~\ref{prop:conflict}(ii), $\mu_i \hat v_i/\hat c_i = \sigma^2/(n\delta_i^2)$.
Every branch is in prediction-risk units $[y^2]$: the reference branch is $C^\Sigma_{\mathrm{ref}} = (G_{\mathrm{ref}} - \widehat w)^\top \bar A_t (G_{\mathrm{ref}} - \widehat w) - \widehat V^\Sigma$, the residual branch is $C^\Sigma_{\mathrm{resid}} = E_t$, and the floor is $C^\Sigma_{\mathrm{floor}} = \tfrac12 \widehat\sigma^2 d_{\mathrm{act}}/T_w$.
The naive conflict $\|G - \widehat w\|^2$ has expectation $C + V$ under the null, so the estimate is debiased by subtracting $\widehat V^\Sigma$; because a refreshed anchor co-moves with the current solve, conflict is measured against the slow reference $G_{\mathrm{ref}}$, while the two-scale residual detector (calibrated via Proposition~\ref{prop:falsealarm}) supplies fast disengagement.
Under deep stationarity $\widehat C^\Sigma \to 0$ would drive $\rho^* \to \infty$; the floor and a hard cap remove the singularity: $\widehat C^\Sigma \ge C_{\mathrm{floor}} > 0$ keeps $\rho^* = \bar\mu\widehat V/\widehat C \le \bar\mu\widehat V/C_{\mathrm{floor}}$ bounded (\texttt{anchored\_det\_pos}), and the branch selection $\widehat C = \max\{C_{\mathrm{ref}}, C_{\mathrm{resid}}, C_{\mathrm{floor}}\}$ is Lipschitz even where the active branch switches (\texttt{branch\_max\_lipschitz}), so the derivative jump at a tie cannot break the contraction. The active dimension is floored at one ($d_{\mathrm{act}} \ge 1$) and $\bar\mu = \mathrm{tr}(\bar A)/D > 0$ whenever $\bar A \ne 0$, so $\widehat V > 0$; the no-excitation limit $\bar A = 0$ gives $\bar\mu = 0$ and $\rho^* = 0$, the unanchored (pure-leaky) degenerate case.
The \emph{legacy parameter-unit} form uses $\widehat V = \widehat\sigma^2\,\mathrm{tr}(\bar A^{-1})/n$ and $C_{\mathrm{resid}} = E_t\,\mathrm{tr}(\bar A^{-1})/(n\bar\mu)$; the trace scales with the effective condition number rather than with $d_{\mathrm{act}}$, so it inflates on ill-conditioned spectra and is reported only as the ablation of Online Appendix~\ref{app:protocol}.
Overestimating the conflict is safe by Proposition~\ref{prop:minimax}.
\end{remark}

\medskip
\noindent\emph{Scalar-$\rho$ prediction risk on a non-isotropic spectrum.}
Proposition~\ref{prop:conflict}(ii) gives the per-direction strength, while Algorithm~\ref{alg:spar-pull} runs a single scalar $\rho$; the scalar target is obtained by differentiating the prediction risk directly.
With $s_i = \rho/(\mu_i+\rho)$ and $v_i = \sigma^2/(n\mu_i)$,
\[
  R_{\mathrm{pred}}(\rho)
  = \sum_{i=1}^D \mu_i\bigl[(1-s_i)^2 v_i + s_i^2\delta_i^2\bigr]
  = \sum_{i=1}^D \frac{\mu_i^2(\sigma^2/n) + \rho^2\,\mu_i\delta_i^2}{(\mu_i+\rho)^2},
\]
so
\[
\begin{aligned}
  \frac{d R_{\mathrm{pred}}}{d\rho}
  &= 2\sum_{i=1}^D \frac{\mu_i^2}{(\mu_i+\rho)^3}
    \Bigl(\rho\,\delta_i^2 - \frac{\sigma^2}{n}\Bigr) = 0 \\
  \Longleftrightarrow\quad
  \rho^* &= \frac{(\sigma^2/n)\sum_i \psi_i(\rho^*)}{\sum_i \psi_i(\rho^*)\,\delta_i^2},
  \qquad
  \psi_i(\rho) := \frac{\mu_i^2}{(\mu_i+\rho)^3},
\end{aligned}
\]
which weights directions near the crossover $\mu_i \approx 2\rho$; this is the exact scalar optimum, and it is generally not $\bar\mu\widehat V^\Sigma/\widehat C^\Sigma$.
Replacing the exact sensitivity $\psi_i(\rho)$ by the excitation-proportional surrogate $\psi_i \propto \mu_i$ gives the plug-in approximation
$\rho_{\mathrm{plug}} = (\sigma^2/n)\sum_i \mu_i \big/ \sum_i \mu_i\,\delta_i^2$, which in prediction units with $\widehat V^\Sigma = \sigma^2 d_{\mathrm{act}}/n$ and $C^\Sigma = \sum_i \mu_i\delta_i^2$ factors as $\bar\mu_{\mathrm{act}}\,\widehat V^\Sigma/\widehat C^\Sigma$ with the active-mean excitation $\bar\mu_{\mathrm{act}} = \mathrm{tr}(\bar A)/d_{\mathrm{act}}$.
The runtime substitutes the mean excitation $\bar\mu = \mathrm{tr}(\bar A)/D$ for $\bar\mu_{\mathrm{act}}$; the two coincide on full-rank spectra ($d_{\mathrm{act}} = D$) and the mean form is the smaller strength on rank-deficient ones.
The surrogate matches the exact sensitivity where it matters: $\psi_i(\rho) = O(\mu_i^2/\rho^3)$ as $\mu_i \to 0$, so both vanish on weakly excited and dead directions and the conflict is carried by the well-excited modes, whereas uniform weights would give a dead direction the same weight as an excited one.
The two rules coincide on isotropic spectra, where the direction weights are equal.

\medskip
\noindent\emph{Crossover horizon for the plug-in estimator.}
By the calibration identity, a static pull $s_0$ pays $(v + c)\,(s_0 - s^*)^2$ over the calibrated optimum, while the plug-in estimate $\hat s$ pays $(v + c)\,(\hat s - s^*)^2$.
If the two-scale noise and conflict estimates are consistent with $\lvert\hat s_t - s^*\rvert \le K\,t^{-1/2}$, the plug-in beats any fixed pull once
\[
  t \;>\; T_{\mathrm{crit}} \;\asymp\; \frac{K^2}{(s_0 - s^*)^2},
\]
the square of the calibration constant over the squared pull mismatch (the risk scale $v+c$ cancels on both sides).
Short streams therefore favor fixed anchoring, and the BCI IV streams ($T \approx 300$ trials) sit on the short side; on the grid the fixed-$\rho$ ablation is $2.2\%$ better at $T = 10^4$, so our runs do not resolve the crossover. The threshold scales quadratically in the prior mismatch $s_0 - s^*$; the plug-in-versus-fixed-shrinkage tradeoff is classical in empirical-Bayes shrinkage \citep{ledoitwolf2004}.

\subsection{Proof of Proposition~\ref{prop:minimax} (minimax safety)}
\label{app:proof-minimax}

\begin{proof}[Proof of Proposition~\ref{prop:minimax}]
For fixed $C$, $R$ is strictly convex in $s$ with unique minimizer $s^*(C) = Dv/(Dv + C)$ (Proposition~\ref{prop:conflict}(ii) applied to the lumped problem; the interpolation identity is Proposition~\ref{prop:conflict}(i)), and $\rho \mapsto s(\rho)$ is a bijection, so $\min_\rho R(\rho, C) = DvC/(Dv + C)$, attained uniquely at $\rho^*(C) = \mu Dv/C$.
This value is strictly increasing in $C$, so the maximin is attained at $C_0$ and equals $DvC_0/(Dv + C_0)$.
For the minimax: for fixed $\rho$ with $s(\rho) > 0$, $R(\rho, C)$ is increasing in $C$, so $\max_{C \le C_0} R(\rho, C) = R(\rho, C_0)$, and minimizing over $\rho$ gives the same value, attained at $\rho^*(C_0)$.
The inequality $DvC/(Dv + C) \le (1 - s)^2 Dv + s^2 C$ for all $s$, which is the computational core of the equality (together with the monotonicity of the value in $C$), is formalized (\texttt{minimax\_lower} in \texttt{Core.lean}).
\end{proof}

\begin{proof}[Proof of Proposition~\ref{prop:minimax}, safety clause]
(i) is Proposition~\ref{prop:conflict}(ii).
(ii) Expand: $R(s) - v = (1-s)^2 v + s^2 C - v = s\bigl(s(v + C) - 2v\bigr)$, which is at most $0$ exactly when $s \le 2v/(v+C) = 2s^*$.
Since $\widehat C \ge C$ implies $\widehat s = v/(v + \widehat C) \le v/(v + C) = s^* < 2s^*$, we get $R(\widehat s) \le v$.
(iii) As $\widehat C \to \infty$, $\rho^* = \widehat V/\widehat C \to 0$ and $s_i = \rho/(\mu_i + \rho) \to 0$ for every direction with $\mu_i > 0$; directions with $\mu_i = 0$ contribute $\mu_i \delta_i^2 = 0$ to the prediction risk.
\end{proof}

\subsection{Proof of Proposition~\ref{prop:falsealarm} (false-alarm calibration)}
\label{app:proof-falseesc}

\begin{proof}[Proof of Proposition~\ref{prop:falsealarm}]
The fast two-scale excess is the causal moving average $Z_t = \sum_{k\ge 0} w_k z_{t-k}$ of a centered sub-exponential martingale-difference sequence, with geometric weights $w_k = \eta_f(1-\eta_f)^k$; its weight energy is the geometric sum
\[
\sum_{k\ge 0} w_k^2 = \frac{\eta_f^2}{1-(1-\eta_f)^2} = \frac{\eta_f}{2-\eta_f} = \frac{1}{m_{\mathrm{eff}}},
\qquad m_{\mathrm{eff}} = \frac{2-\eta_f}{\eta_f},
\]
and $\max_k w_k = \eta_f$.
Freedman's inequality for sub-exponential martingale differences \citep{freedman1975} bounds the weighted sum at each time with predictable variance $V = \nu_z^2/m_{\mathrm{eff}}$ and scale $b = \eta_f b_z$:
\[
\Pr[Z_t > \tau] \;\le\; \exp\Bigl(-\frac{\tau^2}{2\,(V + b\,\tau)}\Bigr),
\]
so the threshold $\tau(\delta) = bL + \sqrt{b^2L^2 + 2VL}$ with $L = \log(T/\delta)$ solves $\tau^2 = 2(V + b\tau)L$ and gives $\Pr[Z_t > \tau(\delta)] \le e^{-L} = \delta/T$.
A union bound over the $T$ time points upgrades this to the sliding (time-uniform) statement $\Pr[\exists\, t \le T: Z_t > \tau(\delta)] \le T \cdot \delta/T = \delta$.
The statistic is the implemented one, with effective sample size $m_{\mathrm{eff}} = (2-\eta_f)/\eta_f \to 1/\eta_f$ for small $\eta_f$, the runtime constant.
The damage of a false escalation is bounded by the cost of running at reduced anchor strength for the escalation holdout: at most $O(v)$ per step for $O(1/\eta_f)$ steps, so with $\delta = 1/T^2$ the total expected false-escalation damage is $o(1)$.
The alternative (drift) is detected with lag $O(h_f + 1/\eta_f)$, where $h_f$ is the fast moving-average horizon.
\end{proof}

\medskip
\noindent\emph{Stationary pinning (effective threshold).}
The variance-floor subtraction is itself a strictly positive threshold: the drift-energy estimator is $\widehat\nu^2_\Sigma=\max\bigl(0,(E_t-\tau_\omega)\omega^2\bigr)$ with $\tau_\omega=\hat c'\,\omega_{\min}/(2-\omega_{\min})$, so $E_t\le\tau_\omega$ implies $\widehat\nu^2_\Sigma=0$ and $\omega_t=\omega_{\min}$; the conflict floor $\tau_\rho=C_{\mathrm{floor}}$ likewise prevents the residual branch from driving the strength below its stationary floor.
Applying the bound of Proposition~\ref{prop:falsealarm} at $\tau=\min(\tau_\omega,\tau_\rho)>0$ gives $\Pr[\omega_t>\omega_{\min}]\le\delta$ for the sliding test under stationarity, upgrading Theorem~\ref{thm:window}'s in-expectation statement to a high-probability pinning guarantee; the bound applies to the implemented EMA directly (Proposition~\ref{prop:falsealarm}).

\begin{remark}[Time uniformity]
The union bound over time points makes the guarantee uniform over the whole stream; the cost is the $\log T$ factor in $\tau(\delta)$.
The measured rate is at or below the nominal level at $\delta = 0.10$ and $0.05$, with a slight exceedance at $\delta = 0.01$ on ill-conditioned spectra (Online Appendix~\ref{app:protocol}).
\end{remark}

\begin{remark}[Mixing and heavy tails]
\label{rem:mixing}
The bound uses only the sub-exponential martingale-difference structure of $z_t$. For a stationary stream with autocorrelations $\rho_k = \operatorname{Corr}(z_t,z_{t-k})$ the EMA variance is
\[
\operatorname{Var}(Z_t) \;\le\; \frac{\sigma_z^2}{m_{\mathrm{eff}}}\Bigl(1 + 2\sum_{k\ge1}|\rho_k|(1-\eta_f)^k\Bigr) \;\le\; \frac{\sigma_z^2}{m_{\mathrm{eff}}}\,(1 + 2S),
\qquad S := \sum_{k\ge1}|\rho_k|,
\]
so the Freedman bound and the time-uniform guarantee of Proposition~\ref{prop:falsealarm} hold verbatim with $\nu_z^2$ replaced by $\nu_z^2(1+2S)$: this covers geometrically mixing streams ($\sum|\rho_k|<\infty$), and, because $|\rho_k|\le\varphi_k^{1/2}$, stationary $\varphi$-mixing streams with summable coefficients. Under heavy tails with $\E|z|^{2+p}<\infty$ ($p>0$) the Gaussian tail is replaced by a Fuk--Nagaev bound \citep{fuk1971}, the threshold scales as $(T/\delta)^{1/(2+p)}$ in place of $\sqrt{\log(T/\delta)}$, and the false-alarm rate remains at most $\delta$ at that larger threshold.
\end{remark}

\subsection{Proof of Proposition~\ref{prop:ocowrap}(i) (projected gradient)}
\label{app:proof-composition}

\begin{proof}[Proof of Proposition~\ref{prop:ocowrap}(i)]
\emph{Step 1 (convexity and subgradient).}
Writing $z_t = U_t^\top x_t$, $w^e_t = U_t^\top \widehat w_t$, and $g^e_t = U_t^\top G_t$, the prediction is $a_t(s) := x_t^\top W_t(s) - y_t = z_t^\top[(1-s)\odot w^e_t + s\odot g^e_t] - y_t$, affine in $s$ on the box; hence $\ell_t = a_t(s)^2$ is convex, with
\[
  \nabla_s \ell_t(s) \;=\; 2\,a_t(s)\,\bigl(z_{t,i}\,(g^e_{t,i} - w^e_{t,i})\bigr)_i .
\]
The exact subgradient identity for the segment family, $(a + b w)^2 + 2b(a + bw)(s - w) \le (a + b s)^2$, is formalized (\texttt{sq\_subgradient} in \texttt{Ogd.lean}), which is the convexity step in the Lean assembly below.

\emph{Step 2 (gradient bound).}
Since $s_i \in [0,1]$ each eigen-coordinate obeys $|W^e_{t,i}| \le \max\{|w^e_{t,i}|, |g^e_{t,i}|\}$, and the Euclidean norm is invariant under $U_t$, so
\[
  \|W_t(s)\|_2^2 \;=\; \|W^e_t(s)\|_2^2 \;\le\; \|\widehat w_t\|_2^2 + \|G_t\|_2^2 \;\le\; 2W_{\max}^2,
\]
and therefore
\[
  |a_t(s)| \;\le\; \|x_t\|_2\, \|W_t(s)\|_2 + |y_t| \;\le\; \sqrt 2\, X W_{\max} + Y .
\]
Using $\|z_t\|_\infty \le \|z_t\|_2 = \|x_t\|_2 \le X$ and $\|g^e_t - w^e_t\|_2 = \|G_t - \widehat w_t\|_2 \le 2W_{\max}$,
\[
  \bigl\|\bigl(z_{t,i}(g^e_{t,i} - w^e_{t,i})\bigr)_i\bigr\|_2
  \;\le\; \|z_t\|_\infty\, \|g^e_t - w^e_t\|_2
  \;\le\; 2\,X W_{\max},
\]
and combining the two factors gives $\|\nabla_s \ell_t\|_2 \le G_B$ with $G_B = 4XW_{\max}(\sqrt 2 XW_{\max} + Y)$.

\emph{Step 3 (projected descent).}
The box $[0,1]^D$ has $\ell_2$-diameter $\sqrt D$, and for any comparison point $s^*$ in the box, $\|s_0 - s^*\|_2 \le \sqrt D$.
Online projected gradient descent on convex losses against an adaptive adversary with step size $\iota = \sqrt{D}/(G_B\sqrt{T})$ incurs regret at most
\[
  \frac{\|s_0 - s^*\|_2^2}{2\iota} + \frac{\iota\, T\, G_B^2}{2}
  \;\le\; G_B\sqrt{DT}
  \;\le\; G_B\sqrt{2DT}
\]
\citep{zinkevich2003}; no probabilistic assumption is used.
The comparison class contains the pure-data profile $\mathbf 0$, the pure-anchor profile $\mathbf 1$, and every fixed scalar-$\rho$ profile $s_\rho$, since all lie in the box; in physical coordinates it is the realized operator sequence $\{U_t\operatorname{diag}(s)U_t^\top\}$, whose basis-invariant sub-class is the isotropic family $\{cI\}$ (constant $s \equiv c$; \texttt{conj\_const\_diagonal}), containing $\mathbf 0$, $\mathbf 1$, and every fixed isotropic ridge. The sublinear rate is confirmed by the diagnostic OGD runs and the $\sqrt D$ premium by the dimension sweep.
The bound is external regret against the realized loss sequence: the anchor's state makes a comparator's counterfactual anchor trajectory differ from the realized one.
\end{proof}

\begin{remark}[Formalized chain]
{\sloppy The 1-dimensional descent chain is mechanized in \texttt{Ogd.lean} (\texttt{sq\_subgradient}, \texttt{projBox\_dist\_sq\_le}, \texttt{ogd\_regret\_box}, and \texttt{ogd\_regret\_segment}), and the full $D$-dimensional squared-loss instantiation is mechanized in \texttt{BoxD.lean} (\texttt{ogd\_regret\_box\_D}, \texttt{sqPredLoss\_subgradient}, \texttt{normSq\_sqPredGrad\_le}, \texttt{ogd\_regret\_sq\_loss\_D}, and \texttt{ogd\_regret\_sq\_loss\_D\_opt} for the tuned bound $G_B\sqrt{DT}\le G_B\sqrt{2DT}$).\par}
The comparator scope is described in Appendix~\ref{app:lean}.
\end{remark}

\subsection{Proof of Proposition~\ref{prop:ocowrap}(ii) (Hedge)}
\label{app:proof-hedge}

\begin{proof}[Proof of Proposition~\ref{prop:ocowrap}(ii)]
\emph{Step 1 (quadratic exponential bound).}
For $x \ge 0$,
\[
  e^{-x} \;\le\; \frac{1}{1+x} \;\le\; 1 - x + x^2,
\]
the first inequality by $e^x \ge 1 + x$, the second by the exact identity $(1 - x + x^2)(1 + x) = 1 + x^3 \ge 1$.
This step is formalized (\texttt{exp\_neg\_le\_one\_sub\_add\_sq} in \texttt{Hedge.lean}).

\emph{Step 2 (potential ratio).}
By the closed form of the weights,
\[
  Z_{t+1}
  \;=\; \sum_i e^{-\eta(L_{i,t} + \ell_{i,t})}
  \;=\; Z_t \sum_i \pi_{i,t}\, e^{-\eta \ell_{i,t}} ,
\]
which is formalized as \texttt{Zfun\_step}.
Applying Step 1 termwise with $\ell_{i,t} \le M$ (so $\ell_{i,t}^2 \le M\,\ell_{i,t}$), and using $\sum_i \pi_{i,t} = 1$ and $1 + u \le e^u$,
\[
  \frac{Z_{t+1}}{Z_t}
  \;\le\; \sum_i \pi_{i,t}\bigl(1 - \eta \ell_{i,t} + \eta^2 \ell_{i,t}^2\bigr)
  \;\le\; 1 - \eta(1 - \eta M)\,m_t
  \;\le\; e^{-\eta(1 - \eta M)\,m_t}
\]
(\texttt{Zfun\_step\_bound}).

\emph{Step 3 (telescope).}
Multiplying over $t < T$ and using $Z_0 = K$ (\texttt{Zfun\_prod\_bound}) gives
$Z_T \le K\, e^{-\eta(1 - \eta M)\sum_t m_t}$, while the $j$-th term of the defining sum gives $Z_T \ge e^{-\eta L_{j,T}}$.
Taking logarithms,
\[
  e^{-\eta L_{j,T}}
  \;\le\; K\, e^{-\eta(1 - \eta M)\sum_t m_t}
  \qquad\Longrightarrow\qquad
  (1 - \eta M)\sum_{t < T} m_t \;\le\; L_{j,T} + \frac{\ln K}{\eta},
\]
which is the stated bound since $1 - \eta M > 0$ (\texttt{hedge\_regret}).
The general $K$-expert mixture-convexity step $(\sum_i \pi_i u_i)^2 \le \sum_i \pi_i u_i^2$ is formalized as \texttt{weighted\_sum\_sq\_le} in \texttt{Core.lean}, and combining it with \texttt{hedge\_regret} yields the fixed-$\eta$ regret bound directly on the realized mixture squared prediction loss (\texttt{mixLoss\_le\_mloss} and \texttt{hedge\_mixture\_regret} in \texttt{Hedge.lean}); the anytime form (the implemented schedule $\varsigma_t = \sqrt{\ln K / S_t}$) is the standard AdaHedge substitution \citep{rooij2014}, a self-confident cumulative-loss schedule in the line of \citet{auer2002}; see also \citet{cesabianchi2006}.
\end{proof}

\medskip
\noindent\emph{Best of both worlds.}
Because \texttt{hedge\_mixture\_regret} (via \texttt{hedge\_regret}) holds for every comparator $j$ in the pool, the frozen anchor and the best adapted window can be chosen simultaneously:
\[
  \sum_{t<T} \ell_t(\mathrm{mix}) \;\le\; \frac{1}{1-\eta M}\Bigl(\min_j L_{j,T} + \frac{\ln K}{\eta}\Bigr),
\]
with the pool containing both: the mixture is never worse than the warm model by more than the regret term, and never worse than the best adapted path by the same term. The SPAR-EG$^+$ pool includes both the frozen anchor and the plug-in method.
\citet{evendar2008} quantify the regret-to-best versus regret-to-average tradeoff.

\subsection{Proof of Proposition~\ref{prop:windup} (conditioning bound)}
\label{app:proof-props}

\begin{proof}[Proof of Proposition~\ref{prop:windup}]
The eigenvalues of $A + n\rho I$ lie in the interval $[\,n\rho,\ n(\mu_{\max} + \rho)\,]$, whose ratio is at most $1 + \mu_{\max}/\rho$.
\end{proof}

The accumulator of Section~\ref{sec:theory} never enters the covariance form: the anchored solve is a batch information-form solve, so Proposition~\ref{prop:windup} holds at every window, including windows where the covariance recursion of leaky RLS loses rank and diverges \citep{sayed2008}.
The contrast is measured in Section~\ref{sec:experiments}: the leaky covariance recursion diverges in $135/135$ main-grid and $20/20$ stress-suite runs at window $100 < D = 128$, while the batch information-form solve of the same objective is stable at every $\lambda$, and no SPAR method diverged in any configuration.

\medskip
\noindent\emph{Eigengap context (Davis--Kahan).}
Let $U, U'$ be the eigenbases of $\bar A$ and $\bar A' = \bar A + E$ with $\|E\|_{\mathrm{op}} \le \epsilon$, and let $\delta$ be the minimum gap between eigenvalues of $\bar A$ involved in a per-direction decision.
Then
\[
  \bigl\|(I - U'U'^\top)\,U_i\bigr\| \;\le\; \frac{\epsilon}{\delta}
\]
by the Davis--Kahan $\sin\Theta$ theorem in the two-subspace form \citep{davis1970,wedin1972}, so the per-direction refresh with a stale basis errs by at most $O(\epsilon/\delta)\,\|W - G\|$ per refresh cycle.
A refresh cycle performed in a basis rotated by more than the pull displacement applies the intended per-direction weights to the wrong coordinates, at a cost proportional to the rotation angle times $\|W - G\|$.
Three regimes follow: with static, well-separated spectra (the ill-conditioned grid) the bound is small and the per-direction estimator is stable and wins pre-drift; with flat spectra ($\delta \to 0$) the eigenbasis is arbitrary and per-direction state is noise; with rotating spectra (covariance drift) the basis drifts within the estimator's memory and per-direction state scrambles.
The scalar-strength method aggregates over directions and is invariant to basis rotation, so it is the safer default.

\subsection{Spectral retention (Theorem~\ref{thm:rmt})}
\label{app:rmt}

This subsection separates the deterministic-equivalent proof from its numerical validation.
Throughout, $\zeta = D\omega$ is the aspect ratio held fixed in the proportional limit, $H$ is the population spectral measure of $\Sigma$, and $\rho_{\mathrm{eff}}$ is the finite-window renormalized anchor strength.
The risk coefficient of Theorem~\ref{thm:window} is written $c' = \sigma^2 d_{\mathrm{eff}}(\rho)$ throughout; it is distinct from $\zeta$ and from the cost branches $C^\Sigma_{\mathrm{ref}}, C^\Sigma_{\mathrm{resid}}, C^\Sigma_{\mathrm{floor}}$ of Algorithm~\ref{alg:spar-pull}.
The scalar geometric weights are $\omega_s = \omega(1-\omega)^s$; $W$, $w^*$, and $\widehat w$ denote the anchored solve, the target parameter, and the data solve as in the main text.

\subsubsection{The deterministic-equivalent system}
\label{app:rmt-system}

For the exponentially weighted accumulator with geometric weights $\omega(1-\omega)^s$ and features $x_s = \Sigma^{1/2} z_s$ ($z_s$ sub-Gaussian with independent entries), the separable-covariance deterministic equivalent \citep{silverstein1995,rubio2011,hachem2007} gives, for the resolvent $R(\rho) = (\bar A + \rho I)^{-1}$ and $T(-\rho) = (\kappa_1 \Sigma + \rho I)^{-1}$, the two-equation fixed-point system
\begin{equation}
\label{eq:rmt-system}
\theta_1 = \int \frac{\tau}{\kappa_1 \tau + \rho}\, dH(\tau),
\qquad
\kappa_1 = \sum_{s \ge 0} \frac{\omega_s}{1 + D \omega_s \theta_1}.
\end{equation}
In the streaming limit ($D \to \infty$, $\omega \to 0$, $D \omega = \zeta$ fixed) the weight sum integrates in closed form.
Substituting $y = (1-\omega)^s$ and using $\sum_s \omega (1-\omega)^s f((1-\omega)^s) \to \int_0^1 f(y)\,dy$,
\begin{equation}
\label{eq:rmt-kappa}
\kappa_1(\theta_1)
= \int_0^1 \frac{dy}{1 + \zeta \theta_1 y}
= \frac{\ln(1 + \zeta \theta_1)}{\zeta \theta_1}.
\end{equation}
The hypotheses used here are the ones stated in Theorem~\ref{thm:rmt}: independent sub-Gaussian entries, a population spectrum whose nonzero part is bounded away from zero ($\tau_{\min}^+ > 0$), and the proportional regime.
The atom at zero is allowed: Corollary~\ref{cor:rmt-active} below shows that exact null directions drop out of the prediction risk, so the system collapses to the active subspace verbatim.

\subsubsection{The collapse lemma}
\label{app:rmt-collapse}

\begin{lemma}[Collapse]
\label{lem:rmt-collapse}
Let $q = 1 - \bar s$ be the average data retention. Under \eqref{eq:rmt-system}--\eqref{eq:rmt-kappa}, $\kappa_1 \theta_1 + \bar s = 1$, and the system collapses to the scalar fixed point
\begin{equation}
\label{eq:rmt-collapse}
q = \int \frac{\tau}{\tau + \rho_{\mathrm{eff}}}\, dH(\tau),
\qquad
\rho_{\mathrm{eff}} = \rho\,\frac{e^{\zeta q} - 1}{\zeta q}.
\end{equation}
\end{lemma}

\begin{proof}
Per population eigenvalue the retention and the pull sum to one, so $\kappa_1 \theta_1 + \bar s = 1$ and $q = \kappa_1 \theta_1$.
Substituting \eqref{eq:rmt-kappa} gives $q = \ln(1 + \zeta \theta_1)/\zeta$, hence $e^{\zeta q} = 1 + \zeta \theta_1$.
Therefore $\kappa_1 = \zeta q/(e^{\zeta q} - 1)$ and
$\rho_{\mathrm{eff}} = \rho/\kappa_1 = \rho (e^{\zeta q} - 1)/(\zeta q)$.
The collapse identity is formalized as \texttt{de\_collapse} in \texttt{Core.lean}; the probabilistic convergence of the deterministic equivalent remains pen-and-paper.
Existence and uniqueness in $(0,1)$: the right side of \eqref{eq:rmt-collapse} is continuous and strictly decreasing in $q$ (the growth factor $(e^{\zeta q}-1)/(\zeta q) \ge 1$ is strictly increasing, and each term $\tau/(\tau + \rho_{\mathrm{eff}})$ decreases in $\rho_{\mathrm{eff}}$), positive at $q \downarrow 0$ because $\tau_{\min} > 0$, and strictly below $1$ at $q = 1$.
\end{proof}

\subsubsection{Variance functional (Theorem~\ref{thm:rmt}(i))}
\label{app:rmt-variance}

With an oracle anchor ($\delta = 0$), the one-step excess prediction risk is its variance component
\begin{equation}
\label{eq:rmt-variance}
\sigma^2\,\E\!\left[\operatorname{tr}\!\left(\Sigma R\, \bar A^{(2)} R\right)\right],
\qquad
\bar A^{(2)} = \sum_s \omega_s^2 x_s x_s^\top .
\end{equation}
Perturb the weights by $\omega_s(\varepsilon) = \omega_s(1 + \varepsilon \omega_s)$, so that $\omega_s'(0) = \omega_s^2$ and $\bar A^{(2)}$ is the first-order term.
Then
\begin{equation}
\label{eq:rmt-derivative}
\E\!\left[\operatorname{tr}\!\left(\Sigma R\, \bar A^{(2)} R\right)\right]
= -\left.\frac{d}{d\varepsilon}\E\operatorname{tr}\!\left(\Sigma R_\varepsilon\right)\right|_{\varepsilon=0}
= -D\,\dot\theta_1,
\end{equation}
where the dot is $d/d\varepsilon$ at $\varepsilon = 0$.
This is the step that produces the amplification denominator.

Differentiate \eqref{eq:rmt-system} in $\varepsilon$.
Writing $J = \int \bigl(\tau/(\tau + \rho_{\mathrm{eff}})\bigr)^2 dH(\tau)$, the first equation gives
\begin{equation}
\label{eq:rmt-theta-dot}
\dot\theta_1 = -\frac{\dot\kappa_1}{\kappa_1^2}\,J,
\end{equation}
because $\kappa_1\tau + \rho = \kappa_1(\tau + \rho_{\mathrm{eff}})$.
For the second equation, the derivative of $\omega_s(\varepsilon)$ and the derivative of the denominator combine:
\begin{equation}
\label{eq:rmt-kappa-dot}
\dot\kappa_1
= \sum_s \left[
\frac{\omega_s^2}{1 + D \omega_s \theta_1}
- \frac{\omega_s \bigl(D \omega_s^2 \theta_1 + D \omega_s \dot\theta_1\bigr)}{(1 + D \omega_s \theta_1)^2}
\right]
= K_2\,\bigl(1 - D\dot\theta_1\bigr),
\end{equation}
with
\begin{equation}
\label{eq:rmt-K2}
K_2 = \sum_s \frac{\omega_s^2}{(1 + D \omega_s \theta_1)^2}.
\end{equation}
Substituting \eqref{eq:rmt-kappa-dot} into \eqref{eq:rmt-theta-dot} and solving the linear equation for $\dot\theta_1$,
\begin{equation}
\label{eq:rmt-upsilon-def}
-D\dot\theta_1
= \frac{\Upsilon}{1 - \Upsilon},
\qquad
\Upsilon := \frac{D K_2 J}{\kappa_1^2}.
\end{equation}
The denominator $1-\Upsilon$ is exactly the Neumann series of the derivative of the fixed-point map; it is positive whenever $\Upsilon < 1$, which is the regime covered by the theorem.

It remains to evaluate $K_2$ in closed form.
With $y = (1-\omega)^s$, $\sum_s \omega^2 (1-\omega)^{2s} f((1-\omega)^s) \to \omega\int_0^1 y f(y)\,dy$, so
\begin{equation}
\label{eq:rmt-K2-integral}
K_2
= \omega\int_0^1 \frac{y}{(1 + \zeta \theta_1 y)^2}\,dy
= \frac{\omega}{(\zeta \theta_1)^2}
\left[\ln(1 + \zeta \theta_1) - \frac{\zeta \theta_1}{1 + \zeta \theta_1}\right].
\end{equation}
Using $e^{\zeta q} = 1 + \zeta \theta_1$ and $\kappa_1^2 = (\zeta q)^2/(\zeta \theta_1)^2$,
\begin{equation}
\label{eq:rmt-K2-collapse}
\frac{D K_2}{\kappa_1^2}
= \frac{\zeta}{(\zeta q)^2}\Bigl[\zeta q - \bigl(1 - e^{-\zeta q}\bigr)\Bigr]
= \frac{1}{q}\left(1 - \frac{1 - e^{-\zeta q}}{\zeta q}\right).
\end{equation}
Combining \eqref{eq:rmt-upsilon-def} and \eqref{eq:rmt-K2-collapse} yields
\begin{equation}
\label{eq:rmt-upsilon}
\Upsilon
= \frac{1}{q}\left(1 - \frac{1 - e^{-\zeta q}}{\zeta q}\right)
\int \left(\frac{\tau}{\tau + \rho_{\mathrm{eff}}}\right)^{2} dH(\tau),
\end{equation}
which is the variance functional of Theorem~\ref{thm:rmt}(i).
The $\zeta \to 0$ limit is consistent: expanding $e^{-\zeta q} = 1 - \zeta q + \tfrac12(\zeta q)^2 + O(\zeta^3)$ gives $\Upsilon = \tfrac12\zeta J + O(\zeta^2)$, so $\sigma^2\Upsilon/(1-\Upsilon) \to \tfrac12\sigma^2\omega\,d_{\mathrm{eff}}(\rho)$, the mean-field variance of Theorem~\ref{thm:window}.

\subsubsection{Pull sensitivities (Theorem~\ref{thm:rmt}(ii))}
\label{app:rmt-sensitivity}

Implicit differentiation of \eqref{eq:rmt-collapse} in $\ln\rho$ gives
\begin{equation}
\label{eq:rmt-sensitivity}
\rho\,\frac{\partial \bar s}{\partial \rho}
= \frac{\mathcal{I}}{1 + \zeta\,\phi(\zeta q)\,\mathcal{I}},
\qquad
\mathcal{I} = \int \tilde s\,(1 - \tilde s)\, dH(\tau),
\end{equation}
with $\tilde s(\tau) = \rho_{\mathrm{eff}}/(\tau + \rho_{\mathrm{eff}})$ and $\phi(u) = (1 - e^{-u})^{-1} - u^{-1} \in (0,1)$.
The anisotropy bound follows from the variance decomposition
\begin{equation}
\label{eq:rmt-anisotropy}
\mathcal{I} = \bar s\,(1 - \bar s) - \operatorname{Var}_H(\tilde s).
\end{equation}
Spectral spread pushes directions into the data-dominated bulk ($\tilde s \approx 0$) or the anchor-pinned tail ($\tilde s \approx 1$), both insensitive, so the closed-loop sensitivity
$S_{\omega\rho} = \mathcal{I}/(q + \zeta\phi(\zeta q)\mathcal{I})$ stays below one on every spectrum.
This is the analytic explanation of the measured coupling margins of Theorem~\ref{thm:closedloop}.

\subsubsection{Conflict bias (Theorem~\ref{thm:rmt}(iii))}
\label{app:rmt-conflict}

Part (i) is the variance component.
The general excess risk adds the bias carried by an anchor offset $\delta = G - w^*$.
Writing the anchored solve as $W = (\bar A + \rho I)^{-1}(\bar B + \rho G)$ and taking expectations over the noise alone, the deterministic part of the deviation is the resolvent action $\rho R(\rho)\delta$ with $R(\rho) = (\bar A + \rho I)^{-1}$.
For \emph{out-of-sample} prediction risk at $x_{\mathrm{new}} \sim (0,\Sigma)$ the conflict contribution is
\begin{equation}
\label{eq:rmt-conflict-quadratic}
\delta^\top \rho^2 R(\rho)\,\Sigma\,R(\rho)\,\delta .
\end{equation}
The in-sample empirical risk replaces $\Sigma$ by $\bar A$ and gives $\mu_i(\rho/(\mu_i + \rho))^2\delta_i^2$ in the classical limit; we state the out-of-sample form, which is the risk the experiments estimate.

In the proportional streaming limit with $\delta$ held fixed, define the direction-weighted conflict measure $H_\delta$ (an atom of mass $\delta_i^2$ at each population direction $\tau_i$, total mass $\|\delta\|^2$).
The deterministic equivalent of \eqref{eq:rmt-conflict-quadratic} is
\begin{equation}
\label{eq:rmt-conflict-de}
\rho_{\mathrm{eff}}^2 \int \frac{\tau}{(\tau + \rho_{\mathrm{eff}})^2}\, dH_\delta(\tau).
\end{equation}
It remains to justify the renormalization by $(1-\Upsilon)^{-1}$.
The perturbation $\varepsilon$ enters \eqref{eq:rmt-conflict-quadratic} only through the scalar $\kappa_1(\varepsilon)$, because the resolvent is a function of $\kappa_1(\varepsilon)\Sigma + \rho I$.
Expanding the resolvent around $\varepsilon = 0$ produces the same scalar Neumann series $1 + \Upsilon + \Upsilon^2 + \cdots$ that produced \eqref{eq:rmt-upsilon-def}, with the same $\Upsilon$ from \eqref{eq:rmt-upsilon}; the conflict quadratic form multiplies that series term by term.
Hence
\begin{equation}
\label{eq:rmt-risk}
R_{\mathrm{pred}}(\rho, \zeta) \;\to\;
\frac{1}{1 - \Upsilon}\left[\sigma^2\,\Upsilon
+ \rho_{\mathrm{eff}}^2 \int \frac{\tau}{(\tau + \rho_{\mathrm{eff}})^2}\, dH_\delta(\tau)\right],
\end{equation}
which is Theorem~\ref{thm:rmt}(iii).
The oracle anchor ($H_\delta \equiv 0$) reduces \eqref{eq:rmt-risk} to Part (i), and the classical limit $\zeta \to 0$ gives $\rho_{\mathrm{eff}} \to \rho$, $\Upsilon \to 0$, recovering Proposition~\ref{prop:conflict}'s per-direction bias.
The derivation is a deterministic-equivalent expansion rather than a finite-$D$ theorem; the rank-deficient case is covered by the corollary below (Appendix~\ref{app:rmt-validation}).

\subsubsection{Active-subspace reduction (rank-deficient spectra)}
\label{app:rmt-active}

The full-rank hypothesis in Theorem~\ref{thm:rmt} is not needed.
Exact null directions carry no prediction risk, and the deterministic equivalent restricts to the range of the population covariance.

\begin{corollary}[Active-subspace deterministic equivalent]
\label{cor:rmt-active}
Let the population spectral measure split as $H = (1-p)\,\delta_0 + p\,H^+$ with $p \in (0,1]$ and $\operatorname{supp}(H^+) \subset [\tau^+_{\min}, \tau^+_{\max}]$, $\tau^+_{\min} > 0$.
Then the deterministic-equivalent system \eqref{eq:rmt-system}--\eqref{eq:rmt-upsilon}, the variance functional $\Upsilon$, and the prediction risk \eqref{eq:rmt-risk}, all evaluated at $(H, \zeta)$, coincide with those evaluated at $(H^+, \zeta^+)$ with the reduced aspect ratio $\zeta^+ = p\,\zeta$.
\end{corollary}

\begin{proof}
The features are $x_s = \Sigma^{1/2} z_s$, so $x_s$ vanishes identically on $\ker\Sigma$ and the accumulator $\bar A$ has an exact zero block there.
In the out-of-sample risk $\E[(x_{\mathrm{new}}^\top(W-w^*))^2] = (W-w^*)^\top\Sigma(W-w^*)$ every error component in $\ker\Sigma$ is multiplied by $\Sigma$, so it contributes nothing.
Algebraically, the atom at zero drops out of the Stieltjes integrals:
\[
\theta_1 = \int \frac{\tau}{\kappa_1\tau+\rho}\,dH = p\,\theta_1^+,
\qquad \theta_1^+ := \int \frac{\tau}{\kappa_1\tau+\rho}\,dH^+,
\]
while the weight sum \eqref{eq:rmt-kappa} depends on $\theta_1$ only through the product
$\zeta\,\theta_1 = (D\omega)(p\,\theta_1^+) = (pD)\,\omega\,\theta_1^+ = \zeta^+ \theta_1^+$.
Hence \eqref{eq:rmt-system} is the $H^+$ system at aspect ratio $\zeta^+$, so $\kappa_1$ and therefore $\rho_{\mathrm{eff}} = \rho/\kappa_1$ are unchanged.
Retentions scale as $q = \kappa_1\theta_1 = p\,q^+$ with the active retention $q^+ := \kappa_1\theta_1^+ = \int \tau/(\tau+\rho_{\mathrm{eff}})\,dH^+$, which is the fixed point \eqref{eq:rmt-collapse} with $\rho_{\mathrm{eff}} = \rho (e^{\zeta^+ q^+}-1)/(\zeta^+ q^+)$.
For the variance functional, $J = \int (\tau/(\tau+\rho_{\mathrm{eff}}))^2\,dH = p\,J^+$ and $D\,K_2/\kappa_1^2$ is a function of $\zeta\theta_1 = \zeta^+\theta_1^+$ alone, so
$\Upsilon = (D K_2/\kappa_1^2)\,J = (D^+ K_2/\kappa_1^2)\,J^+ = \Upsilon^+$.
The conflict integral in \eqref{eq:rmt-risk} carries the weight $\tau/(\tau+\rho_{\mathrm{eff}})^2$, which vanishes on the null atom, so it is unchanged as well.
\end{proof}

On a rank-deficient population the same deterministic equivalent holds on the active subspace, of dimension $D^+ = pD$ and aspect ratio $\zeta^+ = p\zeta$, with the hypothesis reduced to $\tau_{\min}^+ > 0$ on the nonzero part.
The direction-weighted conflict measure $H_\delta$ is likewise supported on $\operatorname{range}(\Sigma)$ in the prediction risk; the $\tau$-weighting is what carries this invariance, and the unweighted trace functional does not.

\subsubsection{The coupling corollary is local}
\label{app:rmt-coupling}

At an interior fixed point the product of cross-sensitivities is invariant under monotone coordinate changes, since $\partial \psi_1(\omega^*)/\partial \psi_2(\rho^*)\cdot \partial \psi_2(\rho^*)/\partial \psi_1(\omega^*) = |(\partial \omega^*/\partial \rho)(\partial \rho^*/\partial \omega)|$; we therefore evaluate in $(\ln \omega, \ln \rho)$ without loss of generality.
(i) The window target $\omega^* \propto (4\nu^2_\Sigma / c')^{1/3}$ with $c' = \sigma^2 d_{\mathrm{eff}}(\rho) \propto q^2$ gives $|\partial \ln \omega^*/\partial \ln \rho| = \tfrac{2}{3}\,\mathcal I / [q(1 + \zeta\phi(\zeta q)\mathcal I)] \le \tfrac{2}{3}\,\bar s/q$, using $\mathcal I \le \bar s(1-\bar s) = q\bar s$.
(ii) The strength target $\rho^*(\omega) = \bar\mu\widehat V(\omega)/\widehat C$ is analyzed by which branch of $\widehat C$ is active:
(a) when $C^\Sigma_{\mathrm{floor}}$ is active, $\widehat C$ is independent of $\omega$, and $|\partial \ln \rho^*/\partial \ln \omega| \le 2/(2-\omega)$;
(b) when $C^\Sigma_{\mathrm{resid}} = E_t$ is active, the drift-active floor $\rho_{\min}^{\mathrm{drift}}$ pins $\rho$, so $L_{\rho\omega} = 0$ and the system is triangular (Theorem~\ref{thm:closedloop}(ii));
(c) when $C^\Sigma_{\mathrm{ref}}$ is active at stationarity ($\nu = 0$), $\widehat w$ is unbiased for $w^*$ in the mean-field limit, so the reference branch is independent of $\omega$ to leading order in $1/n$ and $L_{\rho\omega} = 0$; away from stationarity it inherits the same $(\omega/(2-\omega))$ scaling as (a), with the constant absorbed into $\bar s$.
Branches (b) and (c) carry explicit finite-$n$ corrections. In branch (b) the drift-active floor fixes $\rho$ exactly, so $L_{\rho\omega} = 0$; the residual dependence is only the branch-selection margin, bounded by the $O(1/n)$ fluctuation of the residual excess $E_t$.
In branch (c), under the quasi-static estimator block of Theorem~\ref{thm:closedloop}, the computed quadratic form $R = (G_{\mathrm{ref}}-\widehat w)^\top \bar A_t (G_{\mathrm{ref}}-\widehat w)$ is $\omega$-independent, so the runtime estimate $\widehat C_{\mathrm{ref}}(\omega) = R - \widehat V(\omega)$ has the single explicit $\omega$-dependence of the debiasing term $\widehat V(\omega) = \widehat\sigma^2 d_{\mathrm{act}}\,\omega/(2-\omega)$; differentiating $\rho^* = \bar\mu\widehat V/\widehat C_{\mathrm{ref}}$,
\[
\bigl|L_{\rho\omega}^{(c)}\bigr|
= \frac{2}{2-\omega}\Bigl(1 + \frac{\widehat V}{|\widehat C_{\mathrm{ref}}|}\Bigr),
\]
the leading coefficient plus the measurable debiasing ratio $\widehat V/|\widehat C_{\mathrm{ref}}| = \rho^*/\bar\mu$ at the fixed point, in place of an unbound $O(1/n)$.
The measured coupling product is below $6\times10^{-2}$ across all $20$ diagnostic runs, with one quantity pinned in three of four configurations.
The coupling product in log-coordinates is therefore at most
\begin{equation}
\label{eq:rmt-coupling}
L_{\ln \omega, \ln\rho}\,L_{\ln\rho, \ln \omega}
\;\le\; \frac{2}{3q}\,\mathcal I \cdot \frac{2}{2-\omega}
\;=\; \frac{4\,\mathcal I}{3q(2-\omega)}
\;\le\; \frac{4}{3(2-\omega)}\,\bar s
\;<\; \bar s
\;<\; 1
\end{equation}
for all $\bar s \in (0,1)$ and $\omega \in (0, 2/3)$, with strict improvement from spectral anisotropy ($\operatorname{Var}_H(\tilde s) > 0$ reduces $\mathcal I$ below $q\bar s$).
The bound \eqref{eq:rmt-coupling} is uniform on the parameter box $B = [\omega_{\min},\omega_{\max}]\times[\rho_{\mathrm{lo}},\rho_{\mathrm{hi}}]$: it uses only $\mathcal I \le q\bar s$ and $\bar s \le \bar s_{\max} < 1$, both of which hold at every $\rho$, so it is not tied to the fixed point.
\begin{corollary}[Global contraction on the parameter box]
\label{cor:global}
The mean-field coupled map $F(\omega,\rho) = (\omega^*(\rho),\rho^*(\omega))$ is a contraction on the whole box $B$ in the weighted max-norm $\max\{|\ln\omega|,\, r\,|\ln\rho|\}$, with the fixed weight $r = \sqrt{a_{\max}/b_{\max}}$ and modulus $\gamma_{\mathrm{glob}} = \sqrt{a_{\max}b_{\max}} \le \sqrt{\tfrac{4}{3}\,\bar s_{\max}/(2-\omega_{\max})} < 1$, where $a_{\max} = \sup_B |\partial\ln\omega^*/\partial\ln\rho|$ and $b_{\max} = \sup_B |\partial\ln\rho^*/\partial\ln\omega|$ are finite by \eqref{eq:rmt-coupling}.
The clip is a nonexpansive projection onto $B$, so $B$ is forward-invariant and the clipped recursion contracts on $B$ as well.
By Banach's theorem $F$ has a unique fixed point and every trajectory in $B$ converges to it geometrically at rate $\gamma_{\mathrm{glob}}$: the basin of attraction is the entire box $B$, not a neighbourhood of the fixed point.
The piecewise definition of $\widehat C$ preserves this: $F$ is smooth with the same bound on each branch region, and a pointwise maximum of finitely many Lipschitz maps is Lipschitz with the largest constant.
\end{corollary}

\subsubsection{Numerical validation}
\label{app:rmt-validation}

The closed forms above were checked against direct simulation, separately from the proof.
\emph{Fixed points.}
At $45$ operating points (three spectra $\times$ five nominal windows $1/(1-\lambda) \in \{100,\dots,2000\}$ $\times$ three anchor strengths $\rho \in \{10^{-3}, 10^{-2}, 10^{-1}\}$), \eqref{eq:rmt-collapse} matches the Monte Carlo stationary risk to within $4.0\%$ (median $1.0\%$) across all three spectra, and within $1.8\%$ on the rank-deficient spectrum.
The rank-deficient grid also verifies the active-subspace invariance of Corollary~\ref{cor:rmt-active}: the full-spectrum and reduced-system functionals agree to machine precision ($|\Upsilon-\Upsilon^+| = 0$ at every point).
\emph{Sensitivities.}
Equation~\eqref{eq:rmt-sensitivity} agrees with numerical differentiation of the fixed point to $10^{-6}$ relative error across all nine spectrum--operating-point combinations.
\emph{Conflict increment.}
Paired Monte Carlo (same seeds, conflict-anchor total minus oracle total) matches \eqref{eq:rmt-risk} within $6.2\%$ (median $1.3\%$) for a random conflict direction, and within $10.0\%$ (median $1.6\%$) for a top-eigendirection conflict, across all three spectra including the rank-deficient one; the direction weighting by $\tau/(\tau+\rho_{\mathrm{eff}})^2$ is what keeps the conflict measure correct there.
The random-orientation rank-deficient case is covered by Corollary~\ref{cor:rmt-active}: restricting the deterministic equivalent to $\operatorname{range}(\Sigma)$ removes the null-space mismatch, after which the paired Monte Carlo matches \eqref{eq:rmt-risk} within $6.2\%$ at every point.
\emph{Operating points.}
At the method's own converged control pair the deterministic equivalent matches the estimator's Monte Carlo risk to within $9.8\%$ (isotropic), $12.7\%$ (decay3), and $1.9\%$ (rank-deficient).
\emph{Finite-$D$ concentration.}
Scoring the risk with the population covariance, the out-of-sample quantity Theorem~\ref{thm:rmt} is stated for, the relative discrepancy $|R_{\mathrm{MC}}-R_{\mathrm{DE}}|/R_{\mathrm{DE}}$ at fixed $\zeta$ is within the Monte Carlo standard error at every $D \in \{64,128,256,512\}$ over $\zeta \in \{1,2\}$ and $\rho \in \{10^{-2},10^{-1}\}$: for $D \ge 128$ the maximum relative bias is $1.1\%$ and the median $|\text{bias}|/\text{s.e.}$ is $0.66$, with no systematic growth in $D$, consistent with an $O(1/D)$ deterministic-equivalent error.
The fixed-point grid above scores the risk with the in-sample empirical covariance, which adds an $O(D/T)$ in-sample term ($\approx 1.3\%$ at $D=128$, $T=10^4$) on top of this finite-$D$ error.
All rows of the diagnostic grids, full-rank and rank-deficient, support the theorem.

The convergence step is cited \citep{silverstein1995,rubio2011,hachem2007}; the nearest prior analysis is the large-system exponentially windowed RLS study of \citet{xiao2005}, without anchoring, per-direction pulls, refresh, or self-calibrating estimators, and the unweighted-batch ridge asymptotics \citep{hastie2022,pafka2004} do not carry the finite-window renormalization.

\subsection{Closed-loop stability of the two estimators (Theorem~\ref{thm:closedloop})}

\emph{Part (i): the coupled recursions.}
The recursion is the quasi-static two-time-scale model: the estimator block runs every $k$ steps and is smoothed with horizons $1/\eta_C, 1/\eta_\nu$, both slow relative to the accumulator mixing time $1/\omega$ and the fast EMA horizon $h_f$; finite-$k$ lag is an additive perturbation absorbed into $\xi_\omega, \xi_\rho$ \citep{borkar2008,benveniste1990,kushner2003}.
The window estimator maintains an exponential average of the debiased drift-energy estimate, $\widehat\nu^2_{m+1} = (1-\eta_\nu)\widehat\nu^2_m + \eta_\nu\,\widehat\nu^2_\Sigma(\rho_m)$ with gain $\eta_\nu$ (the estimator of Theorem~\ref{thm:window}); at a frozen target the cubed readout $u = \omega^3 = 4\widehat\nu^2/c'$ is affine in the averaged quantity, and the floor-isolation identity together with the frozen-target contraction for $0 < \eta_\nu < 2$ are formalized (\texttt{debiased\_dial\_estimator\_eq}, \texttt{debiased\_dial\_contraction}, \texttt{debiased\_dial\_tendsto}).
Since $u$ is affine in $\widehat\nu^2$ at frozen $c'$, and $c' = (1-\bar s)^2\sigma^2D$ is Lipschitz in $\rho$ through $\bar s$, the error $e_\omega(m) = |u_m - (\omega^*(\rho_m))^3|$ satisfies
\[
e_\omega(m{+}1) \;\le\; (1-\eta_\nu)\,e_\omega(m) + \eta_\nu L_{\omega\rho}\, e_\rho(m) + \xi_\omega(m),
\]
where $L_{\omega\rho}$ absorbs the readout and $c'$ sensitivities and $\xi_\omega$ collects the drift-energy estimation noise, bounded by the same sub-exponential concentration as Proposition~\ref{prop:falsealarm}.
The clip onto $[\omega_{\min}, \omega_{\max}]$ is a projection onto a convex set, which preserves the one-sided error contraction; a pinned estimator is the triangular case of part~(ii).
The strength estimator tracks the plug-in target $\rho^*(\omega) = \bar\mu\widehat V/\widehat C$ of Proposition~\ref{prop:conflict} through clipping and shrinkage; $\widehat V$ and $\widehat C$ are Lipschitz in the window coordinate on the whole parameter box, with no spectral-gap assumption, once the active dimension is taken as the smooth effective dimension $\mathsf d(\bar A) = \mathrm{tr}(\bar A(\bar A+\rho_{\mathrm{ridge}}I)^{-1}) = \sum_i \mu_i/(\mu_i+\rho_{\mathrm{ridge}})$, the $\tau \to 0$ restriction of the integer count $d_{\mathrm{act}}$; each summand lies in $[0,1)$ and is $1/\rho_{\mathrm{ridge}}$-Lipschitz in its eigenvalue (\texttt{smooth\_dim\_lipschitz} and \texttt{smooth\_effdim\_lipschitz\_sum} in \texttt{Core.lean}), so $\mathsf d(\omega)$ and hence $\widehat V$ are Lipschitz everywhere on the box, unlike the integer count which jumps at eigenvalue crossings; through $\bar\mu(\omega)$, $\mathsf d(\omega)$, and the prediction-unit reference conflict $(G_{\mathrm{ref}}-\widehat w)^\top\bar A(G_{\mathrm{ref}}-\widehat w)$, the maps are Lipschitz, giving
\[
e_\rho(m{+}1) \;\le\; \eta_C L_{\rho\omega}\, e_\omega(m) + (1-\eta_C)\,e_\rho(m) + \xi_\rho(m).
\]
Setting $a = \eta_\nu L_{\omega\rho}$, $b = \eta_C L_{\rho\omega}$, the two recursions form the coupled system of \texttt{Coupled.lean}.

\emph{The weighted-norm contraction.}
With the midpoint weight $r := \tfrac12(a/\eta_\nu + \eta_C/b)$, the interval $(a/\eta_\nu,\ \eta_C/b)$ being nonempty exactly when $ab < \eta_\nu\eta_C$, i.e.\ $L_{\omega\rho}L_{\rho\omega} < 1$, the joint rate is
\[
\gamma_{\mathrm{cl}} \;=\; \max\bigl\{(1-\eta_\nu) + a/r,\ \ (1-\eta_C) + r\,b\bigr\} \;<\; 1,
\]
and the error pair contracts in the weighted max-norm $\max\{e_\omega,\, r\,e_\rho\}$: the one-step bound, geometric decay, and tracking radius $\xi/(1-\gamma)$ are formalized as \texttt{joint\_step}, \texttt{joint\_geometric}, and \texttt{joint\_tracking}, with the rate lemma \texttt{joint\_rate\_lt\_one} and the weight lemmas \texttt{joint\_bounds}, \texttt{joint\_weight\_lo}, \texttt{joint\_weight\_hi}.

\emph{Part (ii): the drift-mode pin.}
When the detector's drift-active floor pins $\rho$, $e_\rho \equiv 0$ and the window error contracts at the free estimator's own rate $1-\eta_\nu$ (\texttt{joint\_geometric\_pinned}); the joint-rate specialization is \texttt{joint\_geometric\_drift\_mode}.
The complementary one-way case (the strength target insensitive to the window) is the triangular specialization \texttt{joint\_geometric\_triangular}, with noisy tracking radius \texttt{joint\_tracking\_triangular}.

\emph{Part (iii): the fluctuation bound.}
The contracted recursion of part (i) is the scalar Lyapunov inequality whose stationary second moment is bounded in Proposition~\ref{prop:fluct}; the same weighted max-norm $V = \max\{|e_\omega|, r|e_\rho|\}$ bounds the cross moment, $|\E[e_\omega e_\rho]| \le \E[V^2]/r \le q_*/(r(1-\gamma_{\mathrm{cl}})^2)$, so the full stationary second-moment matrix, including its off-diagonal cross-covariance, is controlled entrywise by the scalar bound. The weighted max-norm route gives the entrywise bounds that part (iii) requires; entrywise domination of nonnegative error vectors is exactly the form the cross-moment bound uses.

\begin{proposition}[Invariant measure in the co-moving frame]
\label{prop:ergodic}
Assume $(x_t,\varepsilon_t)$ is i.i.d.\ (or stationary ergodic) with $\E\|x_t\|^2 \le \bar\Sigma$, $\varepsilon_t$ sub-Gaussian, and constant drift $\nu$.
Work in the co-moving frame $\tilde B_t := B_t - A_t w^*_t$ and $\tilde G_t := G_t - w^*_t$, and let $\Phi_m$ be the $k$-step skeleton at $t = mk$.
\begin{enumerate}
\item[(i)] $\Phi_m$ is a Feller Markov chain on the product of the compact parameter box with the trace-bounded (tight) accumulator set, and admits an invariant probability measure $\pi$.
\item[(ii)] \sloppy If the mean-field estimator map of Theorem~\ref{thm:closedloop} is contracting ($\gamma_{\mathrm{cl}}<1$) and the full state-to-estimator Jacobian satisfies the small-gain condition $\varrho(J_{\mathrm{state}\to\mathrm{estimator}})\,\varrho(J_{\mathrm{estimator}\to\mathrm{state}}) < 1-\gamma_{\mathrm{cl}}$, then $\pi$ is unique and $\Phi_m$ is geometrically ergodic in $W_1$.
\end{enumerate}
The small-gain condition controls the coupling directly: when the two copies' estimators differ, the accumulator difference carries the cross term $-(\omega^{(1)}-\omega^{(2)})A^{(2)}$, which the condition bounds.
\end{proposition}
\begin{proof}
(i) The co-moving frame removes the linear growth: $A_t$ is tight by the Foster--Lyapunov drift $\E[A_{t+1}\mid\mathcal F_t]\preceq(1-\omega_{\min})A_t+\bar\Sigma$, the centered statistics $\tilde B_t,\tilde G_t$ are bounded through the pull interpolation, and the estimators live in a compact box by construction.
The one-step kernel is $k$-periodic because the anchor refresh and estimator updates occur every $k$ steps, so the Krylov--Bogoliubov argument \citep{meyn2009} is applied to the time-homogeneous $k$-step skeleton $\Phi_m$.
The hard gate $\mathbf 1\{E_t>0\}$ does not break the Feller property: when $\varepsilon_t$ has a continuous density (Gaussian or bounded with a density), the boundary event $\{E_t=0\}$ has conditional probability zero, so the transition expectation of any bounded continuous function is continuous in the state; no artificial smoothing of the gate is needed.
Tightness plus this Feller property yields an invariant probability measure \citep{meyn2009}.
(ii) Linearizing the joint error dynamics around the invariant set gives the feedback system state $\to$ estimator $\to$ state. The small-gain condition makes its spectral radius strictly below one in a weighted norm, and combined with the Foster--Lyapunov drift and the support condition this gives geometric ergodicity and uniqueness \citep[Ch.~18]{meyn2009}.
Uniqueness of the full process requires the small-gain condition; the idealized mean-field estimator process of Theorem~\ref{thm:closedloop} is unique under $\gamma_{\mathrm{cl}}<1$.
\end{proof}

\begin{proposition}[Fluctuation bound]
\label{prop:fluct}
Let the coupled estimator deviations obey the weighted max-norm recursion of part (i), $V(m{+}1) \le \gamma_{\mathrm{cl}}\,V(m) + \|\xi(m)\|_r$, where $\xi(m)$ is any perturbation sequence with uniformly bounded second moment $\sup_m \E\|\xi(m)\|_r^2 \le q_*$.
Then the stationary second moment satisfies $\limsup_m \E[V(m)^2] \le q_*\,/\,(1 - \gamma_{\mathrm{cl}})^2$; in particular each estimator fluctuates within RMS distance $\sqrt{q_*}\,/\,(1 - \gamma_{\mathrm{cl}})$ of its target.
No zero-mean, martingale, or independence assumption is used.
\end{proposition}
\begin{proof}
Young's inequality $(a + b)^2 \le (1 + \epsilon)a^2 + (1 + 1/\epsilon)\,b^2$ with $\epsilon = 1/\gamma_{\mathrm{cl}} - 1$ makes $(1 + \epsilon)\,\gamma_{\mathrm{cl}}^2 = \gamma_{\mathrm{cl}}$ and $(1 + 1/\epsilon) = 1/(1 - \gamma_{\mathrm{cl}})$, so
\[
  \E[V(m{+}1)^2] \;\le\; \gamma_{\mathrm{cl}}\,\E[V(m)^2] \;+\; \E\|\xi(m)\|_r^2\,/\,(1 - \gamma_{\mathrm{cl}}) \;\le\; \gamma_{\mathrm{cl}}\,\E[V(m)^2] + q_*/(1 - \gamma_{\mathrm{cl}}).
\]
Iterating from a bounded initial second moment and summing the geometric series $\sum_j \gamma_{\mathrm{cl}}^j = 1/(1 - \gamma_{\mathrm{cl}})$ gives $\E[V(m)^2] \le \gamma_{\mathrm{cl}}^m\,\E[V(0)^2] + q_*/(1 - \gamma_{\mathrm{cl}})^2$.
\end{proof}
No zero-mean or martingale assumption is used: Young's inequality bounds the cross term $2ab$ pathwise. If $\xi$ were a signed martingale difference and one worked with the signed recursion before taking norms, the cross term would vanish and the denominator would improve from $(1-\gamma_{\mathrm{cl}})^2$ to $1-\gamma_{\mathrm{cl}}^2$; the norm route is deliberately more general and covers the nonnegative triangle-inequality remainder as well.

Together, these results close the analysis of the coupled loop: the single-estimator bounds correspond to the frozen-coupling regime, the product condition is the coupling budget, and the gate enforces it exactly where the targets move fastest.
The deterministic coupled stationary law is the unique fixed point of the composition $\Phi = \rho^*\circ\omega^*$: substituting the window target $\omega^*(\rho) = (4\nu^2_\Sigma/\sigma^2 d_{\mathrm{eff}}(\rho))^{1/3}$ into the strength target $\rho^*(\omega) = \bar\mu\widehat V(\omega)/\widehat C(\omega)$ gives the scalar equation
\[
\rho \;=\; \Phi(\rho), \qquad
\Phi(\rho) \;=\; \frac{\bar\mu\,\widehat V(\omega^*(\rho))}{\widehat C(\omega^*(\rho))},
\]
a contraction with modulus the coupling product $L_{\omega\rho}L_{\rho\omega} < 1$; hence $\Phi$ has a unique fixed point, reached from any start by the iteration $\rho_{m+1} = \Phi(\rho_m)$, and the law admits the convergent telescoping series $\rho^* = \rho_0 + \sum_{m\ge0}(\rho_{m+1}-\rho_m)$ with geometric remainder $O((L_{\omega\rho}L_{\rho\omega})^m)$.
No single elementary closed form is available: $\Phi$ composes a cube root of the rational $d_{\mathrm{eff}}(\rho)$ with the rational strength map, so the fixed-point equation is transcendental and the convergent series is the closed-form handle.
The stochastic stationary law's mean is this fixed point and its second moment is bounded by $q_*/(1-\gamma_{\mathrm{cl}})^2$ (Proposition~\ref{prop:fluct}); the state-dependent innovation $\xi$ makes the density non-elementary, so the moment recursion is the exact form available.

\subsection{Spectral Invariance of the Scalar Method}
\label{app:invariance}

\begin{proposition}[Scalar-method rotation equivariance]
\label{prop:invariance}
For any orthogonal $Q$, the scalar-$\rho$ anchored solve is equivariant under rotating the stream, $x'_t = Qx_t$ with anchor $G' = QG$: the rotated solve is $W' = QW$, every prediction is invariant ($x'^\top_t W' = x^\top_t W$), and the spectral refresh commutes with the rotation.
\end{proposition}
\begin{proof}
The rotated accumulators are $A' = QAQ^\top$ and $B' = QB$, and $QAQ^\top + n\rho I = Q(A + n\rho I)Q^\top$, so
$W' = (QAQ^\top + n\rho I)^{-1}(QB + n\rho\,QG) = Q\,(A + n\rho I)^{-1}(B + n\rho G) = QW$.
Predictions are inner products of rotated vectors against rotated solutions.
The eigenvalues of $\bar A' = Q\bar A Q^\top$ coincide with those of $\bar A$, so the pulls $s_i = \rho/(\mu_i + \rho)$ are unchanged; with the rotated eigenbasis $U' = QU$ the refresh step maps to the rotated refresh, so the equilibrium of Theorem~\ref{thm:cadence} is unchanged as well.
The trace and determinant invariance of the conjugated accumulator are formalized as \texttt{conj\_trace} and \texttt{conj\_det} in \texttt{Core.lean}.
\end{proof}
This equivariance has two practical consequences. First, the scalar method's per-step solve requires no eigendecomposition (it uses only $A$, $B$, $G$), so a covariance rotation mid-stream degrades it through the conditioning of $A + n\rho I$, which Proposition~\ref{prop:windup} bounds, and through the frozen anchor's misalignment until the next refresh. Second, the per-direction method's refresh aligns to the current dominant directions only at each cycle, so between cycles it applies its pulls in a stale basis, at the Davis--Kahan-bounded cost above. The scalar method wins all nine covariance-rotation configurations while the per-direction method is the fragile one (Sections~\ref{sec:experiments} and~\ref{sec:limitations}).

\section{Supporting Theory}
\label{app:support}

This section collects the supporting results behind the anchored solve and the effective-memory-horizon analysis, with short proofs and references to the standard literature.

\subsection{The Anchored Solve as Exact MAP}
\label{app:support-map}

For a forgetting factor $\alpha \in (0,1]$ define the weighted statistics
\[
  A_t = \sum_{i=1}^t \alpha^{t-i}\,\phi_i\phi_i^\top, \qquad
  B_t = \sum_{i=1}^t \alpha^{t-i}\,\phi_i y_i^\top, \qquad
  M_t = \sum_{i=1}^t \alpha^{t-i},
\]
and the anchored objective with prior mean $G$, prior precision $\tau_0 > 0$, and ridge $\lambda \ge 0$,
\[
  J_t(W) \;=\; \sum_{i=1}^t \alpha^{t-i}\,\|\phi_i^\top W - y_i\|_F^2
  \;+\; (\tau_0 + \lambda)\,\|W - G\|_F^2 .
\]
The anchor penalty is undiscounted because $G$ is a fixed prior, not streaming data.

\begin{proposition}[Anchored ridge as exact MAP]
\label{prop:app-map}
Assume $y_i = \phi_i^\top W^\star + \varepsilon_i$ with $\varepsilon_i \sim \mathcal N(0, \sigma^2 I)$ and prior $W \sim \mathcal N(G, \tau^{-1} I)$ with $\tau = (\tau_0 + \lambda)/\sigma^2$.
Then the posterior mean of $W$ given $(\phi_i, y_i)_{i \le t}$ equals
\[
  W_t^\star \;=\; \bigl(A_t + (\tau_0 + \lambda) I\bigr)^{-1}\bigl(B_t + (\tau_0 + \lambda) G\bigr).
\]
\end{proposition}
\begin{proof}
Completing the square in $J_t$: the normal equations are $(A_t + (\tau_0 + \lambda)I)\,W = B_t + (\tau_0 + \lambda)G$ with positive-definite Hessian, and the Gaussian posterior mean solves the same system.
\end{proof}

The identity is the standard ridge-as-MAP correspondence \citep{hoerl1970,rasmussen2006}, with regularization toward a nonzero prior mean as in L2-SP \citep{li2018l2sp}.
Two algebraic facts are used in the text: the solve is ordinary ridge on the augmented design $[\Phi;\ \sqrt{\tau_0 + \lambda}\,I]$ with pseudo-targets $[Y;\ \sqrt{\tau_0 + \lambda}\,G]$, so standard conditioning statements transfer mechanically and the $\tau_0 \to \infty$ limit is exactly the frozen warm solution (the zero-shot endpoint of the WiSE-FT interpolation \citep{wortsman2022wiseft}); and the count-scaled form $W = (A_t + M_t\rho I)^{-1}(B_t + M_t \rho G)$ used by SPAR matches \hyperref[prop:app-map]{the proposition} with the anchor weight proportional to the effective count, whereas an anchor weight that does not scale with the accumulated count dilutes as the stream grows, biasing strength sweeps.

\subsection{Effective-memory-horizon Stability and the Rank Obstruction}
\label{app:support-emh}

\begin{proposition}[Unanchored stability needs excitation over the window]
\label{prop:app-emh}
Let $\alpha = 1 - 1/m$ with integer $m \ge 2$, and suppose every window $\mathcal W_t = \{t - m + 1, \dots, t\}$ satisfies the excitation condition
\[
  \lambda_{\min}\Bigl(\sum_{i \in \mathcal W_t} \phi_i \phi_i^\top\Bigr) \;\ge\; s^2 \;>\; 0 .
\]
If $\|\phi_i\| \le L$ for all $i$, then
\[
  \lambda_{\min}(A_t) \;\ge\; e^{-1} s^2,
  \qquad
  \|A_t^{-1}\| \;\le\; e/s^2,
  \qquad
  \kappa_2(A_t) \;\le\; e\,m\,L^2/s^2 .
\]
\end{proposition}
\begin{proof}
For samples in the window, $\alpha^{t-i} \ge \alpha^{m-1} = (1 - 1/m)^{m-1} = \bigl(1 + \tfrac{1}{m-1}\bigr)^{-(m-1)} \ge e^{-1}$, since $(1 + \tfrac1n)^n \le e$ for all $n \ge 1$ and taking reciprocals gives the required lower bound.
Hence $A_t \succeq e^{-1} \sum_{i \in \mathcal W_t} \phi_i\phi_i^\top \succeq e^{-1} s^2 I$.
For the upper bound, $\lambda_{\max}(A_t) \le L^2 \sum_{k \ge 0} \alpha^k = mL^2$ (with $A_t$ truncated at $t$), which gives the condition-number bound.
\end{proof}

\begin{corollary}[Rank obstruction]
\label{cor:app-rank}
The window sum $\sum_{i \in \mathcal W_t} \phi_i\phi_i^\top$ has rank at most $\min(m, D)$.
Hence if $m < D$ the excitation condition can hold only on a subspace of dimension $m$, and in the generic case the unanchored recursion is ill-posed when $m \ll D$.
This is the covariance windup of aggressive forgetting, whose measured boundary is reported in Section~\ref{sec:experiments}.
\end{corollary}

\begin{corollary}[Anchoring removes the excitation requirement]
\label{cor:app-anchor-excites}
For any $\tau_0 + \lambda > 0$,
$A_t + (\tau_0 + \lambda)I \succeq (\tau_0 + \lambda)I$ regardless of excitation, so
$\|(A_t + (\tau_0 + \lambda)I)^{-1}\| \le 1/(\tau_0 + \lambda)$ and
$\kappa_2 \le 1 + mL^2/(\tau_0 + \lambda)$: the anchor replaces the persistence-of-excitation assumption by a conditioning budget.
\end{corollary}

The practical design rule is $m = 1/\omega \gtrsim D$ (the effective window at least the dimension), or an anchor strength of at least $10^{-2}$ when $m < D$.
Proposition~\ref{prop:app-emh} and its corollaries are the stability statement behind Proposition~\ref{prop:windup} of the main text: the anchored solve never relies on excitation, and the forgetting method that violates the rule is the one that diverges.

\section{The Lean 4 Artifact}
\label{app:lean}

The algebraic core and the no-regret chain of Propositions~\ref{prop:conflict}--\ref{prop:ocowrap} are formalized in Lean 4 \citep{demoura2021} with the Mathlib library \citep{mathlib2020} (\texttt{theory/lean/SparTheorems/} in the artifact).
This section maps the paper's results to the artifact, states the formalization boundary, records the axiom dependencies, and gives the build recipe.

\subsection{Result-to-declaration Map}
Table~\ref{tab:lean-map} summarizes the correspondence between the paper's results and the Lean 4 declarations.

{\scriptsize
\setlength{\tabcolsep}{3pt}
\begin{longtable}{@{}p{1.02in} p{1.12in} p{1.9in} l@{}}
\caption{Map from the paper's results to the Lean artifact.
Probabilistic statements, random-matrix convergence, Davis--Kahan bounds, and wiring of the empirical clipped maps remain pen-and-paper; every algebraic, calculus, regret, and contraction result with a named Lean declaration above is formalized; the remaining rows are pen-and-paper and appear in the list below.}
\label{tab:lean-map}\\
\toprule
paper result & content & Lean declaration(s) & file \\
\midrule
\endfirsthead
\multicolumn{4}{@{}l}{\itshape continued from the previous page}\\
\toprule
paper result & content & Lean declaration(s) & file \\
\midrule
\endhead
\bottomrule
\endlastfoot
Equation~\eqref{eq:pull} & pull identity ($\gamma = 1$ and general) & \texttt{pull\_identity}, \texttt{pull\_identity\_general} & Core.lean \\
Equation~\eqref{eq:pull} (eigenbasis) & diagonal decoupling; coordinatewise anchored solve & \texttt{eigenbasis\_solve\_coord}, \texttt{eigenbasis\_pull\_identity} & Core.lean \\
Proposition~\ref{prop:conflict} & risk interpolation; dominance; calibration identity; optimal pull; multiplicative inversion at the optimum & \texttt{convex\_interp\_sq}, \texttt{convex\_interp\_sq\_le}, \texttt{dominance}, \texttt{dominance\_strict}, \texttt{quadratic\_calibration}, \texttt{quadratic\_calibration\_delta}, \texttt{optimal\_pull\_min}, \texttt{optimal\_pull\_attained}, \texttt{pull\_inverse} & Core.lean \\
Proposition~\ref{prop:minimax} & minimax value (core inequality); safety excess identity; multiplicative safety threshold $s(v+c)\le 2v$ & \texttt{minimax\_lower}, \texttt{quadratic\_calibration}, \texttt{safety\_threshold} & Core.lean \\
Lemma~\ref{lem:zerolag} & lag-invariance algebra & \texttt{lag\_invariance} & Core.lean \\
Theorem~\ref{thm:cadence} & refresh rate identity; finite-cadence fixed point; geometric closed form; contraction; correlated variance identities and the lower endpoint $\Psi\ge(1-s)^2$ & \texttt{refresh\_rate\_identity}, \texttt{refresh\_fixed\_point}, \texttt{refresh\_closed\_form}, \texttt{refresh\_contraction\_rate}, \texttt{refresh\_anchor\_variance}, \texttt{refresh\_variance\_form}, \texttt{refresh\_variance\_lt}, \texttt{refresh\_variance\_lb} & Core.lean \\
Theorem~\ref{thm:cadence} (moving frame) & moving-frame reduction to the affine recursion; midpoint cycle assembly; Gauss sum & \texttt{moving\_frame\_reduction}, \texttt{cycle\_midpoint\_average}, \texttt{cycle\_lag\_assembly}, \texttt{sum\_range\_half} & Core.lean \\
Theorem~\ref{thm:window} & cubic optimum; exact mean-field risk derivatives, strict convexity on $(0,2)$, and unique global optimality; estimator fixed-point dichotomy ($0 < \beta < 2$ contraction and floor isolation) & \texttt{window\_optimum}, \texttt{window\_second\_deriv\_pos}, \texttt{windowRisk\_hasDerivAt}, \texttt{windowRisk\_deriv\_hasDerivAt}, \texttt{windowRisk\_strictConvexOn}, \texttt{windowRisk\_global\_min}, \texttt{naive\_dial\_no\_fixed\_point}, \texttt{debiased\_dial\_closed\_form}, \texttt{debiased\_dial\_contraction}, \texttt{debiased\_dial\_tendsto}, \texttt{debiased\_dial\_estimator\_eq} & Ogd.lean, Core.lean \\
Theorem~\ref{thm:window} (variance coefficient) & exact effective dimension; plug-in gap $D(1-\bar s)^2 \le d_{\mathrm{eff}}$ & \texttt{deff\_sub\_plugin}, \texttt{deff\_sub\_plugin\_nonneg} & Core.lean \\
Proposition~\ref{prop:invariance} & rotation equivariance; refresh commutation & \texttt{conj\_trace}, \texttt{conj\_det} (spectral invariance under orthogonal conjugation) & Core.lean \\
Proposition~\ref{prop:falsealarm} & concentration step & --- (pen-and-paper, \citep{freedman1975}) & --- \\
 & subgradient identity & \texttt{sq\_subgradient} & Ogd.lean \\
 & box-projection contraction (1-dim) & \texttt{projBox\_dist\_sq\_le} & Ogd.lean \\
Proposition~\ref{prop:ocowrap}(i) & Zinkevich step (1-dim); basis-invariant comparators & \texttt{ogd\_regret\_box}, \texttt{ogd\_regret\_segment}, \texttt{conj\_const\_diagonal} & Ogd.lean, Core.lean \\
 & $D$-dimensional box Zinkevich and squared-loss instantiation ($G_B\sqrt{2DT}$, adaptive losses) & \texttt{projBoxD\_dist\_sq\_le}, \texttt{sum\_expand\_sq}, \texttt{ogd\_step\_D}, \texttt{ogd\_regret\_box\_D}, \texttt{sqPredLoss\_subgradient}, \texttt{normSq\_sqPredGrad\_le}, \texttt{ogd\_regret\_sq\_loss\_D}, \texttt{ogd\_regret\_sq\_loss\_D\_opt} & BoxD.lean \\
Proposition~\ref{prop:ocowrap}(ii) & quadratic exponential bound & \texttt{exp\_neg\_le\_one\_sub\_add\_sq} & Hedge.lean \\
 & $K$-expert mixture convexity & \texttt{weighted\_sum\_sq\_sub}, \texttt{weighted\_sum\_sq\_le} & Core.lean \\
 & potential step, ratio, and product bounds & \texttt{Zfun\_step}, \texttt{Zfun\_step\_bound}, \texttt{Zfun\_prod\_bound} & Hedge.lean \\
 & fixed-$\eta$ Hedge regret on realized mixture prediction loss & \texttt{hedge\_regret}, \texttt{mixLoss\_le\_mloss}, \texttt{hedge\_mixture\_regret} & Hedge.lean \\
Proposition~\ref{prop:windup} & matrix pull identity; anchored resolvent; floor nonsingularity; eigenbasis conditioning ratio (general PSD reduction on paper); effective-dimension trace & \texttt{pull\_matrix\_sum\_one}, \texttt{diagonal\_resolvent}, \texttt{anchored\_det\_pos}, \texttt{anchored\_cond\_bound}, \texttt{resolvent\_trace\_effdim} & Core.lean \\
Theorem~\ref{thm:closedloop} & joint $2\times2$ contraction; geometric decay; tracking radius; drift-mode, one-way (noiseless and noisy), and pinned-estimator specializations; component Lipschitz bounds & \texttt{joint\_step}, \texttt{joint\_geometric}, \texttt{joint\_tracking}, \texttt{joint\_tracking\_triangular}, \texttt{joint\_bounds}, \texttt{joint\_weight\_lo}, \texttt{joint\_weight\_hi}, \texttt{joint\_rate\_lt\_one}, \texttt{joint\_geometric\_drift\_mode}, \texttt{joint\_geometric\_triangular}, \texttt{joint\_geometric\_pinned}, \texttt{smooth\_dim\_lipschitz}, \texttt{smooth\_effdim\_lipschitz\_sum}, \texttt{branch\_max\_lipschitz} & Coupled.lean, Core.lean \\
Propositions~\ref{prop:ergodic},~\ref{prop:fluct} & invariant measure; second-moment (Lyapunov) bound & --- (pen-and-paper, \citealp{meyn2009}) & --- \\
\midrule
Corollary~\ref{cor:mixed} & mixed-drift master law & --- (pen-and-paper, Appendix~\ref{app:proofs}) & --- \\
Theorem~\ref{thm:rmt} & spectral retention and streaming risk; collapse/renormalization identities & \texttt{de\_collapse} (convergence pen-and-paper) & Core.lean \\
Props.~\ref{prop:app-map},~\ref{prop:app-emh}; Cors.~\ref{cor:app-rank},~\ref{cor:app-anchor-excites} & exact MAP; excitation/rank; anchoring removes the excitation requirement & --- (pen-and-paper, Appendix~\ref{app:support}) & --- \\
\end{longtable}}

\subsection{Coverage and Formalization Boundary}
The artifact contains 111 theorem and lemma declarations, all proved: 49 in \texttt{Core.lean}, 16 in \texttt{Ogd.lean}, 12 in \texttt{Hedge.lean}, 17 in \texttt{BoxD.lean}, and 17 in \texttt{Coupled.lean}, together with 9 concrete non-vacuity witnesses in \texttt{NonVacuity.lean} verifying joint satisfiability of the main theorem hypotheses.
The algebraic core covers the pull identity, eigenbasis decoupling, risk-interpolation and calibration identities, minimax core, lag-invariance step, and the finite-cadence refresh chain (rate identity, fixed point, geometric closed form, contraction, variance identities, $\Psi \ge (1-s)^2$, moving-frame reduction, and cycle-average assembly).
The online-learning and window modules cover the subgradient and box-projection steps, the 1-dimensional and full $D$-dimensional squared-loss Zinkevich regret bounds ($G_B\sqrt{2DT}$), general $K$-expert mixture convexity and fixed-$\eta$ Hedge mixture-loss regret, basis-invariant comparators, the cubic window optimum, and the exact mean-field risk's two derivatives, strict convexity on $(0,2)$, and unique global optimality, together with the exact effective-dimension decomposition of the window coefficient and the contracted debiased dial ($0 < \beta < 2$).
The matrix and stability modules cover spectral invariance under orthogonal conjugation, eigenbasis anchored-system identities (pull-weight sum, resolvent, floor nonsingularity, eigenvalue conditioning ratio, effective-dimension trace), smooth effective-dimension Lipschitzness, branch continuity, deterministic-equivalent collapse, and the coupled contraction chain.

The probabilistic concentration step, the deterministic-equivalent convergence of Theorem~\ref{thm:rmt}, the Davis--Kahan bound, the anytime AdaHedge learning-rate schedule $\varsigma_t = \sqrt{\ln K / S_t}$, the reduction from general symmetric positive-semidefinite matrices to the eigenbasis identities, and the wiring showing SPAR's clipped empirical maps satisfy the coupled linear error bounds of \texttt{Coupled.lean} remain pen-and-paper.
Table~\ref{tab:lean-map} and \texttt{SCOPE.md} mark each row as formalized or pen-and-paper.

No declaration introduces a custom axiom: the sources contain no \texttt{axiom}, \texttt{sorry}, or \texttt{unsafe} construct, and the headline theorems' axiom dependencies are logged in \texttt{theory/lean/axioms.log}, regenerated by \texttt{theory/lean/print\_axioms.sh}, all resting only on \texttt{propext}, \texttt{Classical.choice}, and \texttt{Quot.sound}.

\subsection{Build Recipe}
{\sloppy The project ships its own \texttt{theory/lean/lakefile.toml}, \texttt{lean-toolchain}, \texttt{lake-manifest.json}, and root import module, and builds directly from \texttt{theory/lean/} with \texttt{lake exe cache get \&\& lake build} (and \texttt{lake env lean SparTheorems/TestAx.lean} for axioms). Equivalently, \texttt{theory/lean/build\_ogd.sh} copies the sources and configuration into a standalone lake project (default \texttt{\$HOME/SparTheorems}) and builds:\par}
\begin{quote}
\ttfamily
bash theory/lean/build\_ogd.sh\\
bash theory/lean/print\_axioms.sh
\end{quote}
On first use the script resolves Mathlib with \texttt{lake update} and fetches the pre-built cache with \texttt{lake exe cache get}, so no Mathlib compilation is required.

%% file: appendix_online.tex
\section{Extended Experimental Protocol}
\label{app:protocol}

\subsection{Factorial Grid}
\emph{Definition.}
The grid of Section~\ref{sec:experiments} is generated by \texttt{scripts/synthetic\_factorial.py}.
The main suite is the full factorial
\[
  \underbrace{\{\text{iso},\ \text{decay3},\ \text{rankdef}\}}_{\text{spectra}}
  \times
  \underbrace{\{\text{abrupt},\ \text{covrot},\ \text{both}\}}_{\text{drift}}
  \times
  \underbrace{\{\text{oracle},\ \text{warm},\ \text{adversarial}\}}_{\text{prior}}
\]
with Gaussian noise, giving 27 configurations, plus an extra suite of four: loss of persistent excitation on the decay3 spectrum, oscillating drift on the isotropic spectrum, a stationary control (no drift), and a covariance-rotation configuration with heavy-tailed $t_3$ noise, for 31 configurations in total.
Spectra: isotropic ($\mu_i \equiv 1$), decay3 (log-spaced from $1$ to $10^{-3}$, condition number $10^3$), and rank-deficient (a fixed fraction of zero eigenvalues).
Drifts: an abrupt jump of the target, a rotation of the input covariance, both combined, none, the death of excitation in a subspace, and an oscillating target.
Priors: oracle (the pre-drift target), warm (a perturbed warm-up solution), and adversarial (a perturbation independent of the target).
All configurations use $D = 128$, $T = 10^4$ steps with a warm-up of $T_w = 2000$, noise variance $\sigma^2 = 0.01$, and 5 seeds per configuration; every method runs predict-then-update.

\begin{figure}[htbp]
\centering
\includegraphics[width=0.55\textwidth]{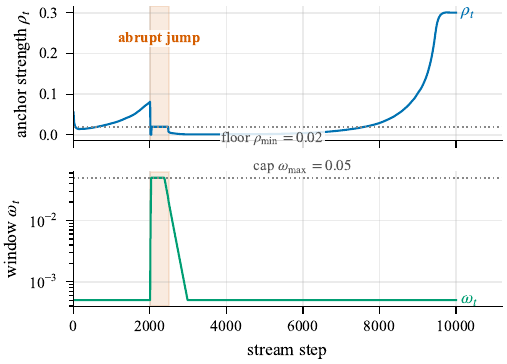}
\caption{Both estimators self-calibrate on a stream with an abrupt jump ($D = 128$, decay-3, seed 0): the anchor strength $\rho_t$ falls on conflict (Proposition~\ref{prop:conflict}) and the window $\omega_t$ opens on drift (Theorem~\ref{thm:window}), both recovering; dotted lines mark the $\rho$ floor and the $\omega$ cap.}
\label{fig:cliffdials}
\end{figure}

\begin{table}[htbp]
\centering
\small
\caption{Stress suite (4 configurations $\times$ 5 seeds): mean and worst overall MSE, one fixed setting per method. The suite is deliberately outside the main grid's operating envelope (loss of persistent excitation, oscillating drift, a stationary control, heavy-tailed rotation noise); the true-noise Kalman is an oracle comparator, and the covariance-form recursion diverges in all 20 runs.}
\label{tab:stress}
\begin{tabular}{lcc}
\toprule
method & mean MSE & worst MSE \\
\midrule
Kalman, true noise ($q = 10^{-4}$) & $1.29$ & $6.26$ \\
SPAR-EG$^+$ & $1.60$ & $7.83$ \\
SPAR-EG & $1.60$ & $7.77$ \\
leaky RLS, information form ($\lambda = 0.99$) & $1.61$ & $7.70$ \\
SPAR-Pull & $1.66$ & $8.10$ \\
SGD ($\mathrm{lr} = 0.01$) & $2.07$ & $10.03$ \\
Adam \citep{kingma2015adam} ($\mathrm{lr} = 0.01$) & $2.82$ & $13.32$ \\
WiSE-FT ($\alpha = 1$) & $3.62$ & $17.39$ \\
anchor-free solve & $4.08$ & $19.70$ \\
frozen warm solution & $5.91$ & $28.83$ \\
leaky RLS, covariance form ($\lambda = 0.99$) & \multicolumn{2}{c}{diverged (20/20)} \\
\bottomrule
\end{tabular}
\end{table}

On the stress suite SPAR-EG$^+$ is the best oracle-free method in mean MSE ($1.60$), $23\%$ below SGD and $43\%$ below Adam; the true-noise Kalman leads it by $23\%$, consistent with the suite residing outside the self-calibration's design envelope (short streams and stationary controls favor fixed shrinkage, Section~\ref{sec:limitations}).

\begin{table}[htbp]
\centering
\scriptsize
\setlength{\tabcolsep}{3pt}
\caption{Marginal mean MSE over the 27 main configurations (each row averages 9 configurations $\times$ 5 seeds), addressing the aggregate-only view of Table~\ref{tab:grid}. The two SPAR methods agree within $0.03$ on every slice; the Kalman and VFF leaders are prior-independent and lose the covariance-rotation slice. The \texttt{abrupt} and \texttt{both} rows coincide to three decimals because both streams carry the same abrupt target jump, and the added covariance rotation shifts the drift-axis mean by less than $10^{-3}$.}
\label{tab:marginal}
\begin{tabular}{llccccc}
\toprule
axis & value & SPAR-Pull & SPAR-EG$^+$ & KF-slow & VFF & leaky-info \\
\midrule
spectrum & isotropic & $0.673$ & $0.644$ & $0.507$ & $0.562$ & $0.636$ \\
 & decay-3 & $0.114$ & $0.115$ & $0.065$ & $0.120$ & $0.257$ \\
 & rank-deficient & $0.131$ & $0.132$ & $0.071$ & $0.119$ & $0.134$ \\
\midrule
drift & abrupt & $0.454$ & $0.440$ & $0.310$ & $0.387$ & $0.487$ \\
 & covariance rotation & $0.011$ & $0.011$ & $0.023$ & $0.027$ & $0.054$ \\
 & both & $0.454$ & $0.440$ & $0.310$ & $0.387$ & $0.487$ \\
\midrule
prior & oracle & $0.301$ & $0.294$ & $0.215$ & $0.267$ & $0.342$ \\
 & warm & $0.305$ & $0.295$ & $0.215$ & $0.267$ & $0.342$ \\
 & adversarial & $0.313$ & $0.303$ & $0.215$ & $0.267$ & $0.342$ \\
\bottomrule
\end{tabular}
\end{table}

\emph{Methods.}
Each configuration runs the baseline families below in a common streaming protocol, plus the slow-horizon estimated-$R$ Kalman, the variable forgetting factor baselines, the two OGD variants (scalar segment and $D$-dimensional profile), SPAR-EG$^+$, and the causally-tuned Kalman filters appended post hoc. The families are the frozen warm solution, an anchor-free solve, leaky RLS at $\lambda \in \{0.999, 0.997, 0.995, 0.99\}$ in both the covariance-form recursion and the batch information-form solve, directional forgetting \citep{kulhavy1984,cao2000} at the matched grid $\lambda_f \in \{0.999, 0.997, 0.995, 0.99, 0.98, 0.95, 0.90\}$, AROW \citep{crammer2009arow} at $r \in \{0.01, 0.05, 0.1, 0.3, 0.5\}$, random-walk Kalman filters \citep{kalman1960,sayed1994} at $q \in \{10^{-6}, 10^{-5}, 10^{-4}\}$ with the true noise level plus the estimated-$R$ variants at both horizons, SGD at $\mathrm{lr} \in \{0.003, 0.01\}$ and Adam \citep{kingma2015adam} at $\mathrm{lr} \in \{0.0003, 0.001, 0.003, 0.01\}$, variable forgetting factor RLS at two floors under both the single-step and the nominal-length calibrations, and the WiSE-FT $\alpha$-envelope (11 values). The SPAR methods are scalar, diagonal, two-estimator, and expert-simplex.
Sweeps are reported per value; the table reports each family's best setting under both the per-configuration rule (oracle-tuned baselines, the conservative treatment) and the global single-setting rule, and SPAR methods use the same configuration in every cell.

\emph{Baseline verification.}
Before any comparison, the \texttt{verify} step of the pipeline checks the implementation identities numerically: directional forgetting at $\lambda_f = 1$ coincides with RLS, the Kalman filter at $q = 0, R = 1$ coincides with RLS, and the WiSE-FT endpoints coincide with the frozen and fully adapted solves (maximum absolute deviation of order $10^{-15}$).
The same step checks the algebraic identities of Section~\ref{sec:theory} symbolically with SymPy \citep{meurer2017sympy}, and the step exits nonzero on any failure.

\emph{Estimated-noise Kalman.}
The grid includes two estimated-$R$ Kalman filters.
The fast variant updates $R$ with a 100-step moving average of squared residuals; a fast horizon attributes part of the drift-induced residual to measurement noise, which suppresses the filter's gain precisely when adaptation is needed, and it collapses to $1.09/5.36$.
The slow variant uses a $20{,}000$-step horizon (matching the SPAR noise baseline) and attains $0.215/0.93$ (best single $q = 10^{-4}$), within $5.4\%$ of the true-noise filter ($0.204/0.86$) and ahead of SPAR-Pull overall ($0.306/1.22$), though SPAR-Pull leads it on all nine covariance-rotation configurations ($0.011$ vs.\ $0.023$ per configuration mean).
Together, the two rows show that the apparent gap was caused by the estimation horizon, not by access to the true noise level.

\emph{Rank-deficiency note.}
The prediction-unit estimator never forms $\mathrm{tr}(\bar A^{-1})$: $d_{\mathrm{act}}$ counts the eigenvalues of $\bar A$ above the relative threshold $10^{-8}\max_j\mu_j(\bar A)$, so dead directions contribute neither variance nor conflict and feature rescaling leaves $\rho$ invariant (Appendix~\ref{app:hyper}).
The floor/cap pair bounds the resulting strength in all $135$ main-grid and $20$ stress-suite runs; no SPAR method diverged on any spectrum.

\begin{figure}[htbp]
\centering
\includegraphics[width=3.1in]{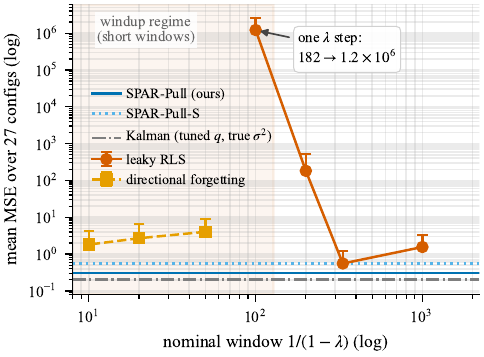}
\caption{Mean MSE over 27 configurations versus nominal window $1/(1-\lambda)$ (log--log; $+1$ s.d.\ whiskers). Leaky RLS degrades in one $\lambda$ step ($182 \to 1.2 \times 10^6$ at window $200 \to 100$); shading marks the windup regime ($\mathrm{window} < D = 128$); directional forgetting is divergence-free but $3.3\times$ worse at its best; SPAR-Pull and the scalar SPAR method use one fixed configuration everywhere.}
\label{fig:cliffapp}
\end{figure}

\emph{Divergence boundary.}
The leaky covariance-form recursion is one tuning step from failure on this grid: $\lambda = 0.997$ attains $0.555$ mean MSE, $\lambda = 0.995$ attains $182$ (worst $3{,}878$), and $\lambda = 0.99$ attains $1.2 \times 10^6$, diverging in $135/135$ main-grid and $20/20$ stress-suite runs (nominal window $100 < D = 128$); Figure~\ref{fig:cliffapp} plots the sweep.
The batch information-form solve of the same objective is stable at every $\lambda$ ($0.34$ mean at $\lambda = 0.99$), which locates the explosion in the covariance recursion rather than the objective; the family's $\lambda$-sensitivity remains (its best stable window is $1.1\times$ SPAR's).

A diagnostic log of the forgetting factor (\texttt{diagnostics.py} \texttt{--diag vff}, \texttt{results/vff\_diag.csv}) confirms the degeneracy: under the reference calibration $\xi = \sigma_0^2$ the factor remains pinned at its floor on $96\%$ of steps on both stationary and drifting streams (mean $\lambda = 0.990$ at floor $0.99$), and a recursive-scale variant of the rule (a Fortescue-style running $\xi$ \citep{fortescue1981}) still pins at the floor ($92\%$ of steps, overall MSE within $2\%$ of the best fixed-$\lambda$ method).
Residual-driven forgetting degenerates to fixed-$\lambda$ leaky RLS whenever $\xi$ tracks the residual scale, a saturation documented in the variable-forgetting-factor literature \citep{paleologu2008}.
The covariance-form VFF inherits the same calibration sensitivity: the single-step normalization $\xi = \sigma_0^2$ drives the factor below $D$ and it diverges identically, while the nominal-length normalization $\xi = N_0\sigma_0^2$ of the original formulation keeps the nominal window near $N_0$ and is stable (best single setting $N_0 = 100$: $0.267/0.98$; the single-step calibration diverges in all 135 runs), the horizon choice $N_0$ playing the role of the fixed window that SPAR's window estimator sets online.
The first-order family terminates at its own boundary: at $\mathrm{lr} = 0.02$, one grid step above the swept maximum, SGD diverges on 45 of 135 runs, and at $\mathrm{lr} = 0.05$ the recursion overflows outright (\texttt{diagnostics.py} \texttt{--diag sgd}); the reported grid $\{0.003, 0.01\}$ sits at the stable frontier of the constant-step family.

\subsection{Windup Mechanism}
The divergence of the covariance-form recursion is governed by a single amplification mechanism acting through two channels: any subspace component not downdated by the rank-one correction inflates by $(1{-}\omega)^{-1}$ per step, growing as $e^{\omega T}$ over the stream.
On full-rank spectra the unseen component is roundoff: the exponentially weighted covariance matrix is a positive combination of $T \gg D$ outer products, full-rank almost surely at every window, with limiting spectral support bounded away from zero (the lower edge obeys $\lambda - \ln\lambda = 1 + c$, positive for all $c$); the iterated inverse $P \leftarrow (P - kk^\top)/(1{-}\omega)$ instead accumulates antisymmetric roundoff error, and a one-line symmetrization $P \leftarrow \tfrac12(P + P^\top)$ isolates this channel (\texttt{diagnostics.py} \texttt{--diag windup}; 2 representative configurations $\times$ 2 seeds per row):

\begin{table}[h]
\centering
\small
\caption{Mean MSE of the covariance-form leaky recursion on full-rank spectra, with and without per-step symmetrization of the iterated inverse.
At the longest window both variants agree to machine precision; at short windows the unsymmetrized recursion diverges at both dimensions while the symmetrized variant remains stable at every window tested.}
\label{tab:windup-mech}
\begin{tabular}{lccc}
\toprule
$\lambda$ ($1/(1-\lambda)$) & $D$ & standard recursion & symmetrized \\
\midrule
$0.997$ ($333$) & 128 & $1.15$ & $1.15$ \\
$0.995$ ($200$) & 128 & $25.1$ & $0.80$ \\
$0.99$ ($100$) & 128 & $2.8 \times 10^{4}$ & $0.61$ \\
$0.99$ ($100$) & 512 & $7.0 \times 10^{4}$ & $10.4$ \\
\bottomrule
\end{tabular}
\end{table}

On rank-deficient spectra the second channel is deterministic: the dead directions receive no data, so nothing downdates the corresponding block of $P$ and it inflates by $(1{-}\omega)^{-t}$ in exact arithmetic, symmetrized or not.
The symmetrized grid comparison confirms the split (\texttt{rls\_sym} rows of \texttt{results/factorial\_extras.csv}): stable on all 90 full-rank runs ($0.13$ mean on the ill-conditioned spectrum, $0.60$ isotropic, at $\lambda = 0.99$) and divergent on 42 of 45 rank-deficient runs ($45$ of $45$ at $\lambda = 0.97$).
The anchored information form is immune to both channels by construction, since it never iterates an inverse, and Proposition~\ref{prop:windup} bounds its conditioning for any data and any window.
The single-step VFF calibration enters the same regime through its floor (windows $20$--$100$ at $\lambda_{\min} \in \{0.95, 0.99\}$); the nominal-length calibration keeps windows near $N_0$, above the amplification threshold at every setting tested.

\subsection{False-escalation Calibration}
The calibration experiment of Proposition~\ref{prop:falsealarm} runs 300 stationary streams per spectrum through the SPAR estimator with a fixed, calibrated model, records the two-scale residual statistic $z_t = (\widehat\sigma^2_{\mathrm{fast}} - \widehat\sigma^2_{\mathrm{slow}})/\sigma^2$, and measures the fraction of streams whose maximum exceeds the sliding-EMA threshold $\tau(\delta) = \sqrt{2\nu_z^2\log(T/\delta)/m_{\mathrm{eff}}}$ with $m_{\mathrm{eff}} = (2-\eta_f)/\eta_f$ at the fast gain $\eta_f = 1/h_f$ and the Gaussian excess proxy $\nu_z^2 = 2$.
The measured rate is $0.003$ on isotropic spectra at nominal $\delta = 0.10$ and $0.040$ on ill-conditioned spectra at nominal $\delta = 0.05$, at or below the nominal level; at $\delta = 0.01$ the ill-conditioned rate is $0.033$. The sub-exponential threshold's linear branch predicts this exceedance: at $\delta = 0.01$ and $m_{\mathrm{eff}} \approx 199$, the linear term $\eta_f b_z \log(T/\delta)$ dominates the quadratic branch, so the Gaussian-form threshold $\sqrt{2\log(T/\delta)/m_{\mathrm{eff}}}$ used in the calibration experiment is provably too low by the $b_z$-scaling of the moving-average's heavy tail, and the measured $0.033$ exceeds the nominal $0.01$ accordingly.

\subsection{Ablation and Sensitivity}
The ablation (\texttt{run\_all.py --steps ablation}) runs the prediction-unit method and its ablations on 4 representative configurations (abrupt and oscillating drifts, adversarial prior, ill-conditioned spectrum) $\times$ 5 seeds.
Removing the window estimator (fixed $\omega$) costs $+189\%$ overall MSE, removing the two-scale residual detector $+190\%$, and removing the pull-identity refresh $+189\%$; setting $\gamma = 0.5$ costs $+560\%$, which measures the $\gamma = 1$ principle of Appendix~\ref{app:proof-dominance}.
Freezing the strength estimator to its warm-up value \emph{lowers} overall MSE by $2.2\%$ on these configurations, and removing the variance-floor subtraction costs $+1.7\%$ under the nominal cap, with the undebiased estimator pinning the window at its cap as Theorem~\ref{thm:window} predicts.
The three-way dissociation in Table~\ref{tab:ablation} localizes the prior-immunity mechanism: freezing the spectral anchor refresh costs $2.2\times$ on the warm-prior configurations and $7.4\times$ on the adversarial ones (adversarial early MSE $47.0$ vs.\ $16.4$), whereas freezing $\rho$ leaves both within $2.6\%$.
Prior immunity is therefore carried by the refresh, and the strength estimator's role is the drift-onset conflict response measured in Appendix Figure~\ref{fig:cliffdials}; it matters only when the warm-up prior scale is unknown, where it guards against an oversized anchor.
An alternating-trust diagnostic probes the strength estimator's average-case value directly (\texttt{diagnostics.py} \texttt{--diag alt\_trust}: the target switches every 500 steps between a noisy revisit of the anchor and a drifted target, two spectra, 5 seeds): the adaptive rule attains $6.79$ overall MSE against $6.64$--$7.00$ for pinned strengths spanning three orders of magnitude ($10^{-3}$ to $1$), a $\pm 3\%$ spread, because the anchor \emph{refresh}, not the strength estimator, carries the phase-to-phase adaptation.

\begin{table}[h]
\centering
\footnotesize
\caption{Component dissociation on the 4-configuration ablation suite (mean $\pm$ s.d.\ over 20 runs per cell; early MSE is the first 500 post-jump steps). Freezing the anchor refresh is the only variant that loses prior immunity; freezing the strength estimator is indistinguishable from the full method.}
\label{tab:ablation}
\begin{tabular}{lccc}
\toprule
variant & warm overall & adversarial overall & adversarial early \\
\midrule
full & $2.26 \pm 3.28$ & $1.04 \pm 0.21$ & $16.4 \pm 3.3$ \\
fixed $\rho$ (refresh on) & $2.20 \pm 3.16$ & $1.05 \pm 0.21$ & $16.5 \pm 3.3$ \\
frozen anchor (no refresh) & $4.98 \pm 6.49$ & $7.69 \pm 1.62$ & $47.0 \pm 10.2$ \\
no variance floor & $2.30 \pm 3.33$ & $1.06 \pm 0.21$ & $16.7 \pm 3.3$ \\
\bottomrule
\end{tabular}
\end{table}

This behavior reflects a sharp fixed-point dichotomy. The unsubtracted estimator $\hat\nu^2_{\mathrm{naive}}(\omega) = (\widehat\E[r_t^2] - \hat\sigma^2)\omega^2$ yields a cubed estimator $(T_{\mathrm{naive}}(\omega))^3 = (\omega^*)^3 + 4\omega^3/(2-\omega) > 2\omega^3$ for every $\omega \in (0,2)$, so the cubed trajectory doubles each iteration and reaches the cap in $O(\log)$ steps even at zero drift (\texttt{naive\_dial\_no\_fixed\_point}), while the variance-floor-subtracted estimator isolates $\nu^2_\Sigma$ and induces the affine contraction $u_{m+1} = (1-\eta_\nu)u_m + \eta_\nu(\omega^*)^3 \to (\omega^*)^3$ for $0 < \eta_\nu < 2$ (\texttt{debiased\_dial\_estimator\_eq}, \texttt{debiased\_dial\_contraction}, \texttt{debiased\_dial\_tendsto}); all are formalized in \texttt{Core.lean}.
Replacing the inverse trace $\mathrm{tr}(\bar A^{-1})$ with the ridge-damped variance $\mathrm{tr}(\bar A(\bar A + \rho I)^{-2})$ on the rank-deficient configurations (\texttt{results/damped\_trace.csv}) yields near-identical error ($0.100$ vs.\ $0.101$ overall): the residual-detector and floor branches govern the estimator where the trace saturates, and the damped variant is the numerically preferred implementation on ill-conditioned spectra.
The sensitivity variants (legacy parameter-unit method, \texttt{results/ablation\_spar\_kf.csv}; the control constants are shared with the prediction-unit method) scale the residual horizons ($\pm$), the refresh cadence ($10 \to 50$), the reference-anchor cadence, and the window cap ($0.05 \to 0.02$ and $0.10$): six of the seven variants change overall MSE by at most $\pm 10\%$, and the largest change is the window cap $\omega_{\max} = 0.02$ at $+16\%$.

\subsection{The Conflict Estimator at Drift Onset}
\label{app:transient}

Table~\ref{tab:transient} pins the strength at $\rho \in \{0.01, 0.1, 1\}$ with the refresh on and compares it with the adaptive estimator on the abrupt-jump stream with the decay-3 spectrum (5 seeds; \texttt{scripts/transient\_test.py}, \texttt{results/transient\_test.csv}).
At cadence $k = 10$ the adaptive estimator is $1\%$ below the best fixed strength and the large-anchor policy costs $87\%$ more; at $k = 50$ it is within $5\%$ of the best fixed strength while the large-anchor policy costs $2.8\times$.
The penalty is pre-drift, not post-jump: $\rho = 1$ multiplies the pre-drift MSE by $4.2$ relative to the small fixed anchor ($0.049$ vs.\ $0.012$ at $k = 10$) because the anchor holds a warm prior that the decaying spectrum no longer supports, whereas the $50$-step post-jump window is dominated by the jump magnitude and is flat across the adaptive and small-anchor policies, if anything slightly better for the large anchor ($7.7$ vs.\ $9.5$).
On the isotropic spectrum, where a scalar strength has no structural effect, every policy has the same pre-drift MSE ($0.0104$) and differs only through the post-jump term.
We therefore read $\rho_t$ as a guard against a badly oversized anchor, the failure mode a good warm prior invites, not as a post-jump accelerator.

\begin{table}[!htbp]
\centering
\footnotesize
\caption{Transient strength sweep on the abrupt-jump stream, decay-3 spectrum, 5 seeds: pre-drift MSE, MSE over the $50$ steps after the jump, and overall MSE.
The refresh is on in every row; the adaptive row is the paper's method and the others pin $\rho$ for the whole stream.}
\label{tab:transient}
\begin{tabular}{ccccc}
\toprule
$k$ & strength policy & pre-drift MSE & post-jump MSE $[0,50]$ & overall MSE \\
\midrule
10 & adaptive & $0.0136$ & $9.48$ & $\mathbf{0.166}$ \\
10 & fixed $\rho = 0.01$ & $0.0116$ & $9.48$ & $0.168$ \\
10 & fixed $\rho = 0.1$ & $0.0271$ & $9.49$ & $0.195$ \\
10 & fixed $\rho = 1$ & $0.0489$ & $7.72$ & $0.311$ \\
\midrule
50 & adaptive & $0.0197$ & $9.52$ & $0.254$ \\
50 & fixed $\rho = 0.01$ & $0.0146$ & $9.54$ & $\mathbf{0.243}$ \\
50 & fixed $\rho = 0.1$ & $0.0338$ & $9.56$ & $0.308$ \\
50 & fixed $\rho = 1$ & $0.0495$ & $8.52$ & $0.723$ \\
\bottomrule
\end{tabular}
\end{table}

\subsection{Coupling Verification}
Theorem~\ref{thm:closedloop}(i)'s condition is measured directly (\texttt{diagnostics.py} \texttt{--diag coupling}, \texttt{results/coupling\_diag.csv}): on the four ablation configurations $\times$ 5 seeds, each estimator is pinned at $\pm 10\%$ of its converged value while the other adapts, and the two sensitivities $L_{\omega\rho}$ (of the cubed window target) and $L_{\rho\omega}$ (of the strength target) are finite-differenced at the empirical coupled fixed point.
The product $L_{\omega\rho} L_{\rho\omega} < 1$ holds in 20 of 20 runs, with maximum $5.7 \times 10^{-3}$, which is $174$ to $1381$ times below one on the oscillating configuration, the only one with both estimators interior.
On three of the four configurations one estimator saturates at its cap or floor, making the corresponding sensitivity exactly zero. A pinned estimator realizes the triangular specialization of Theorem~\ref{thm:closedloop}(ii), and the large cubed-coordinate slopes recorded at floored windows are boundary artifacts of perturbing a microscopic $u = \omega^3$, multiplied by an exactly-zero factor: the caps act as the coupling control, and the cross-signs at the interior fixed point match the theory ($u^*$ increasing in $\rho$, $\rho^*$ decreasing in $u$).

\subsection{Spectral-risk Verification}
Theorem~\ref{thm:rmt}(i)'s closed form was validated at three levels (\texttt{diagnostics.py} \texttt{--diag rmt}).
First, at $45$ fixed operating points (three spectra, including a rank-deficient one, $\times$ five nominal windows $1/(1-\lambda) \in \{100, \dots, 2000\}$ $\times$ three anchor strengths $\rho \in \{10^{-3}, 10^{-2}, 10^{-1}\}$), the formula matches direct Monte Carlo simulation of the anchored solve ($40$ stationary streams per point, oracle anchor) within $4.4\%$ relative error across the entire range, including aspect ratios $\zeta = D\omega$ above one and the rank-deficient spectrum, whose active directions the formula restricts to exactly (\texttt{results/rmt\_grid.csv} records the grid, \texttt{results/rmt\_rankdef.csv} the active-subspace invariance, and \texttt{results/rmt\_verify.csv} the method-level checks).
Second, at the method's own operating points (the converged estimator pairs $(\omega, \rho)$ measured on stationary streams per spectrum), the formula predicts the method's stationary pre-drift MSE within $1.9\%$ (rank-deficient), $12.7\%$ (ill-conditioned), and $9.8\%$ (isotropic, where the strength estimator sits at its cap), with no fitted parameter.
Third, the closed-form sensitivities of part~(ii) agree with numerical differentiation of the fixed point to $10^{-6}$ relative error.
The pre-drift MSE reported in Section~\ref{sec:experiments} ($1.01$--$1.14 \times \sigma^2$ across spectra) is the segment average including the post-warm-up transient; the formula describes the stationary value at the converged operating pair.

\subsection{Per-step Cost}
Table~\ref{tab:app-ops} reports the per-step wall-clock cost of the closed-form methods at $D = 2048$ (fp64 CPU, \texttt{scripts/bench\_ops.py}; run alone for clean timing).
The cached-basis per-direction head is $O(D^2)$ per step with $O(D^3)$ at refreshes; the scalar method solves the anchored system exactly at $O(D^3)$ per step, and its spectral refresh reuses the solve's factorization in $O(D^2)$ (Section~\ref{sec:method-solve}).
The information form deliberately trades the covariance recursion's $O(D^2)$ rank-one update for a factorization that does not wind up on rank-deficient spectra (Proposition~\ref{prop:windup}); where per-step cost dominates, the cached-basis head is the intended path.

\begin{table}[h]
\centering
\footnotesize
\caption{Per-step cost at $D = 2048$ (fp64 CPU).
The cached-basis per-direction head projects each feature into the last refresh's eigenbasis and solves the diagonal system ($O(D^2)$/step, $O(D^3)$ every $k$ refreshes); the scalar method performs a Cholesky factorization plus triangular solves per step ($O(D^3)$), which motivates the per-direction path at large $D$.}
\label{tab:app-ops}
\begin{tabular}{lc}
\toprule
method & ms/step at $D = 2048$ \\
\midrule
leaky RLS & $32.9$ \\
leaky RLS, information form & $189$ \\
AROW & $30.6$ \\
Kalman & $50.5$ \\
directional forgetting & $58.6$ \\
SPAR-Pull-D (cached basis, $O(D^2)$/step) & $51.3$ \\
SPAR-Pull (Algorithm~\ref{alg:spar-pull}, $O(D^3)$/step) & $363$ \\
SPAR-Pull-S (exact scalar solve, $O(D^3)$/step) & $225$ \\
\bottomrule
\end{tabular}
\end{table}
SPAR-EG and SPAR-EG$^+$ run four batch solves per step (the frozen anchor is the remaining expert in the pool), so their per-step cost is about four times the scalar SPAR-Pull solve timed here; the cached-basis path applies to them as well.

\subsection{Mismatched Generators}

\begin{figure}[h]
\centering
\includegraphics[width=2.9in]{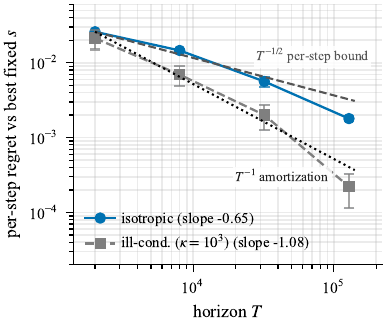}
\caption{Average per-step regret of the segment (scalar-pull) OGD method versus horizon $T$ (log--log; whiskers: $\pm 1$ s.d.\ over 5 seeds; the drift phase lasts a fixed $1{,}000$ steps at every horizon).
Fitted slopes $-0.65$ (isotropic) and $-1.08$ (ill-conditioned, $\kappa = 10^3$) lie inside the $T^{-1/2}$ per-step rate (dashed) implied by the scalar $O(\sqrt T)$ Zinkevich bound \citep{zinkevich2003} (the $D = 1$ member of the Proposition~\ref{prop:ocowrap}(i) family, formalized as \texttt{ogd\_regret\_segment} in \texttt{Ogd.lean}), with the $T^{-1}$ amortization limit (dotted) as the mechanical floor.}
\label{fig:regret}
\end{figure}

At a fixed horizon, the dimension sweep separates the two gradient variants.
With $T = 9{,}000$, the drift phase held at $1{,}000$ steps, $D \in \{8, 32, 128\}$, and 5 seeds (\texttt{scripts/regret\_dscaling.py}, \texttt{results/regret\_dscaling.csv}), both methods' per-step regret is measured against the best fixed scalar pull (an 11-point segment grid, each comparator method carrying its own refresh trajectory).
The profile method's regret exceeds the segment method's at every operating point, and the gap widens monotonically with $D$, from $0.001$ to $0.078$ on isotropic spectra and from $3 \times 10^{-6}$ to $0.011$ on ill-conditioned spectra, the empirical signature of the $\sqrt{D}$ dimension factor in the Proposition~\ref{prop:ocowrap}(i) bound; on the full grid the same premium appears as the profile method's higher mean MSE ($0.707$ vs.\ $0.585$, Table~\ref{tab:grid}).

\begin{figure}[h]
\centering
\begin{subfigure}[t]{0.48\textwidth}
\centering
\includegraphics[width=2.55in]{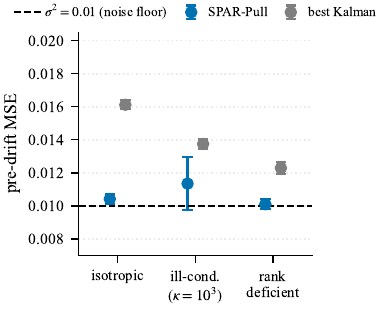}
\caption{}
\label{fig:floor}
\end{subfigure}\hfill
\begin{subfigure}[t]{0.48\textwidth}
\centering
\includegraphics[width=2.55in]{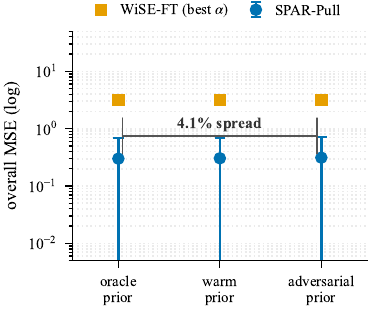}
\caption{}
\label{fig:prior}
\end{subfigure}
\caption{(a) Pre-drift MSE remains at the irreducible floor $\sigma^2 = 0.01$ across all three spectra (dots show 5-seed means $\pm 1$ s.d.\ against the best tuned-$q$ Kalman with true $\sigma^2$), illustrating Proposition~\ref{prop:conflict} at the optimum.
(b) Overall MSE (log scale) across oracle, warm, and adversarial priors (means over all runs).
While the WiSE-FT best-$\alpha$ envelope sits near $3.2$ under every prior, SPAR-Pull varies by only $4.1\%$ across all three priors (bracket): Proposition~\ref{prop:minimax}.}
\label{fig:floorprior}
\end{figure}

Three generators violate the linear-Gaussian, constant-rate model class: nonlinear targets ($y = \tanh(x^\top w^*) + \varepsilon$; the linear head is misspecified), covariate shift (the feature scale doubles mid-stream), and a noise-regime change ($\sigma^2$ rising from $0.01$ to $0.05$ at $t_1$, with no target drift).
On nonlinear targets (four configurations: decay3 and isotropic spectra under abrupt, gradual, and stationary drifts; 5 seeds each) SPAR-EG$^+$ attains $0.221$ overall MSE, SPAR-EG $0.222$, the best single leaky window ($\lambda = 0.997$) $0.232$, and SPAR-Pull $0.273$: the mixture shifts weight toward the expert that tolerates the nonlinearity.
Under covariate shift the Kalman family leads ($0.268$ with the true noise, $0.284$ with the slow estimated $R$), while SPAR-EG$^+$ and SPAR-Pull lead the anchored and leaky methods ($0.490$ and $0.492$), ahead of the batch-solved leaky method ($0.550$), SPAR-EG ($0.566$), and the best single leaky window ($0.627$), the two-scale residual detector catching the scale change.
Under the noise-regime change (two configurations, decay3 and isotropic spectra, 5 seeds each) the fast-horizon estimated-$R$ Kalman attains $0.069$--$0.098$, its $20{,}000$-step-horizon sibling $0.084$--$0.110$, and the true-noise filter $0.096$--$0.114$ (its fixed $R$ is wrong after the change), while the SPAR-EG mixture attains $0.047$--$0.087$ and SPAR-EG$^+$ $0.047$--$0.053$; SPAR-Pull holds $0.052$ on the decay3 spectrum but lags at $0.080$ on the isotropic spectrum, where its slow noise baseline tracks the pre-change level.
No SPAR method diverged under any mismatch, while the $\lambda = 0.995$ leaky method diverged on nonlinear targets (two of five seeds on the decay3 spectrum under abrupt drift) and degraded under the noise change ($9.1$ and $9.4\times 10^3$ overall MSE).

\subsection{A Real High-dimensional Stream: Hourly Electricity Load}
\label{app:realstream}

The factorial grid is synthetic, so we also run the grid-default configuration, unchanged, on a real stream: the public hourly electricity-load panel (321 clients, $26{,}304$ hours) \citep{trindade2015electricity,lai2018lstd}.
Features at hour $t$ are the standardized log-loads of all 321 clients plus an intercept and hour-of-day sine/cosine ($D = 324$); the target is the log-load of one held-out client (the 25th-percentile client by mean load) one hour ahead, centered on its warm-up mean.
The warm-up is $2{,}000$ hours and the evaluation horizon $6{,}000$ hours, so $T > 5{,}000$ and $D > 128$ as required.
The stream is not stationary: load carries daily and weekly cycles and a seasonal trend, and the target series moves with them.

Both SPAR methods use the grid defaults of Table~\ref{tab:hyper} unchanged: window range $[5 \times 10^{-4}, 0.05]$, refresh cadence 10, residual horizon 30, and the drift-active floor $0.02$.
The baselines are swept on the same stream and reported at their best setting, which is oracle tuning in their favour: leaky RLS over $\lambda \in \{0.99, 0.995, 0.999\}$ and the slow estimated-$R$ Kalman over $q \in \{10^{-4}, 10^{-5}\}$.

\begin{table}[!htbp]
\centering
\footnotesize
\setlength{\tabcolsep}{5pt}
\caption{Real stream (hourly electricity load, $D = 324$, $T = 6{,}000$ evaluation hours, 25th-percentile held-out client): MSE of the next-hour log-load over the full horizon and over the final $2{,}000$ hours.
SPAR methods use the factorial-grid defaults with no stream-specific tuning; baseline hyperparameters are swept on the same stream and reported at their best.}
\label{tab:electricity}
\begin{tabular}{lcc}
\toprule
method & full horizon & final $2{,}000$ h \\
\midrule
leaky RLS, information form ($\lambda = 0.999$) & $\mathbf{0.0209}$ & $\mathbf{0.0134}$ \\
leaky RLS ($\lambda = 0.995$) & $0.0219$ & $0.0160$ \\
leaky RLS ($\lambda = 0.99$) & $0.0228$ & $0.0174$ \\
Kalman, slow estimated $R$ ($q = 10^{-5}$) & $0.0660$ & $0.0170$ \\
Kalman, slow estimated $R$ ($q = 10^{-4}$) & $0.3054$ & $0.0183$ \\
\midrule
SPAR-EG$^+$ (grid defaults) & $0.0212$ & $0.0143$ \\
SPAR-Pull (grid defaults) & $0.0232$ & $0.0154$ \\
\bottomrule
\end{tabular}
\end{table}

\begin{table}[!htbp]
\centering
\footnotesize
\setlength{\tabcolsep}{5pt}
\caption{Real stream, held-out client sweep: MSE of the next-hour log-load over the full $6{,}000$-hour horizon for the 25th-, 50th-, and 75th-percentile clients by mean load.
The oracle column is the best of the three leaky windows ($\lambda \in \{0.99, 0.995, 0.999\}$) chosen \emph{per client} ($\lambda = 0.999, 0.995, 0.995$ at the 25th, 50th, 75th percentiles), tuning in the baseline's favour, while the SPAR methods use the grid defaults with no stream-specific tuning.
Bold marks the best entry in each row.}
\label{tab:electricity-clients}
\begin{tabular}{lccc}
\toprule
held-out client & best oracle leaky & SPAR-EG$^+$ & SPAR-Pull \\
\midrule
25th percentile & $\mathbf{0.0209}$ & $0.0212$ & $0.0232$ \\
50th percentile & $0.0086$ & $\mathbf{0.0083}$ & $0.0090$ \\
75th percentile & $\mathbf{0.0112}$ & $0.0113$ & $0.0130$ \\
\midrule
three-client mean & $\mathbf{0.0136}$ & $\mathbf{0.0136}$ & $0.0151$ \\
\bottomrule
\end{tabular}
\end{table}

Across the sweep, SPAR-EG$^+$ lands within $1.2\%$ of the per-client oracle window and $4.3\%$ below it at the 50th percentile, and it matches the oracle on the three-client mean ($0.0136$ both); SPAR-Pull is $4.2$--$16.5\%$ above the oracle.
The ordering of the closed-form methods is stable across the three clients, so the untuned transfer is not specific to a single reported percentile.

Three observations.
First, the untuned SPAR-EG$^+$ lands $1.1\%$ above the best oracle-tuned leaky window over the full horizon and $6.3\%$ above it on the final $2{,}000$ hours, while SPAR-Pull is $10.7\%$ and $14.7\%$ above; a smooth seasonal stream has a long and nearly constant optimal window, which is exactly the regime in which a fixed long window is close to optimal and adaptation has little to recover.
Second, the slow estimated-$R$ Kalman is far worse over the full horizon ($0.051$ and $0.257$) and still $26\%$ above the leaky baseline on the final $2{,}000$ hours, i.e.\ its early noise estimates are poor and it only recovers late.
Third, no method diverges and the ordering of the closed-form methods is stable across the horizon, so the control law transfers to a real $D = 324$ stream without retuning.

\section{BCI IV Details}
\label{app:eeg}

\emph{Datasets and protocol.}
BCI Competition IV 2a \citep{brunner2008bci2a} and 2b \citep{leeb2008bci2b}, nine subjects each, loaded from the npz cache described in Appendix~\ref{app:repro}.
Cross-session protocol: each method fits on the training session and streams the evaluation session (288 trials on 2a, 320 on 2b) predict-then-update.
All online methods use supervised adaptation: the label of trial $t$ enters after its prediction, so every prediction is prequential.
Labels are one-hot $Y_t \in \{0,1\}^K$; the streaming head maintains a $D \times K$ coefficient matrix and the scalar residual driving both estimators is $r_t^2 = \|Y_t - \widehat Y_t\|_2^2/K$.
The metric is balanced accuracy; with $n = 9$ subjects per data set we report paired tests and effect sizes, not significance alone.

\begin{table}[h]
\centering
\scriptsize
\setlength{\tabcolsep}{2.5pt}
\renewcommand{\arraystretch}{0.86}
\caption{BCI IV cross-session, balanced accuracy (mean $\pm$ s.d.\ over 9 subjects per data set).
Supervised online protocol for adaptive methods; classical methods are static. $^{\mathrm{T}}$Session-level Euclidean alignment from the unlabeled evaluation session (transductive); the online-EA row is strictly causal.
Highest value per column in bold.}
\label{tab:eeg}
\begin{tabular}{lcc}
\toprule
method & BCI IV 2a & BCI IV 2b \\
\midrule
\multicolumn{3}{l}{\emph{classical (canonical implementations)}}\\
MDM (pyriemann) & $0.566 \pm 0.160$ & $0.728 \pm 0.148$ \\
FgMDM (pyriemann) & $0.641 \pm 0.171$ & $0.755 \pm 0.149$ \\
LDA (raw features) & $0.480 \pm 0.069$ & $0.651 \pm 0.107$ \\
LDA (raw) + session EA$^{\mathrm{T}}$ & $0.563 \pm 0.076$ & $0.677 \pm 0.133$ \\
CSP-LDA (mne) & $0.607 \pm 0.151$ & $0.749 \pm 0.141$ \\
tangent cov + LDA & $0.637 \pm 0.165$ & $0.752 \pm 0.147$ \\
cov-LDA, offline session-EA$^{\mathrm{T}}$ & $0.664 \pm 0.173$ & $0.728 \pm 0.159$ \\
\midrule
\multicolumn{3}{l}{\emph{ours (closed-form streaming, predict-then-update)}}\\
CSP + SPAR-Pull & $0.621 \pm 0.170$ & $0.768 \pm 0.145$ \\
tangent cov + SPAR-Pull & $0.537 \pm 0.154$ & $0.773 \pm 0.128$ \\
hybrid + SPAR-Pull$^{\mathrm{T}}$ & $0.577 \pm 0.133$ & $0.808 \pm 0.103$ \\
hybrid + SPAR-Pull, grid defaults$^{\mathrm{T}}$ & $0.560 \pm 0.132$ & $0.800 \pm 0.103$ \\
online-EA + hybrid + SPAR-Pull & $0.574 \pm 0.127$ & $0.800 \pm 0.106$ \\
hybrid + AAL head ($\gamma{=}1$, fixed anchoring)$^{\mathrm{T}}$ & $0.608 \pm 0.139$ & $0.799 \pm 0.107$ \\
hybrid + leaky head$^{\mathrm{T}}$ & $0.554 \pm 0.123$ & $0.812 \pm 0.108$ \\
hybrid + SPAR-EG$^{\mathrm{T}}$ & $0.580 \pm 0.130$ & $\mathbf{0.816 \pm 0.102}$ \\
hybrid + SPAR-EG$^{+\,{\mathrm{T}}}$ & $0.578 \pm 0.135$ & $0.813 \pm 0.107$ \\
\midrule
\multicolumn{3}{l}{\emph{backprop references}}\\
EEGNet-lite (train-split tuned, zero-shot) & $0.556 \pm 0.080$ & $0.776 \pm 0.129$ \\
WiSE-FT on the net (causal, $\alpha$-envelope) & $0.591 \pm 0.079$ & $0.811 \pm 0.115$ \\
\midrule
post-hoc ensemble (cov-LDA + hybrid warm fit)$^{\mathrm{T}}$ & $\mathbf{0.668 \pm 0.175}$ & $0.771 \pm 0.104$ \\
\bottomrule
\end{tabular}
\end{table}

\emph{Compute.}
The complete closed-form suite (feature extraction plus all closed-form methods) for all nine 2b subjects runs in $17$ s on the CPU used here ($1.9$ s per subject; \texttt{scripts/bench\_eeg\_tier1.py}); one EEGNet-lite training configuration ($120$ epochs) trains on the WSL GPU environment, and the full backprop protocol trains three configurations per subject plus the online fine-tuning pass.

\emph{Feature families.}
Three feature families feed the readouts of Table~\ref{tab:eeg}.
CSP: 6-component CSP (Ledoit--Wolf shrinkage \citep{ledoitwolf2004}, log output) computed with MNE \citep{gramfort2013mne} on bandpassed epochs (8--30 Hz, 4th-order Butterworth), the classical motor-imagery standard \citep{ramoser2000csp}.
Tangent covariance: per-trial log-Euclidean covariance tangent features \citep{arsigny2007} (lower triangle) on bandpassed epochs, the feature family of pyriemann's tangent-space methods \citep{barachant2012mdm,barachant2013}.
Hybrid: the statistical pipeline (summary statistics, lags $(1, 2, 4, 8)$, FFT bins) concatenated with the tangent-covariance features.
Classical readouts are MDM and FgMDM \citep{barachant2012mdm,barachant2010riemannian} as implemented in pyriemann \citep{barachant2020pyriemann}, raw-feature LDA with and without session-level Euclidean alignment \citep{he2020ea}, CSP-LDA, tangent-covariance LDA (shrinkage LDA in all cases), and a covariance-LDA method with the same offline alignment.
Streaming readouts are the two-estimator SPAR-Pull head (anchor refresh every 5 trials), the AAL head (fixed $\gamma = 1$ anchoring), and a forgetting-only leaky head, each applied to CSP, tangent-covariance, or hybrid features.

\emph{Alignment protocol.}
The hybrid methods use session-level Euclidean alignment: per-trial covariance whiteners are computed from each session separately, so the alignment is transductive at the session level (the offline classical methods share this convention).
The strictly online method recomputes its whitener causally from a trailing window of the evaluation stream (40 trials, refreshed every 10) for \emph{both} feature halves: the statistical features and the tangent-covariance features are each whitened with the trailing-window estimate, so no stream feature depends on trials beyond the current one.
All SPAR methods in Table~\ref{tab:eeg} run the same variance-floor-subtracted window inversion as the factorial grid (Theorem~\ref{thm:window}), with the estimator ranges rescaled to the $288$-trial stream (window range $[5\times10^{-3},0.02]$, refresh cadence 5, residual horizon 40, calibration blocks 2).
The \emph{grid defaults} row is the same hybrid head with the factorial-grid window range, refresh cadence, residual horizon, and calibration blocks ($[5\times10^{-4},0.05]$, 10, 30, 10); the gap to the EEG-range row prices those four settings ($0.018$ balanced accuracy on 2a and $0.007$ on 2b).
The reduced range therefore helps, but the ranking of the readouts and the qualitative conclusion (closed-form streaming matches the causal backprop reference) do not depend on it.

\emph{WiSE-FT protocol measurement.}
Transplanting the original WiSE-FT protocol \citep{wortsman2022wiseft} (fine-tune on the target data, then evaluate on it) into cross-session evaluation is train-on-test.
Measured directly, the offline fine-tuned endpoint reaches accuracy $1.000$ by memorizing the evaluation stream.
The comparison we report is therefore the causal reformulation: the network is fine-tuned online with one SGD step per trial \emph{after} predicting it, trial $t$ is predicted with the interpolation $(1 - \alpha)\,W_{\mathrm{zero}} + \alpha\,W_t$ where $W_t$ is the online model before trial $t$'s label, and the $\alpha$-envelope is swept post-hoc over 11 values with all trajectories causal (the prequential evaluation of Table~\ref{tab:eeg}).

\emph{Backprop references.}
EEGNet-lite is a small EEGNet variant \citep{lawhern2018eegnet} ($F_1 = 8$, $D = 2$, $F_2 = 16$) trained with Adam \citep{kingma2015adam}.
Its hyperparameters (epochs, learning rate) are selected on an 80/20 split of the training session from the grid $(120, 10^{-3})$, $(300, 10^{-3})$, or $(300, 3 \times 10^{-4})$ with no evaluation-session access, then retrained on the full training session and evaluated zero-shot on the evaluation session.
The WiSE-FT-on-network reference uses this zero-shot model as its frozen endpoint and the causal fine-tuning protocol above.

\section{Hyperparameters and Provenance}
\label{app:hyper}

\emph{Fixed constants and the default control parametrization.}
The complete fixed control parametrization is enumerated in Table~\ref{tab:hyper}: the three headline constants (anchor-refresh cadence 10 steps, fast-EMA residual horizon 30 steps, drift-active anchor-strength floor $\rho \ge 0.02$), the clip ranges of both estimators, the slow-noise-EMA baseline horizon (20{,}000, matched to the estimated-noise Kalman), the reference-anchor EMA horizon (200 blocks, i.e. $2{,}000$ steps), the conflict-floor multiplier ($\tfrac12$), the estimator smoothing gains ($\eta_C = 0.15$, $\eta_\nu = 0.2$), and the $10^{-8}$ numerical ridge of the trace and active-dimension computations.
These were fixed once, before the factorial grid, and are shared by every SPAR method on it; all remaining behavior is self-calibrated at runtime by the two online estimators.
The sensitivity analysis of Appendix~\ref{app:protocol} perturbs the headline constants directly.

\begin{table}[htbp]
\centering
\small
\caption{Fixed default control parametrization (grid: $T = 10^4$). The EEG head uses a reduced estimator range and shorter horizons for the short evaluation streams, hand-set once and disclosed here; no per-configuration search was performed for any SPAR method.}
\label{tab:hyper}
\begin{tabular}{lll}
\toprule
parameter & grid default & EEG head \\
\midrule
refresh cadence $k$ & $10$ & $5$ trials \\
fast-EMA horizon $h_f$ & $30$ & $40$ \\
warm-up anchor strength $\rho_0$ & $0.1$ & $0.1$ \\
stationary $\rho$ floor & $10^{-5}$ & $10^{-5}$ \\
drift-active $\rho$ floor & $0.02$ & --- \\
anchor-strength cap $\rho_{\max}$ & $1.0$ & $1.0$ \\
window clip $[\omega_{\min}, \omega_{\max}]$ & $[5\times10^{-4},\ 0.05]$ & $[5\times10^{-3},\ 0.02]$ \\
slow-noise-EMA horizon & $20{,}000$ & $2{,}000$ \\
reference-anchor EMA horizon & $200$ ($2{,}000$ steps) & $30$ \\
conflict-floor multiplier & $\tfrac12$ & $\tfrac12$ \\
smoothing gains $(\eta_C, \eta_\nu)$ & $(0.15,\ 0.2)$ & $(0.2,\ 0.1)$ \\
active-dimension / trace ridge & $10^{-8}$ & $10^{-8}$ \\
\bottomrule
\end{tabular}
\end{table}

\emph{Estimator floors and caps.}
The window estimator is clipped to $\omega \in [5 \times 10^{-4},\ 0.05]$ on the grid (the floor is no forgetting at the grid's horizon; the cap bounds the forgetting rate), and the anchor-strength estimator enforces the drift-active floor above together with a hard cap.
The EEG head uses a reduced range suited to the short evaluation streams ($\omega \in [5 \times 10^{-3},\ 0.02]$, fast horizon 40, refresh every 5 trials, reference-anchor horizon 30, smoothing gains $(\eta_C,\eta_\nu) = (0.2, 0.1)$) and omits the drift-active floor; these are the only stream-dependent settings in the paper.

\emph{Conflict estimator branches (prediction units).}
Algorithm~\ref{alg:spar-pull} step (iv) takes the maximum of three branches, all in prediction-risk units $[y^2]$ (the residual EMAs are $\sigma^2_{\mathrm{fast}}$ = fast EMA of $r_t^2$ at horizon $h_f$, $\sigma^2_{\mathrm{slow}}$ = slow EMA at horizon $20{,}000$; $d_{\mathrm{act}}$ is the number of eigenvalues of $\bar A$ above the relative threshold $10^{-8}\max_j\mu_j(\bar A)$):
(i) the reference branch $C^\Sigma_{\mathrm{ref}} = (G_{\mathrm{ref}} - \widehat w)^\top \bar A\,(G_{\mathrm{ref}} - \widehat w) - \widehat V^\Sigma$, where $\widehat w$ is the minimum-ridge data solve, $\widehat V^\Sigma = \hat\sigma^2 d_{\mathrm{act}}/n$, and $G_{\mathrm{ref}}$ is the reference anchor, an EMA of the refreshed solve with a $2{,}000$-step horizon (updated every $k$ steps with $\eta_{\mathrm{ref}} = k/2{,}000$);
(ii) the residual branch $C^\Sigma_{\mathrm{resid}} = E_t = \max(0, \sigma^2_{\mathrm{fast}} - \sigma^2_{\mathrm{slow}})$;
(iii) the floor $C^\Sigma_{\mathrm{floor}} = \tfrac12\,\hat\sigma^2 d_{\mathrm{act}}/T_w$ (the warm-up prediction variance is an irreducible lower bound on the conflict of a warm-up anchor).
The per-clip strength is $\rho = \mathrm{clip}(\bar\mu\,\widehat V^\Sigma/\widehat C^\Sigma_t,\ \rho_{\min}^{\mathrm{drift}},\ \rho_{\max})$ with the drift-active floor $0.02$ engaged while the window is at least half open ($\omega_t \ge \tfrac12\omega_{\max}$) and the stationary floor $10^{-5}$ otherwise.
Every branch has units $[y^2]$ and $\rho$ has units $[x^2]$; no inverse of $\bar A$ is formed, so the estimator is invariant to feature rescaling and immune to the rank-deficient trace blow-up.
The legacy parameter-unit form (with $\mathrm{tr}(\bar A^{-1})$ and $C_{\mathrm{resid}}=E_t\mathrm{tr}(\bar A^{-1})/(n\bar\mu)$) is reported only as the ablation of Appendix~\ref{app:protocol}.

\emph{Method identifiers.}
The paper's display names map to frozen artifact identifiers (internal Python/CSV keys) as follows: SPAR-Pull is \texttt{spar\_pull} (its legacy parameter-unit form is \texttt{spar\_kf2}), SPAR-Pull-S is \texttt{spar\_scalar}, SPAR-Pull-D is \texttt{spar\_diag}, SPAR-EG is \texttt{spar\_eg}, SPAR-EG$^+$ is \texttt{spar\_eg\_plus}, the causally-tuned Kalman filters are \texttt{kf\_causal} and \texttt{kf\_est\_causal}/\texttt{kf\_est\_slow\_causal}, the AAL head is \texttt{hybrid-pam\_g1}, leaky RLS is \texttt{leaky[$\lambda$]} in covariance form and \texttt{leaky\_batch} in information form, directional forgetting is \texttt{df[$\lambda_f$]}, AROW is \texttt{arow[$r$]}, the Kalman filters are \texttt{kf[$q$]} (true noise), \texttt{kf\_est} (100-step estimated $R$), and \texttt{kf\_est\_slow} (20{,}000-step estimated $R$), the gradient-descent baselines are \texttt{sgd} and \texttt{adam}, the OGD variants are \texttt{ogd\_profile} (scalar segment) and \texttt{ogd\_profile\_d} (full $D$-dimensional profile), the variable forgetting factor baseline is \texttt{vff[fortescue]} in covariance form and \texttt{vff\_info[fortescue]} in information form, and the WiSE-FT envelope is \texttt{wiseft}.
All renaming in the paper is display-layer only; the CSVs, scripts, and tests key on the artifact identifiers.

\emph{Seeds.}
All seeds are pinned in the drivers: the factorial grid uses seeds 0--4 per configuration; the ablation, regret scaling, coupling diagnostic, and mismatched-generator studies use seeds 0--4; the false-escalation calibration uses 300 streams (seeds 1000--1299); the estimator-trajectory figure uses seed 0; the EEG evaluation uses subjects 1--9 per data set (all available subjects); the BP references train with per-subject seeds equal to the subject index.
Seeds are shared across methods at each configuration.

\emph{Data manifests.}
Synthetic streams are generated from the pinned seeds by the committed generator (\texttt{rls\_baselines.make\_stream\_fac}), so no synthetic data files are needed.
The EEG evaluation reads the committed npz cache (\texttt{data/eeg\_cache/\{2a,2b\}/subject*.npz}), exported from the MNE data directory by \texttt{scripts/export\_eeg\_npz.py}.
Every result table in the paper is machine-printed from a committed CSV under \texttt{results/} (Appendix~\ref{app:repro}).

\section{Reproducibility}
\label{app:repro}

\emph{One driver.}
A single driver regenerates every figure and every result table from the committed caches:
\begin{quote}
\ttfamily
py scripts/run\_all.py --list\ \ list the pipeline steps\\
py scripts/run\_all.py --steps quick\ \ verify+aggregate+figures\\
py scripts/run\_all.py\ \ everything (hours; resumable)
\end{quote}
The quick preset takes about 2 minutes; the full run takes hours and is resumable.
\begin{table}[!htbp]
\centering
\footnotesize
\setlength{\tabcolsep}{4pt}
\caption{One driver, \texttt{scripts/run\_all.py}, runs these stages in order; each is resumable.}
\label{tab:pipeline}
\begin{tabular}{@{}p{0.72in}p{2.95in}p{1.55in}@{}}
\toprule
stage & driver & output \\
\midrule
\texttt{verify} & \texttt{verify\_spar\_theorems.py} & symbolic identities \\
\texttt{factorial} & \texttt{synthetic\_factorial.py} & \texttt{factorial\_d\_*.csv} \\
\texttt{sparkf} & \texttt{run\_all.py --steps sparkf} & \texttt{factorial\_extras.csv} \\
\texttt{append} & \texttt{run\_all.py --steps append} & \texttt{factorial\_pu\_refresh.csv} \\
\texttt{aggregate} & \texttt{aggregate\_factorial.py} & Table~\ref{tab:grid} \\
\texttt{ablation} & \texttt{run\_all.py --steps ablation} & \texttt{ablation\_spar\_pull.csv} \\
\texttt{falseesc} & 300-stream calibration & \texttt{false\_esc.csv} \\
\texttt{regret} & horizon and dimension scaling & \texttt{regret\_*.csv} \\
\texttt{extra} & \texttt{ablation\_spar\_kf.py}, \texttt{dead\_direction.py}, \texttt{finite\_d\_conc.py}, \texttt{moving\_targets.py}, \texttt{sustained\_drift.py}, \texttt{transient\_test.py}, \texttt{bench\_stream\_real.py}, \texttt{bench\_stream\_extra.py} & \texttt{ablation\_spar\_kf.csv}, \texttt{dead\_direction.csv}, \texttt{finite\_d\_conc.csv}, \texttt{moving\_targets.csv}, \texttt{sustained\_drift.csv}, \texttt{transient\_test.csv}, \texttt{stream\_*.csv} \\
\texttt{ops} & wall-clock timing at $D=2048$ & \texttt{ops\_table.csv} \\
\texttt{eeg} & \texttt{bench\_eeg\_tier1.py} & \texttt{eeg\_tier1\_2*.csv} \\
\texttt{bp} & backprop baselines & \texttt{bp\_tta\_2*.csv} \\
\texttt{figures} & \texttt{make\_figures.py}, \texttt{make\_schematic.py}, \texttt{make\_geometry.py} & \texttt{results/figures/} \\
\bottomrule
\end{tabular}
\end{table}

\emph{Environment.}
Python 3.12 with the committed \texttt{requirements.txt} (JAX \citep{bradbury2018jax}, numpy, scipy, pandas, scikit-learn, matplotlib, SymPy \citep{meurer2017sympy}).
Float64 is required for the streaming solves and is enabled by setting \texttt{JAX\_ENABLE\_X64=1} (the drivers set it); a CPU runtime is recommended for the fp64 streams.
The backprop methods run on the WSL GPU virtual environment: \texttt{scripts/run\_bp\_tta\_2a.sh} and \texttt{scripts/run\_bp\_tta\_2b.sh}, with \texttt{BP\_PYTHON} pointing at that environment's interpreter and GPU extras in \texttt{requirements-wsl.txt}.

\emph{EEG data.}
Raw BCI IV files live under the MNE data directory (\texttt{MNE\_DATA} environment variable or the mne configuration; default \texttt{\~{}/mne\_data}) at the BNCI database path.
The loader prefers the npz cache: run \texttt{scripts/export\_eeg\_npz.py} once to produce \texttt{data/eeg\_cache/\{2a,2b\}/subject*.npz} for all nine subjects per data set.

\emph{Electricity data.}
The real stream of Appendix~\ref{app:realstream} reads the public hourly electricity-load panel (321 clients, $26{,}304$ rows, headerless comma-separated) from \texttt{data/ts/electricity.csv} (the loader also accepts the \texttt{.txt} variant); the file is a public download and is not shipped with the artifact.
It regenerates with \texttt{scripts/bench\_stream\_real.py --clients 11,82,171 --fresh} (\texttt{results/stream\_electricity.csv}); the script accepts any client indices, and the tables report the 25th-, 50th-, and 75th-percentile clients.
The transient strength sweep of Appendix~\ref{app:transient} regenerates with \texttt{scripts/transient\_test.py} (\texttt{results/transient\_test.csv}).
The grid-default BCI row of Table~\ref{tab:eeg} is the \texttt{hybrid-spar\_kf\_grid} column of \texttt{scripts/bench\_eeg\_tier1.py}; on 2a it is also reproduced by the cheaper \texttt{scripts/eeg\_grid\_default\_probe.py}, which saves \texttt{results/eeg\_griddefault\_2a.csv}.

\emph{Lean.}
The proof artifact builds with the elan-managed Lean 4 \citep{demoura2021} toolchain and the Mathlib library \citep{mathlib2020} via lake; the full recipe, including the Mathlib cache fetch, is \texttt{theory/lean/build\_ogd.sh} (Appendix~\ref{app:lean}).

\emph{Artifact layout.}
Every number in the paper is machine-printed into \texttt{results/*.csv} by the drivers above, and every figure is redrawn from those CSVs: the factorial tables (\texttt{factorial\_*}), the ablation (\texttt{ablation\_spar\_pull.csv}), the false-escalation calibration (\texttt{false\_esc.csv}), the regret scaling (\texttt{regret\_scaling.csv}), the cost table (\texttt{ops\_table.csv}), the EEG tables (\texttt{eeg\_tier1\_2*.csv}, \texttt{spar\_eeg\_2*.csv}, \texttt{bp\_tta\_2*.csv}), the estimator trajectory (\texttt{estimator\_trajectory.csv}), the real electricity stream (\texttt{stream\_electricity.csv}), and the transient strength sweep (\texttt{transient\_test.csv}).
The offline session-EA cov-LDA method and the post-hoc ensemble (members: the cov-LDA method and the static hybrid warm fit, identifier \texttt{ens-covlda+hybrid}) regenerate from the \texttt{eeg} step's baseline pass (\texttt{scripts/bench\_eeg.py}, recorded in \texttt{eeg\_2*.csv}); all other rows regenerate from the tier-1 pass.
The Lean material (seven module files including \texttt{NonVacuity.lean} and \texttt{TestAx.lean}, the root module, the build scripts, and the axiom-listing script) is under \texttt{theory/lean/}, with a theorem-by-theorem map in \texttt{SCOPE.md}.